\documentclass[10pt,twocolumn,letterpaper]{article}

\usepackage{cvpr}
\usepackage{times}
\usepackage{epsfig}
\usepackage{graphicx}
\usepackage{amsmath}
\usepackage{amssymb}

\usepackage{subcaption}
\usepackage[export]{adjustbox}

\usepackage{hyperref}
\usepackage{multirow}

\usepackage[dvipsnames]{xcolor}

\usepackage{pifont}

\usepackage{enumitem}
\setlist[enumerate]{itemsep=0mm}

\usepackage{authblk}

\cvprfinalcopy 

\ifcvprfinal\fi

\begin{document}

\title{Graphical Contrastive Losses for Scene Graph Parsing}


\renewcommand{\thefootnote}{\fnsymbol{footnote}}

\author[1,2]{Ji Zhang}
\author[2]{Kevin J. Shih}
\author[1]{Ahmed Elgammal}
\author[2]{Andrew Tao}
\author[2]{Bryan Catanzaro}
\affil[1]{Department of Computer Science, Rutgers University}
\affil[2]{Nvidia Corporation}
\affil[1]{{\tt\small \{jz462,elgammal\}@cs.rutgers.edu}}
\affil[2]{{\tt\small \{kshih,atao,bcatanzaro\}@nvidia.com}}

\renewcommand\Authands{, }

\maketitle

\begin{abstract}
Most scene graph parsers use a two-stage pipeline to detect visual relationships: the first stage detects entities, and the second predicts the predicate for each entity pair using a softmax distribution. We find that such pipelines, trained with only a cross entropy loss over predicate classes, suffer from two common errors. The first, Entity Instance Confusion, occurs when the model confuses multiple instances of the same type of entity (e.g. multiple cups). The second, Proximal Relationship Ambiguity, arises when multiple subject-predicate-object triplets appear in close proximity with the same predicate, and the model struggles to infer the correct subject-object pairings (e.g. mis-pairing musicians and their instruments). We propose a set of contrastive loss formulations that specifically target these types of errors within the scene graph parsing problem, collectively termed the Graphical Contrastive Losses. These losses explicitly force the model to disambiguate related and unrelated instances through margin constraints specific to each type of confusion. We further construct a relationship detector, called RelDN, using the aforementioned pipeline to demonstrate the efficacy of our proposed losses. Our model outperforms the winning method of the OpenImages Relationship Detection Challenge by 4.7\% (16.5\% relative) on the test set. We also show improved results over the best previous methods on the Visual Genome and Visual Relationship Detection datasets.
\end{abstract}


\section{Introduction}
\label{sec:introduction}

\begin{figure}[t!]
  \centering
  \begin{subfigure}{\columnwidth}
    \centering
    \begin{subfigure}{0.49\columnwidth}
      \centering
      \includegraphics[width=\textwidth]{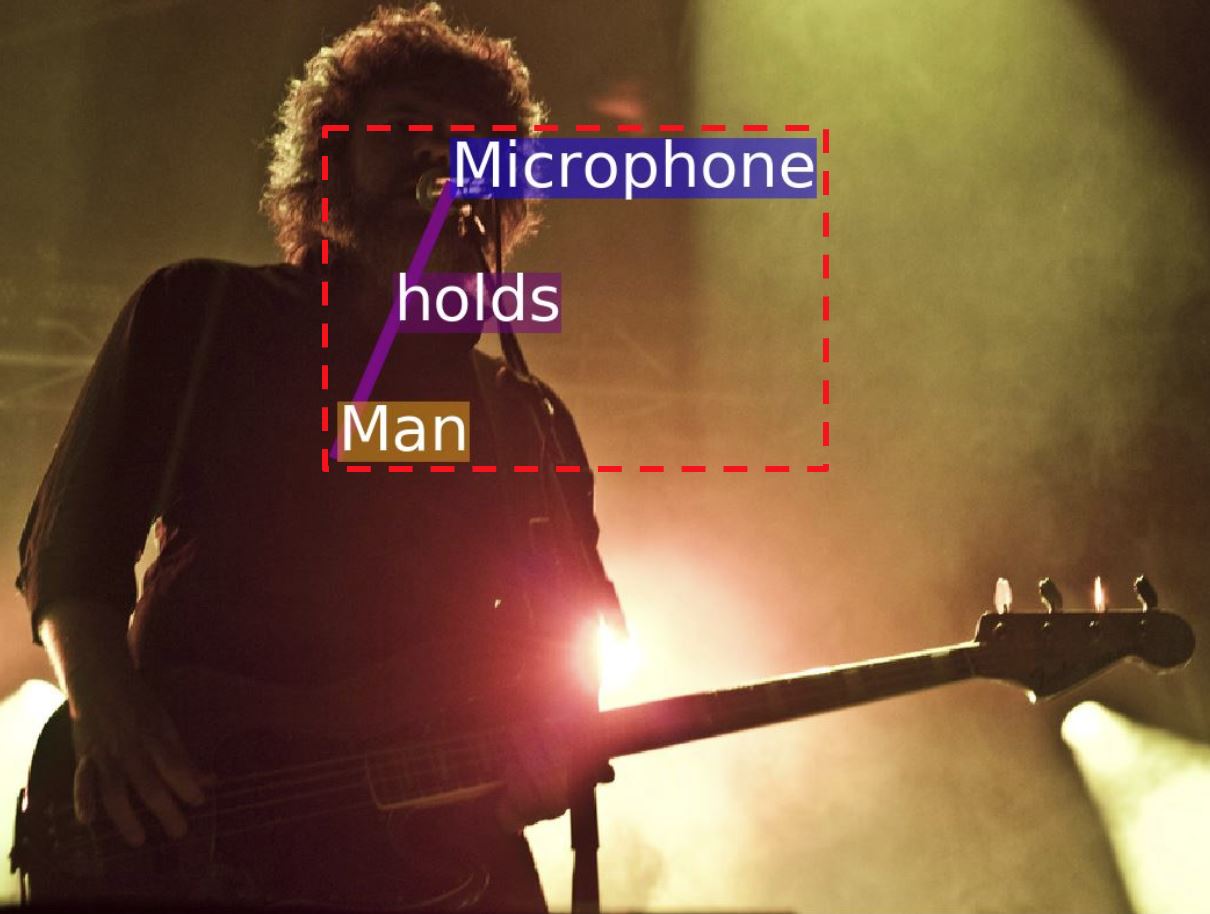}
    \caption{without our losses}
    \label{fig:teaser_L0}
    \end{subfigure}
    \centering
    \begin{subfigure}{0.49\columnwidth}
      \centering
      \includegraphics[width=\textwidth]{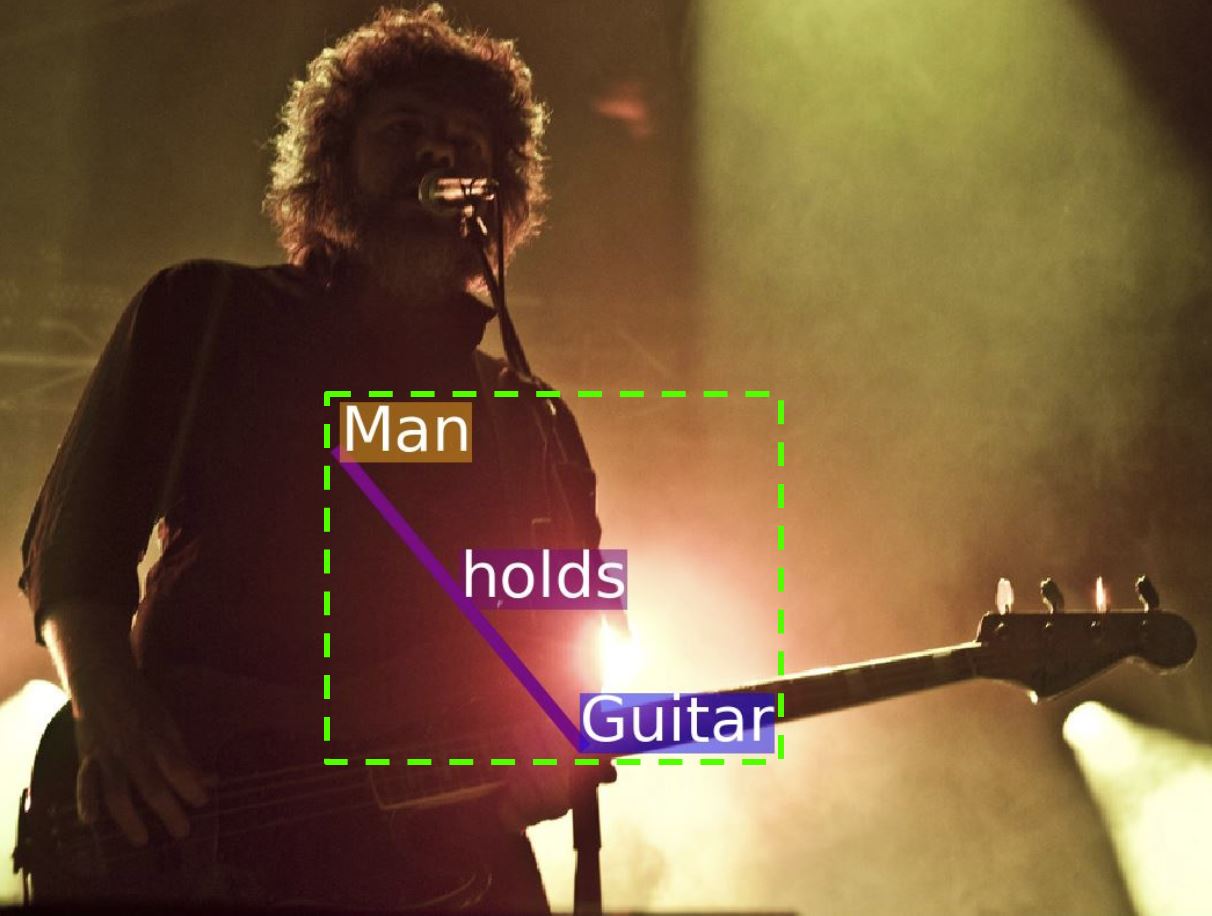}
    \caption{with our losses}
    \label{fig:teaser_ours}
    \end{subfigure}
  \end{subfigure}
\setlength\belowcaptionskip{-2ex}
\captionsetup{font=small}
\caption{Example of failure of models without our losses and success of our losses. (a) RelDN learned with only multi-class cross-entropy loss incorrectly relates the man with the microphone, while (b) RelDN learned with our \textit{Graphical Contrastive Losses} detects the correct relationship $\langle man, holds, guitar\rangle$.}
\label{fig:teaser}
\end{figure}

Given an image, the aim of scene graph parsing is to infer a visually grounded graph comprising localized entity categories, along with predicate edges denoting their pairwise relationships. This is often formulated as the detection of $\langle subject, predicate, object \rangle$ triplets within an image, e.g. $\langle man, holds, guitar \rangle$
 in  Figure \ref{fig:teaser_ours}.
Current state-of-the-art methods achieve this goal by a two-stage mechanism: first detecting entities,
then predicting a predicate for each pair of entities. 

\begin{figure}[t!]
  \centering
  \begin{subfigure}{0.49\columnwidth}
    \centering
    \begin{subfigure}{0.96\columnwidth}
      \centering
      \includegraphics[width=\linewidth]{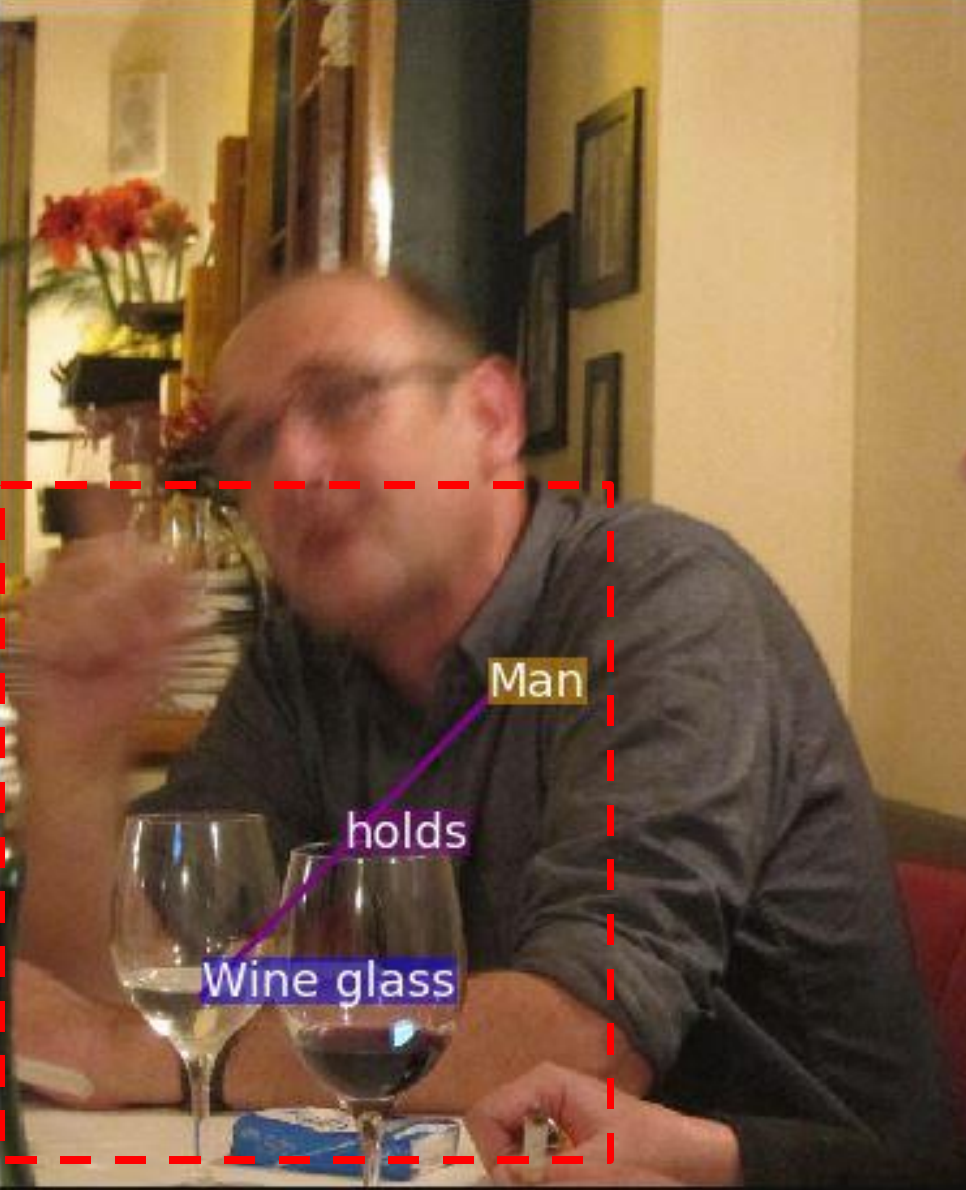}
    \end{subfigure}
  \captionsetup{font=small}
  \caption{Entity Instance Confusion}
  \label{fig:mo_1}
  \end{subfigure}
  \centering
  \begin{subfigure}{0.49\columnwidth}
    \centering
    \begin{subfigure}{0.9\columnwidth}
      \centering
      \includegraphics[width=\linewidth]{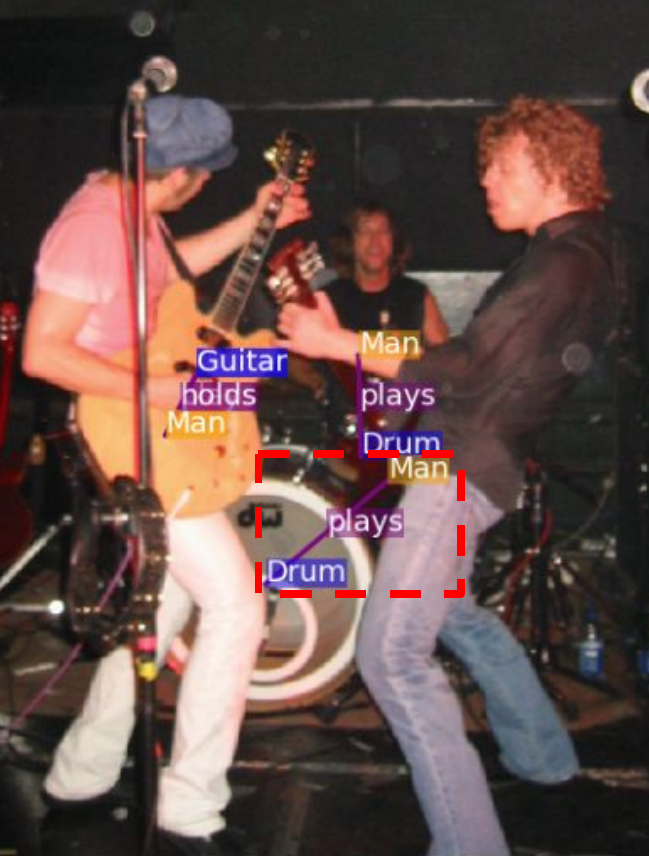}
    \end{subfigure}
  \captionsetup{font=small}
  \caption{Proximal Rel Ambiguity}
  \label{fig:mo_2}
  \end{subfigure}
\setlength\belowcaptionskip{-2ex}
\captionsetup{font=small}
\caption{Examples of Entity Instance Confusion and Proximal Relationship Ambiguity. Red boxes highlight relationships our baseline model predicts incorrectly. (a) the man is not holding the predicted wine glass. (b) the guitar player on the right is not playing drum.}
\label{fig:mo}
\end{figure}

We find that scene graph parsing models using such pipelines tend to struggle with two types of errors. The first is \textbf{Entity Instance Confusion}, in which the subject or object is related to one of many instances of the same class, and the model fails to distinguish between the target instance and the others. We show an example in Figure \ref{fig:mo_1}, in which the model identifies the man is holding a wine glass, but struggles to determine exactly which of the 3 visually similar wine glasses is being held. The incorrectly predicted wine glass is transparent and intersecting with the left arm, which makes it look like being held. The second type of error, \textbf{Proximal Relationship Ambiguity}, occurs when the image contains multiple subject-object pairs interacting in the same way, and the model fails to identify the correct pairing. An example can be seen in the multiple musicians "playing" their respective instruments in Figure \ref{fig:mo_2}. Due to their close proximity, visual features for each musician-instrument pair overlap significantly, making it difficult for the scene graph models to identify the correct pairings. 



The primary cause of these two failures lies in the inherent difficulty of inferring relationships like ``hold'' and ``play'' from visual cues. Concretely, which glass is being held is determined by the small part of the hand that covers the glass. Whether a player is playing the drum can only be inferred by very subtle visual cues such as his standing pose or where his fingers are placed. It is challenging for any model to learn to attend to these details precisely, and it would be impractical to specify which details to focus on for all kinds of relationships, let alone to learn all these details. These challenges motivate the need for a mechanism that can automatically learn fine details that determine visual relationships, and explicitly discriminate related entities from unrelated ones, for all types of relationships. This is the goal of our work.

In this paper we propose a set of \textit{Graphical Contrastive Losses} to tackle these issues. The losses use the form of the margin-based triplet loss, but 
are specifically designed to address the two aforementioned errors. It adds additional supervision in the form of hard negatives specific to Entity Instance Confusion and Proximal Relationship Ambiguity.
To demonstrate the effectiveness of our proposed losses, we design a relationship detection network named \textit{RelDN} using the aforementioned pipeline with our losses. Figure \ref{fig:teaser} shows a result of RelDN with N-way cross-entropy loss only \vs with our additional contrastive losses. Our best model achieves 0.328 on the Private set of the OpenImages Relationship Detection Challenge, outperforming the winning model by a significant 4.7\% (16.5\% relative) margin. It also attains state-of-the-art performance on the Visual Genome\cite{krishnavisualgenome} and VRD\cite{lu2016visual} datasets.


In this paper, we denote subject, predicate, object and attribute with  $s,pred,o,a$. We use ``entity'' to describe individual detected objects to distinguish from ``object'' in the semantic sense, and use ``relationships'' to describe the entire $\langle s,pred,o \rangle$ tuple, not to be confused with ``predicate," which is an element of said tuple.

\section{Related Work}
\label{sec:related}

\noindent\textbf{Scene Graph Parsing:} A large number of scene graph parsing approaches have emerged during the last couple of years. They use the same pipeline that first either uses off-the-shelf detectors \cite{lu2016visual,Zhuang_2017_ICCV,zhang2017visual,dai2017detecting,yu17iccv,Xu_2018_ECCV} or detectors fine-tuned with relationship datasets \cite{LiCVPR2017,xu2017scenegraph,zellers2018neural,zhang2017relationship,zhang2018large,Yin_2018_ECCV,Yang_2018_ECCV} to detect entities, then predicts the predicate using proposed methods. Most of them \cite{lu2016visual,Zhuang_2017_ICCV,zhang2017visual,dai2017detecting,yu17iccv,Yin_2018_ECCV,LiCVPR2017,xu2017scenegraph,zellers2018neural,zhang2017relationship,zhang2018neurips,zhang2018introduction} model the second step as a classification task that takes features of each entity pair as input and output a label independently from other pairs. \cite{zhang2018large} instead learn embeddings for subjects, predicates and objects and use nearest neighbor searching during testing to predict predicates. Nevertheless, the prediction is still done on each entity pair individually. We show that this pipeline struggles with two major scenarios. We find that ignoring the intrinsic graph structure of relationships and predicting each predicate separately is the main cause. Our proposed losses compensate for such drawback by contrasting positive against negative edges for each node, providing global supervision to the classifier and significantly alleviating those two issues.

The scene graph parsing work most related to ours is Associative Embedding \cite{newell2017pixels}. They use use a \emph{push} and \emph{pull} contrastive loss to train embeddings for entities within a visual genome scene graph. Our work differs in that we propose to have different sets of hard negatives to target specific error types within scene graph parsing.

\noindent\textbf{Phrase Grounding and Referring Expressions:}
Phrase grounding and referring expression models aim to localize the region described by a given expression, with the latter focusing more on cases of possible reference confusion \cite{yu2018mattnet,mao2016generation,yu2016modeling,nagaraja2016modeling,hu2016natural,luo2017comprehension,rohrbach2016grounding,wang2016learning,liu2017referring,chen2017query,hu2017modeling, plummer2015flickr30k}. It can be abstracted as a bipartite graph matching problem, where nodes on the visual side are the regions and nodes on the language side are the expressions, and the goal is to find all matched pairs. In contrast, scene graphs are arbitrarily connected graphs whose nodes are visual entities and edges are predicates with rich semantic information. Our losses are designed to leverage that information to better discriminate between related and non-related entities.


\noindent\textbf{Contrastive Training:} Contrastive training using a triplet loss \cite{kiros2014unifying} has wide application in both computer vision and natural language processing. Representative works include Negative Sampling \cite{mikolov2013distributed} and Noise Contrastive Sampling \cite{mnih2013learning}. More recent work also utilizes it to solve multi-modal tasks such as phrase grounding, image captioning, VQA, and vector embeddings \cite{wang2016learning,gupta2017aligned,yu2018mattnet,newell2017pixels}.
Our setting differs in that we define hard negative contrastive margins along the known structure of the annotated scene graph, allowing us to specifically target entity instance and proximal relationship confusion. By adding our losses as additional supervision on top of the N-way cross-entropy loss, we are able to improve the model by significant margins.



\section{Graphical Contrastive Losses}
\label{sec:gcloss}


Our Graphical Contrastive Losses encompass three types of loss, each addressing the two aforementioned issues in their own way: 1) \textbf{Class Agnostic}: contrasts positive/negative entity pairs regardless of their relation and adds contrastive supervision for generic cases; 2) \textbf{Entity Class Aware}: addresses the issue in Figure \ref{fig:mo_1} by focusing on entities with the same class; 3) \textbf{Predicate Class Aware}: addresses the issue in Figure \ref{fig:mo_2} by focusing on entity pairs with the same potential predicate. We define our contrastive losses over an affinity term $\Phi(s,o)$, which can be interpreted as the probability that subject $s$ and object $o$ have some relationship or interaction. Given a model that outputs the distribution over predicate classes conditioned on a subject and object pair $p(pred | s, o)$, we define $\Phi(s,o)$ as:
\begin{equation}
    \Phi(s,o) = 1 - p(pred = \varnothing | s, o)
\end{equation}
where $\varnothing$ is the class symbol representing \texttt{no\_relationship}. This is equivalent to summing over all predicate classes except $\varnothing$.

\subsection{Class Agnostic Loss}
Our first contrastive loss term aims to maximize the affinity of the lowest scoring positive pairing and minimize the affinity of the highest scoring negative pairing. For a subject indexed by $i$ and an object indexed by $j$, the margins we wish to maximize can be written as:
\begin{equation}
\begin{aligned}
m_1^s(i)=\min_{j \in \mathcal{V}_i^+} \Phi(s_i,o_j^+)-\max_{k \in \mathcal{V}_i^-} \Phi(s_i,o_k^-) \\
m_1^o(j)=\min_{i \in \mathcal{V}_j^+} \Phi(s_i^+,o_j)-\max_{k \in \mathcal{V}_j^-} \Phi(s_k^-,o_j)
\label{eq:m_1}
\end{aligned}
\end{equation}

where $\mathcal{V}_i^+$ and $\mathcal{V}_i^-$ represent sets of objects related to and not related to subject $s_i$; $\mathcal{V}_j^+$ and $\mathcal{V}_j^-$ are defined similarly for object $j$ as the sets of subjects related to and not related to $o_j$.

The class agnostic loss for all sampled positive subjects and objects is written as:
\begin{align}
\begin{split}
L_1=&\frac{1}{N}\sum_{i=1}^{N}\max(0,\alpha_1-m_1^s(i)) \\ +&\frac{1}{N}\sum_{j=1}^{N}\max(0,\alpha_1-m_1^o(j))
\end{split}
\label{eq:class_agnostic}
\end{align}
where $N$ is the number of annotated entities and $\alpha_1$ is the margin threshold. 

This loss tries to contrast positive and negative $(s,o)$ pairs, ignoring any class information, and is similar to the triplet losses used referring expression and phrase-grounding literature. We found it works as well in our scenario and even better with the following class-aware losses, as shown in Table \ref{tab:abl_L}.

\subsection{Entity Class Aware Loss}
The Entity Class Aware loss deals with entity instance confusion, in which the model struggles to determine interactions between a subject (object) and multiple instances of a same-class object (subject). It can be viewed as an extension of the Class Agnostic loss where we further specify a class $c$ when populating the positive and negative sets $\mathcal{V}^+$ and $\mathcal{V}^-$. We extend the formulation in equation (\ref{eq:class_agnostic}) as:
\begin{equation}
\begin{aligned}
m_2^s(i,c)=\min_{j \in \mathcal{V}_i^{c+}} \Phi(s_i,o_j^+)-\max_{k \in \mathcal{V}_i^{c-}} \Phi(s_i,o_k^-)\\
m_2^o(j,c)=\min_{i \in \mathcal{V}_j^{c+}} \Phi(s_i^+,o_j)-\max_{k \in \mathcal{V}_j^{c-}} \Phi(s_k^-,o_j)
\label{eq:m_2}
\end{aligned}
\end{equation}
where $\mathcal{V}_i^{c+}$, $\mathcal{V}_i^{c-}$, $\mathcal{V}_j^{c+}$ and $\mathcal{V}_j^{c-}$ are now constrained to instances of class $c$.

The entity class aware loss for all sampled positive subjects and objects is defined as
\begin{align}
\begin{split}
L_2=&\frac{1}{N}\sum_{i=1}^{N}\frac{1}{|\mathcal{C}(\mathcal{V}_i^+)|}\sum_{c\in\mathcal{C}(\mathcal{V}_i^+)}\max(0,\alpha_2-m_2^s(i,c)) \\
+&\frac{1}{N}\sum_{j=1}^{N}\frac{1}{|\mathcal{C}(\mathcal{V}_j^+)|}\sum_{c\in\mathcal{C}(\mathcal{V}_j^+)} \max (0,\alpha_2-m_2^o(j,c))
\end{split}
\label{eq:o_class_aware}
\end{align}
where $\mathcal{C}()$ returns the set of unique classes of the sets $\mathcal{V}_i^+$ and $\mathcal{V}_j^+$ as defined in the class agnostic loss. 
Compared to the class agnostic loss which maximizes the margins across all instances, this loss maximizes the margins between instances of the same class. It forces a model to disentangle confusing entities illustrated in Figure \ref{fig:mo_1}, where the subject has several potentially related objects with the same class.

\subsection{Predicate Class Aware Loss}
Similar to the entity class aware loss, this loss maximizes the margins within groups of instances determined by their associated predicates. It is designed to deal with the proximal relationship ambiguity as exemplified in Figure \ref{fig:mo_2}, where instances joined by the same predicate class are within close proximity of each other. In the context of Figure \ref{fig:mo_2}, this loss would encourage the correct pairing of who is playing which instrument by penalizing wrong pairing, \ie, ``man plays drum'' in the red box. Replacing the class groupings in equation (\ref{eq:m_2}) with predicate groupings restricted to predicate class $e$, we define our margins to maximize as:
\begin{align}
\begin{split}
m_3^s(i,e)=&\min_{j \in \mathcal{V}_i^{e+}} \Phi(s_i,o_j^+)-\max_{k \in \mathcal{V}_i^{e-}} \Phi(s_i,o_k^-)\\
m_3^o(j,e)=&\min_{i \in \mathcal{V}_j^{e+}} \Phi(s_i^+,o_j)-\max_{k \in \mathcal{V}_j^{e-}} \Phi(s_k^-,o_j)
\end{split}
\label{eq:m_3}
\end{align}

Here, we define the sets $\mathcal{V}_i^{e+}$ and $\mathcal{V}_j^{e+}$ as the sets of subject-object pairs where the ground truth predicate between $s_i$ and $o_j$ is $e$, anchored with respect to subject $i$ and object $j$ respectively. We define the sets $\mathcal{V}_i^{e-}$ and $\mathcal{V}_j^{e-}$ as is the set of instances where the model \emph{incorrectly predicts} (via argmax) the predicate to be $e$, anchored with respect to subject $i$ and object $j$ respectively.


The predicate class aware loss for all sampled positive subjects and objects is defined as
\begin{align}
\begin{split}
L_3=&\frac{1}{N}\sum_{i=1}^{N}\frac{1}{|\mathcal{E}(\mathcal{V}_i^+)|}\sum_{e\in\mathcal{E}(\mathcal{V}_i^+)}\max (0,\alpha_3-m_3^s(i,e)) \\
+&\frac{1}{N}\sum_{j=1}^{N}\frac{1}{|\mathcal{E}(\mathcal{V}_j^+)|}\sum_{e\in\mathcal{E}(\mathcal{V}_j^+)}\max (0,\alpha_3-m_3^o(j,e))
\end{split}
\label{eq:p_class_aware}
\end{align}
where $\mathcal{E}()$ returns the set of unique predicates associated with the input (excluding $\varnothing$). The final loss is expressed as:
\begin{align}
L = L_0 + \lambda_1 L_1 + \lambda_2 L_2 + \lambda_3 L_3
\label{eq:total}
\end{align}
where $L_0$ is the cross-entropy loss over predicate classes. 

\subsection{Complexity Analysis}

We look at the case where the subject $s_i$ is fixed and we vary object for positive/negative pairings. The reverse case (object fixed, subject varies) has the same complexity. All sampling is conducted on the entities of a single image per batch. The set of entities include ground truth bounding boxes, as well as any detector output with $>=0.5$ IOU to ground truth entities.

For the Class Agnostic Loss $L_1$, the computational complexity of the sampling procedure is $O(N^2)$, where $N$ is the upper bounded on number of sampled entities per image. In practice, for each subject, we randomly sample at most $K$ non-related objects (negative pairings), which makes the actual complexity $O(NK)$.

For the Entity Class Aware Loss $L_2$, the sampling procedure is the same as with $L_1$, except that we need to keep only those non-related objects that are of class $c$, \ie, the object class of the current $o$ in the sampled $(s, o)$ pair. This involves a filtering operation on the $K$ objects which takes $O(K)$ time, therefore the overall complexity is still $O(NK)$.

The analysis for the Predicate Class Aware Loss $L_3$ is similar to that of $L_2$, except that the filtering operation looks at the predicate class $e$ instead of the object class $c$. The overall complexity is also $O(NK)$.

We set $N=512$ and $K=64$ per batch in practice.

\begin{figure*}[t]
  \centering
  \includegraphics[width=0.9\textwidth]{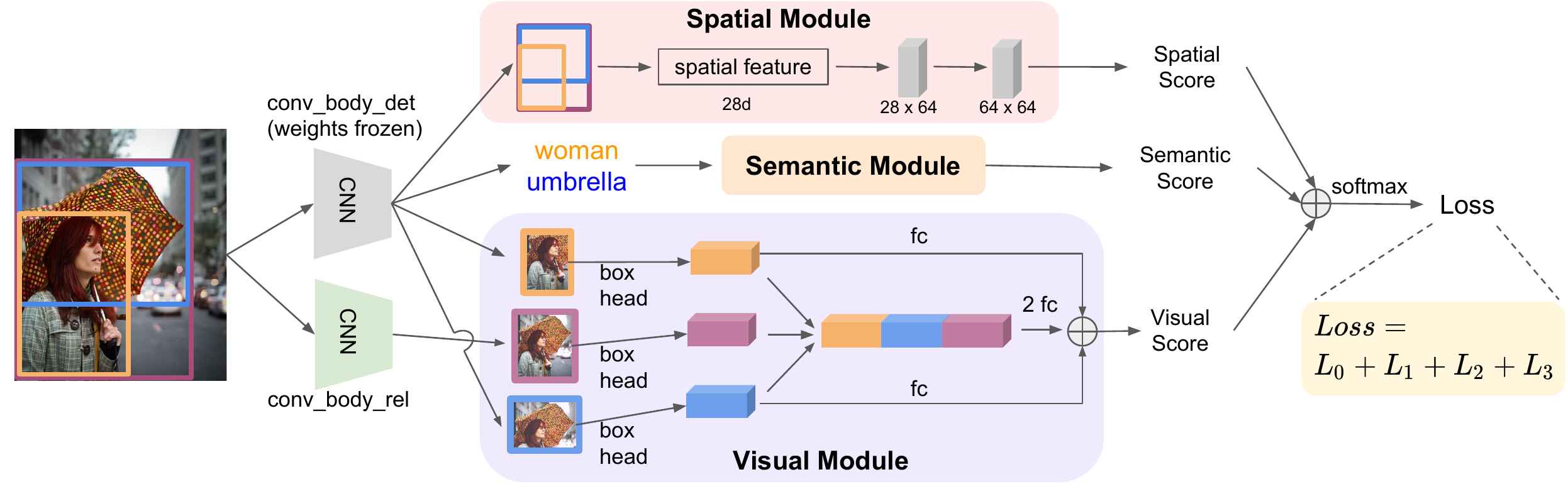}
\setlength\belowcaptionskip{-2ex}
\captionsetup{font=small}
\caption{The RelDN model architecture. The structures of conv\_body\_det and conv\_body\_rel are identical. We freeze the weights of the former and only train the latter.}
\label{fig:architecture}
\end{figure*}

\section{RelDN}
\label{sec:reldn}
\begin{figure}[t!]
  \centering
  \begin{subfigure}{\columnwidth}
    \centering
    \begin{subfigure}{0.32\columnwidth}
      \centering
      \includegraphics[width=\textwidth]{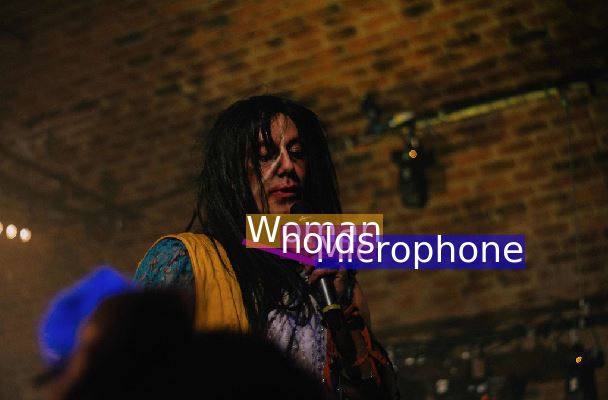}
    \end{subfigure}
    \centering
    \begin{subfigure}{0.32\columnwidth}
      \centering
      \includegraphics[width=\textwidth]{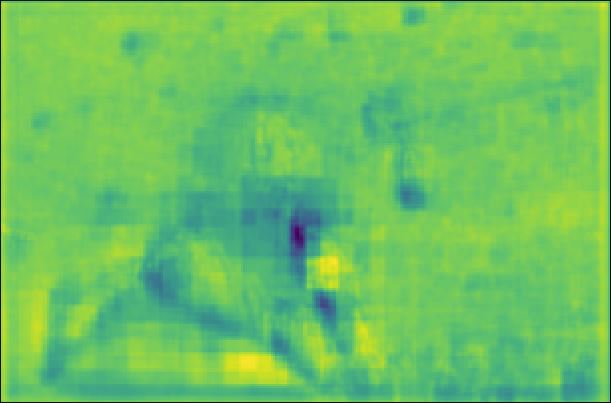}
    \end{subfigure}
    \centering
    \begin{subfigure}{0.32\columnwidth}
      \centering
      \includegraphics[width=\textwidth]{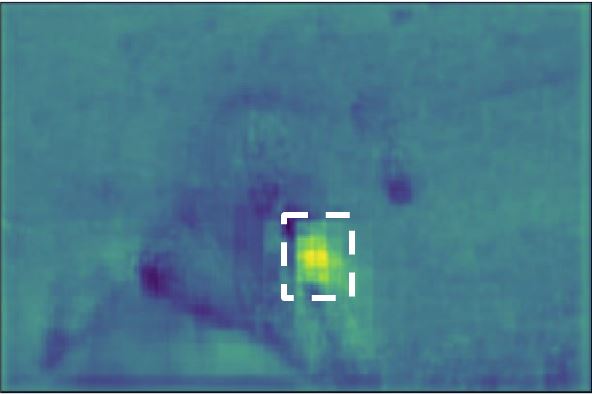}
    \end{subfigure}
  \end{subfigure}
  
  
  \centering
  \begin{subfigure}{\columnwidth}
    \centering
    \begin{subfigure}{0.32\columnwidth}
      \centering
      \adjincludegraphics[width=\textwidth,trim={0 {.05\height} 0 {.25\height}},clip]{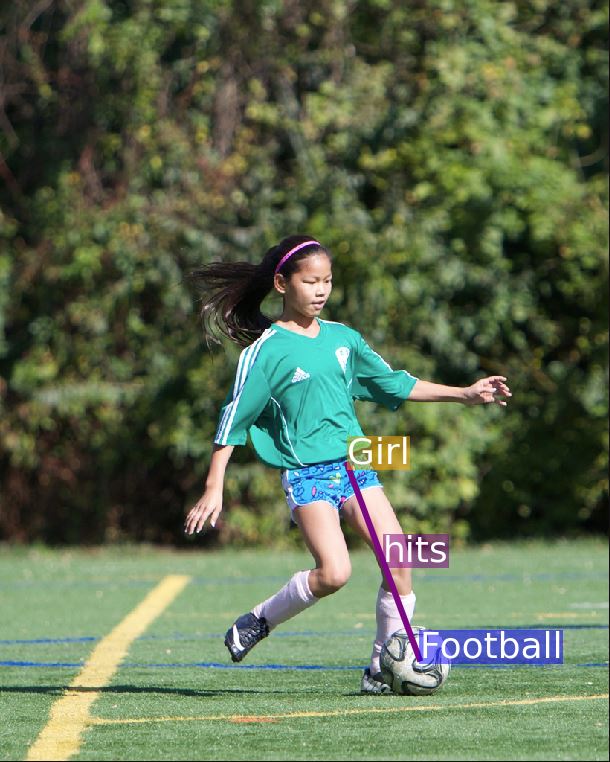}
    \caption{ground truth}
    \label{fig:gt}
    \end{subfigure}
    \centering
    \begin{subfigure}{0.32\columnwidth}
      \centering
      \adjincludegraphics[width=\textwidth,trim={0 {.05\height} 0 {.25\height}},clip]{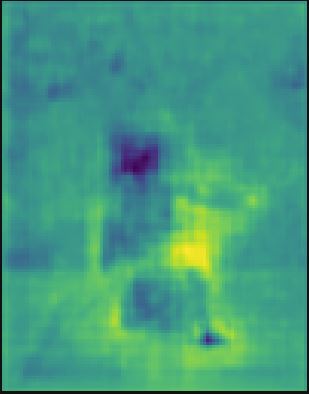}
    \caption{conv\_body\_det}
    \label{fig:det_conv}
    \end{subfigure}
    \centering
    \begin{subfigure}{0.32\columnwidth}
      \centering
      \adjincludegraphics[width=\textwidth,trim={0 {.05\height} 0 {.25\height}},clip]{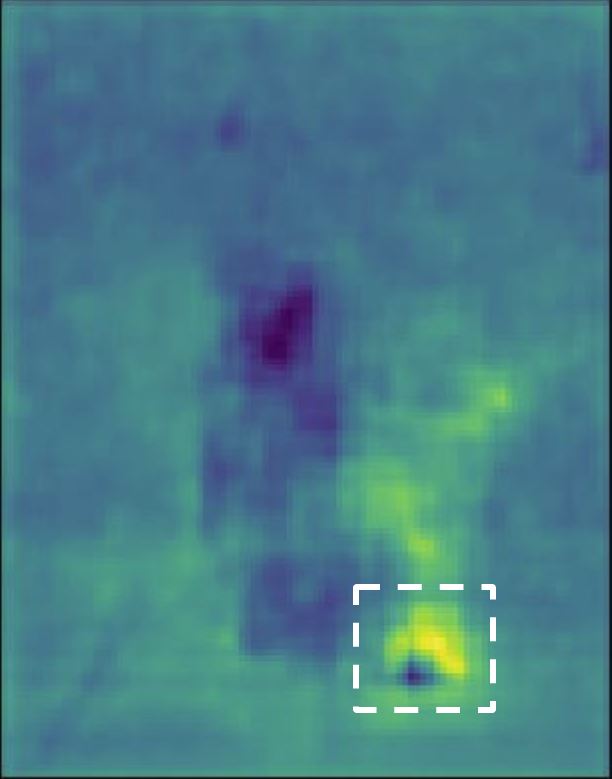}
    \caption{conv\_body\_rel}
    \label{fig:prd_conv}
    \end{subfigure}
  \end{subfigure}
  
  
\setlength\belowcaptionskip{-2ex}
\captionsetup{font=small}
\caption{Visualization of CNN features by averaging over the channel dimension of convolution feature maps \cite{zagoruyko2016paying}. (a) shows the image ground truth relationships, (b) shows the convolution feature from the entity detector backbone, and (c) shows the feature from the predicate backbone. In all the three examples there are clear shifts of salience from large entities to small areas that strongly indicate the predicates (highlighted in white boxes).}
\label{fig:vis_prd}
\end{figure}

We demonstrate the efficacy of our proposed losses with our Relationship Detection Network (RelDN). The RelDN follows a two stage pipeline: it first identifies a proposal set of likely subject-object relationship pairs, then extracts features from these candidate regions to perform a fine-grained classification into a predicate class. We build a separate CNN branch for predicates (conv\_body\_rel) with the same structure as that of entity detector CNN (conv\_body\_det) to extract predicate features. The intuition for having a separate branch is that we want visual features for predicates to focus on the interactive areas of subjects and objects as opposed to individual entities. As Figure \ref{fig:vis_prd} illustrates, the predicate CNN clearly learns better features which concentrate on regions that strongly imply relationships. 


The first stage of the RelDN exhaustively returns bounding box regions containing every pair. In the second stage, it computes three types of features for each relationship proposal: semantic, visual, and spatial. Each feature is used to output a set of class logits, which we combine via element-wise addition, and apply softmax normalization to attain a probability distribution over predicate classes. See Figure \ref{fig:architecture} for our model pipeline.


\noindent\textbf{Semantic Module:} The semantic module conditions the predicate class prediction on subject-object class co-occurrence frequencies. It is inspired by Zeller, et al. \cite{zellers2018neural} which introduced a frequency baseline that performs reasonably well on Visual Genome by counting frequencies of predicates given subject and object. Its motivation is that in general, the combination of relationships between two entities is usually very limited, \eg, the relationship between a person-horse subject-object pairing is most likely to be “ride”, “walk”, or “feed”, and unlikely to be “stand on” or “wear”. For each training image, we count the occurrences of predicate class $pred$ given subject and object classes $s$ and $o$ in the ground truth annotations. This gives us an empirical distribution $p(pred|s,o)$. We assume that the test set is also drawn from the same distribution. 

\noindent\textbf{Spatial Module:} The spatial module conditions the predicate class predictions on the relative positions of the subject and object. One of the major predicate types are about positions, for example, ``on'', ``under'', or ``inside\_of.'' These predicate types can often be inferred using only relative spatial information. We capture spatial information by encoding the box coordinates of subjects and objects using the box delta \cite{ren2015faster} and normalized coordinates.

We define the delta feature between two sets of bounding box coordinates as follows:
\begin{equation}
\Delta(b_1,b_2)=\langle\frac{x_1-x_2}{w_2},\frac{y_1-y_2}{h_2},\log\frac{w_1}{w_2},\log\frac{h_1}{h_2}\rangle
\end{equation}
where $b_1$ and $b_2$ are two coordinate tuples in the form of $(x, y, w, h)$.

We then compute the normalized coordinate features for a bounding box $b$ as follows:
\begin{equation}
c(b)=\langle\frac{x}{w_{img}},\frac{y}{h_{img}},\frac{x+w}{w_{img}},\frac{y+h}{h_{img}},\frac{wh}{w_{img}h_{img}}\rangle
\end{equation}
where $w_{img}$ and $h_{img}$ are the width and height dimensions of the image. Our spatial feature vector for the subject, object, and predicate bounding boxes $b_s$, $b_o$, $b_{pred}$ is represented as:
\begin{equation}
\langle\Delta(b_s,b_o),\Delta(b_s,b_{pred}),\Delta(b_{pred},b_o),c(b_s),c(b_o)\rangle
\end{equation}
Note that $b_{pred}$ is the tightest bounding box around $b_s$ and $b_o$. This feature vector is fed through an MLP to attain predicate class logit scores.

\noindent\textbf{Visual Module:} The visual module produces a set of class logits conditioned ROI feature maps, as in the fast-RCNN pipeline. We extract subject and object ROI features from the entity detector's convolution layers (conv\_body\_det in Figure \ref{fig:architecture}) and extract predicate ROI features from the relationship convolution layers (conv\_body\_rel in Figure \ref{fig:architecture}). The subject, object, and predicate feature vectors are concatenated and passed through an MLP to attain the predicate class logits.

We also include two skip-connections projecting subject-only and object-only ROI features to the predicate class logits. These skip connections are inspired by the observation that many relationships, such as human interactions \cite{gkioxari2017interactnet}, can be accurately inferred by the appearance of only the subjects or objects. We show an improvement from adding these skip connections in \ref{subsec:model_analysis}.


\noindent\textbf{Module Fusion:} As illustrated in Figure \ref{fig:architecture}, we obtain the final probability distribution over predicate classes by adding the three scores followed by softmax normalization:
\begin{align}
\textbf{p}^{pred}=softmax(\textbf{f}_{vis}+\textbf{f}_{spt}+\textbf{f}_{sem})
\label{eq:p_pred}
\end{align}
where $\textbf{f}_{vis},\textbf{f}_{spt},\textbf{f}_{sem}$ are unnormalized class logits from the visual, spatial, semantic modules. 

\section{Implementation Details}
We train the entity detector CNN (conv\_body\_det) independently using entity annotations, then fix it when training our model. While previous works \cite{LiCVPR2017,dai2017detecting,Yin_2018_ECCV} claim it is beneficial to fine-tune the entity detector end-to-end with the second stage of the pipeline, we opt to freeze our entity detector weights for simplicity. We initialize the predicate CNN (conv\_body\_rel) with the entity detector's weights and fine-tune it end-to-end with the second stage.

During training, we independently sample positive and negative pairs for each loss, subject to their respective constraints. For $L_0$, we sample 512 pairs in total where 128 of them are positive. For our class-agnostic loss, we sample 128 positive subjects, then for each of them sample the two closet contrastive pairs according to Eq.\ref{eq:m_1}; we do the sampling symmetrically for objects. For our entity and predicate aware losses, we sample in the same way with class-agnostic except that negative pairs are grouped by entity and predicate classes, as described in Eq.\ref{eq:m_2},\ref{eq:m_3}. We set $\lambda_1=1.0, \lambda_2=0.5, \lambda_3=0.1$, determined by cross-validations, for all experiments.

During testing, we take up to 100 outputs from the entity detector and exhaustively group all pairs as relationship proposals/entity pairs. We rank relationship proposals by multiplying the predicted subject, object, predicate probabilities as $\textbf{p}^{det}(s)\cdot \textbf{p}^{pred}(pred)\cdot \textbf{p}^{det}(o)$ where $\textbf{p}^{det}(s),\textbf{p}^{det}(o)$ are the probabilities of the predicted subject and object classes from the entity detector, and $\textbf{p}^{pred}(pred)$ is the probability of the predicted predicate class from the result of Eq.\ref{eq:p_pred}.


To match the architectures of previous state-of-the-art methods, We use ResNeXt-101-FPN \cite{xie2017aggregated,lin2017feature} as our OpenImages backbone and VGG-16 on Visual Genome (VG) and Visual Relationship Detection (VRD).


\section{Experiments}
\label{sec:experiments}

\begin{table*}[t!]
\centering
\resizebox{2.1\columnwidth}{!}{
\begin{tabular}{c c c c | c c c c | c c c c c c c c c }
\hline
& & & & & & & & \multicolumn{9}{c}{AP$_{rel}$ per class} \\
$L_0$ & $L_1$ & $L_2$ & $L_3$ & R@50 & wmAP$_{rel}$ & wmAP$_{phr}$ & score$_{wtd}$ & at & on & holds & plays & interacts\_with & wears & inside\_of & \textit{under} & \textit{hits} \\
\hline
\hline
\checkmark & & & & 74.67 & 34.63 & 37.89 & 43.94 & 32.40 & 36.51 & 41.84 & 36.04 & 40.43 & 5.70 & 44.17 & \textit{25.00} & \textit{55.40} \\
\checkmark & \checkmark & & & 75.06 & 35.25 & 38.37 & 44.46 & 32.78 & 36.96 & 42.93 & 37.55 & 43.30 & \textbf{9.01} & 44.15 & \textit{100.00} & \textit{50.95} \\
\checkmark & & \checkmark & & 74.64 & 35.03 & 38.18 & 44.21 & 32.76 & 36.82 & 42.24 & 37.17 & 40.47 & 8.53 & 44.71 & \textit{33.33} & \textit{49.68} \\
\checkmark & & & \checkmark & 74.88 & 35.19 & 38.27 & 44.36 & 32.88 & 36.73 & 42.38 & 38.03 & 43.53 & 6.71 & 44.18 & \textit{16.67} & \textit{52.06} \\
\checkmark & \checkmark & \checkmark & & 75.03 & 35.38 & 38.50 & 44.56 & \textbf{32.95} & \textbf{37.10} & 42.82 & 38.58 & 43.66 & 6.79 & 43.72 & \textit{20.00} & \textit{50.24} \\
\checkmark & \checkmark & & \checkmark & \textbf{75.30} & 35.30 & 38.27 & 44.49 & 32.92 & 36.73 & 42.58 & 38.81 & 44.13 & 6.35 & 42.74 & \textit{100.00} & \textit{51.40} \\
\checkmark & & \checkmark & \checkmark & 75.00 & 35.12 & 38.34 & 44.39 & 32.79 & 36.47 & 42.31 & 39.74 & 41.35 & 6.11 & 43.57 & \textit{25.00} & \textit{55.12} \\
\checkmark & \checkmark & \checkmark & \checkmark & 74.94 & \textbf{35.54} & \textbf{38.52} & \textbf{44.61} & 32.92 & 37.00 & \textbf{43.09} & \textbf{41.04} & \textbf{44.16} & 7.83 & \textbf{44.72} & \textit{50.00} & \textit{51.04} \\
\hline
\end{tabular}
}
\captionsetup{font=small}
\caption{Ablation Study on our losses. We report a frequency-balanced wmAP instead of mAP, as the test set is extremely imbalanced and would fluctuate wildly otherwise (see fluctuations in columns ``under" and ``hits"). We also report score$_{wtd}$, which is the official OI scoring formula but with wmAP in place of mAP. ``Under'' and ``hits'' are not highlighted due to having too few instances.}
\label{tab:abl_L}
\end{table*}

\begin{table*}[t!]
\centering
\resizebox{2.1\columnwidth}{!}{
\begin{tabular}{c | c c c c | c c c c c | c c c c c}
\hline
 & & & & & \multicolumn{5}{c|}{AP$_{rel}$ per class} & \multicolumn{5}{c}{AP$_{phr}$ per class} \\
 & R@50 & wmAP$_{rel}$ & wmAP$_{phr}$ & score$_{wtd}$ & at & holds & plays & interacts\_with & wears & at & holds & plays & interacts\_with & wears \\
\hline
\hline
$L_0$ & 61.72 & 25.80 & 33.15 & 35.92 & 14.77 & 26.34 & 42.51 & 21.33 & 21.03 & 21.76 & 35.88 & \textbf{48.57} & 38.74 & 31.92 \\
$L_0+L_1+L_2+L_3$ & \textbf{62.65} & \textbf{27.37} & \textbf{34.58} & \textbf{37.31} & \textbf{16.18} & \textbf{30.39} & \textbf{42.73} & \textbf{22.40} & \textbf{22.14} & \textbf{22.67} & \textbf{39.60} & 48.09 & \textbf{40.96} & \textbf{32.64} \\
\hline
\end{tabular}
}
\captionsetup{font=small}
\caption{Comparison of our model with Graphical Contrastive Loss \vs without the loss on 100 images containing the 5 classes that suffer from the two aforementioned confusions, selected via visual inspection on a random set of images.}
\label{tab:abl_special_img}
\end{table*}

\begin{table*}[t!]
\centering
\resizebox{1.9\columnwidth}{!}{
\begin{tabular}{c | c c c c | c c c c c c c c c }
\hline
 & & & & & \multicolumn{9}{c}{AP$_{rel}$ per class} \\
 & R@50 & wmAP$_{rel}$ & wmAP$_{phr}$ & score$_{wtd}$ & at & on & holds & plays & interacts\_with & wears & inside\_of & \textit{under} & \textit{hits} \\
\hline
\hline
sem only & 72.98 & 28.73 & 33.07 & 39.32 & 28.62 & 24.52 & 37.04 & 27.33 & 38.37 & 3.16 & 16.34 & \textit{25.00} & \textit{38.45} \\
sem + $\langle$S,P,O$\rangle$ & 74.97 & 34.70 & 37.96 & 44.06 & 32.26 & 36.26 & 42.44 & 38.47 & 41.63 & 6.50 & 40.97 & \textit{20.00} & \textit{54.38} \\
sem + vis & \textbf{75.12} & 35.22 & 38.33 & 44.44 & 32.68 & 36.83 & 42.09 & \textbf{41.53} & 42.58 & \textbf{8.49} & 42.31 & \textit{33.33} & \textit{53.95} \\
sem + vis + spt & 74.94 & \textbf{35.54} & \textbf{38.52} & \textbf{44.61} & \textbf{32.92} & \textbf{37.00} & \textbf{43.09} & 41.04 & \textbf{44.16} & 7.83 & \textbf{44.72} & \textit{50.00} & \textit{51.04} \\
\hline
\end{tabular}
}
\captionsetup{font=small}
\caption{Ablation Study on RelDN modules. \textit{sem only} means using only the semantic module without training any model; \textit{$\langle$S,P,O$\rangle$} means using only the $\langle$S,P,O$\rangle$ concatenation without the separate S,O layers in the visual module; \textit{vis} means our full visual module, and \textit{spt} means spatial module. ``Under'' and ``hits'' are not highlighted due to having too few instances.}
\label{tab:abl_model}
\end{table*}

\begin{table}[t!]
\centering
\resizebox{0.7\columnwidth}{!}{
\begin{tabular}{c | c c c c}
\hline
 & R@50 & wmAP$_{rel}$ & wmAP$_{phr}$ & score$_{wtd}$ \\
\hline
\hline
m = 0.1 & \textbf{75.09} & 35.29 & 38.43 & 44.51 \\
m = 0.2 & 74.94 & \textbf{35.54} & \textbf{38.52} & \textbf{44.61} \\
m = 0.5 & 74.64 & 35.14 & 38.39 & 44.34 \\
m = 1.0 & 74.28 & 34.17 & 37.75 & 43.62 \\
\hline
\end{tabular}
}
\setlength\belowcaptionskip{-2ex}
\captionsetup{font=small}
\caption{Ablation Study on the margin threshold m. We use $m=0.2$ everywhere in our experiments.}
\label{tab:abl_m}
\end{table}


We present experimental results on three datasets: OpenImages (OI) \cite{openimages}, Visual Genome (VG) \cite{krishnavisualgenome} and Visual Relationship Detection (VRD) \cite{lu2016visual}. We first report evaluation settings, followed by ablation studies and finally external comparisons.

\subsection{Evaluation Settings}


\noindent\textbf{OpenImages:} The full train and val sets contains 53,953 and 3,234 images, which takes our model 2 days to train. For quick comparisons, we sample a ``mini" subset of 4,500 train and 1,000 validation images where predicate classes are sampled proportionally with a minimum of one instance per class in train and val. We first conduct parameter searches on the mini set, then train and compare with the top model of the OpenImages VRD Challenge \cite{openimages_challenge} on the full set. We show two types of results, one using the same entity detector from the top model, and the other using a detector trained by our own initialized by COCO pre-trained weights. 

In the OpenImages Challenge, results are evaluated by calculating Recall@50 (R@50), mean AP of relationships (mAP$_{rel}$), and mean AP of phrases (mAP$_{phr}$). The final score is obtained by $\text{score}=0.2 \times R@50 + 0.4 \times mAP_{rel} + 0.4 \times mAP_{phr}$. The mAP$_{rel}$ evaluates AP of $s,pred,o$ triplets where \emph{both} the subject and object boxes have an IOU of at least 0.5 with ground truth. The mAP$_{phr}$ is similar, but applied to the enclosing relationship box\footnote{More details of evaluation can be found on the official page: \url{https://storage.googleapis.com/openimages/web/vrd_detection_metric.html}}. In practice, we find mAP$_{rel}$ and mAP$_{phr}$ to suffer from extreme predicate class imbalance. For example, 64.48\% of the relationships in val have the predicate ``at'', while only 0.03\% of them are ``under''. This means a single ``under" relationship is worth much more than the more common ``at" relationships. We address this by scaling each predicate category by their relative ratios in the val set, which we refer to as the weighted mAP (wmAP). We use wmAP in all of our ablation studies (Table \ref{tab:abl_L}-\ref{tab:abl_m}), in addition to reporting score$_{wtd}$ which replaces mAP with wmAP in the score formula.

\begin{table}[t!]
\centering
\resizebox{\columnwidth}{!}{
\begin{tabular}{c c c c | c c c c c c c }
\hline
$L_0$ & $L_1$ & $L_2$ & $L_3$ & R@50 & mAP$_{rel}$ & mAP$_{phr}$ & score & mAP$_{rel}^\textbf{*}$ & mAP$_{phr}^\textbf{*}$ & score$^\textbf{*}$ \\
\hline
\hline
\checkmark & & & & 74.67 & 35.28 & 41.04 & 45.46 & 33.87 & 38.99 & 44.08 \\
\checkmark & \checkmark & & & 75.06 & \textbf{44.18} & \textbf{50.19} & \textbf{52.76} & 35.24 & 40.30 & 45.23 \\
\checkmark & & \checkmark & & 74.64 & 36.19 & 41.71 & 46.09 & 34.67 & 39.61 & 44.64 \\
\checkmark & & & \checkmark & 74.88 & 34.80 & 40.47 & 45.08 & 34.92 & 40.01 & 44.95 \\
\checkmark & \checkmark & \checkmark & & 75.03 & 35.10 & 41.18 & 45.52 & 35.09 & 40.22 & 45.13 \\
\checkmark & \checkmark & & \checkmark & \textbf{75.30} & 43.96 & 49.61 & 52.49 & 34.89 & 39.87 & 44.96 \\
\checkmark & & \checkmark & \checkmark & 75.00 & 35.83 & 41.32 & 45.86 & 34.62 & 39.70 & 44.73 \\
\checkmark & \checkmark & \checkmark & \checkmark & 74.94 & 39.09 & 44.47 & 48.41 & \textbf{35.82} & \textbf{40.43} & \textbf{45.49} \\
\hline
\end{tabular}
}
\captionsetup{font=small}
\setlength\belowcaptionskip{-2ex}
\caption{Ablation Study on our losses with the official mAP$_{rel}$, mAP$_{phr}$ and score metrics. Metric marked with a \textbf{*} means ``under'' and ``hits'' are excluded from evaluation. The fluctuating numbers in mAP$_{rel}$, mAP$_{phr}$ and score indicate that the mAP metrics are unstable and unreliable, while when ``under'' and ``hits'' are excluded, all the results become consistent with Table \ref{tab:abl_L}.}
\label{tab:abl_L_no_weights}
\end{table}

\begin{table}[t!]
\centering
\resizebox{0.85\columnwidth}{!}{
\begin{tabular}{c | c c c c }
\hline
& R@50 & mAP$_{rel}$ & mAP$_{phr}$ & score \\
\hline
\hline
$L_0$ & 61.72 & 25.20 & 35.37 & 36.57 \\
$L_0+L_1+L_2+L_3$ & \textbf{62.65} & \textbf{26.77} & \textbf{36.79} & \textbf{37.95} \\
\hline
\end{tabular}
}
\captionsetup{font=small}
\caption{Comparison of our model with Graphical Contrastive Loss \vs without the loss on 100 images containing the 5 classes that suffer from the two aforementioned confusions, selected via visual inspection on a random set of images. The metrics are the official mAP$_{rel}$, mAP$_{phr}$ and the score. The ``under'' and ``hits'' predicates are not in this 100 image subset.}
\label{tab:abl_special_img_no_weights}
\end{table}

\begin{figure*}[t!]
  \centering
  \begin{minipage}{\linewidth}
    \centering
    \begin{subfigure}{0.33\columnwidth}
        \centering
        \begin{subfigure}{0.492\columnwidth}
          \centering
          \adjincludegraphics[width=\textwidth,trim={0 0 0 {.2\height}},clip]{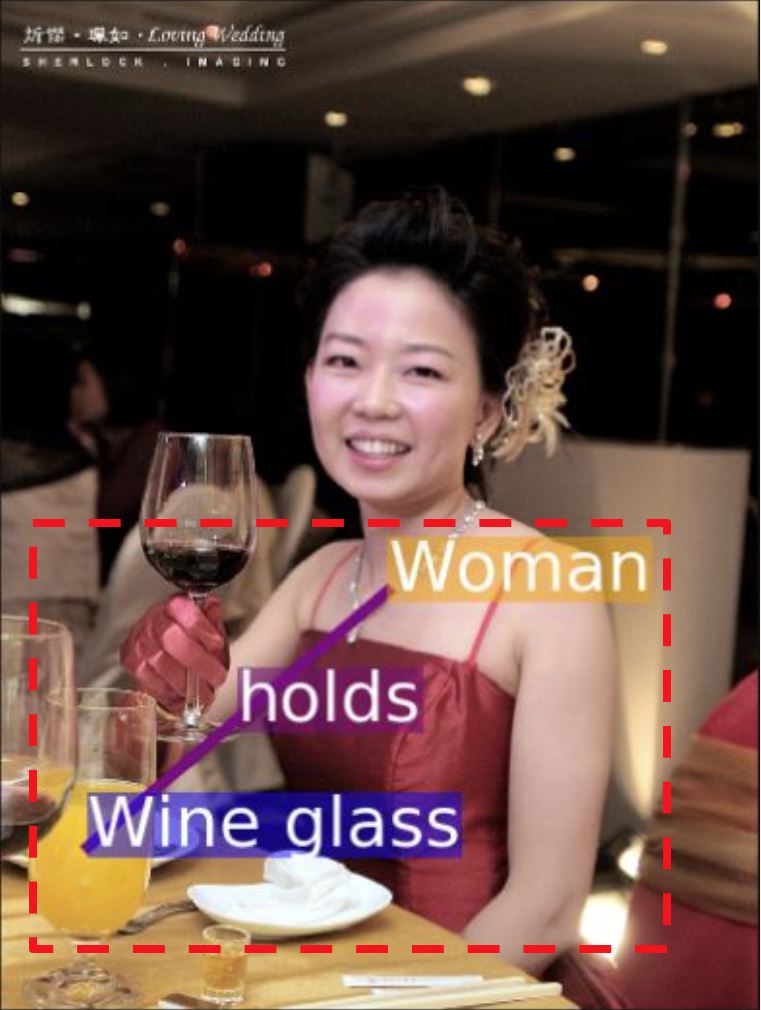}
        \end{subfigure}
        \centering
        \begin{subfigure}{0.49\columnwidth}
          \centering
          \adjincludegraphics[width=\textwidth,trim={0 0 0 {.2\height}},clip]{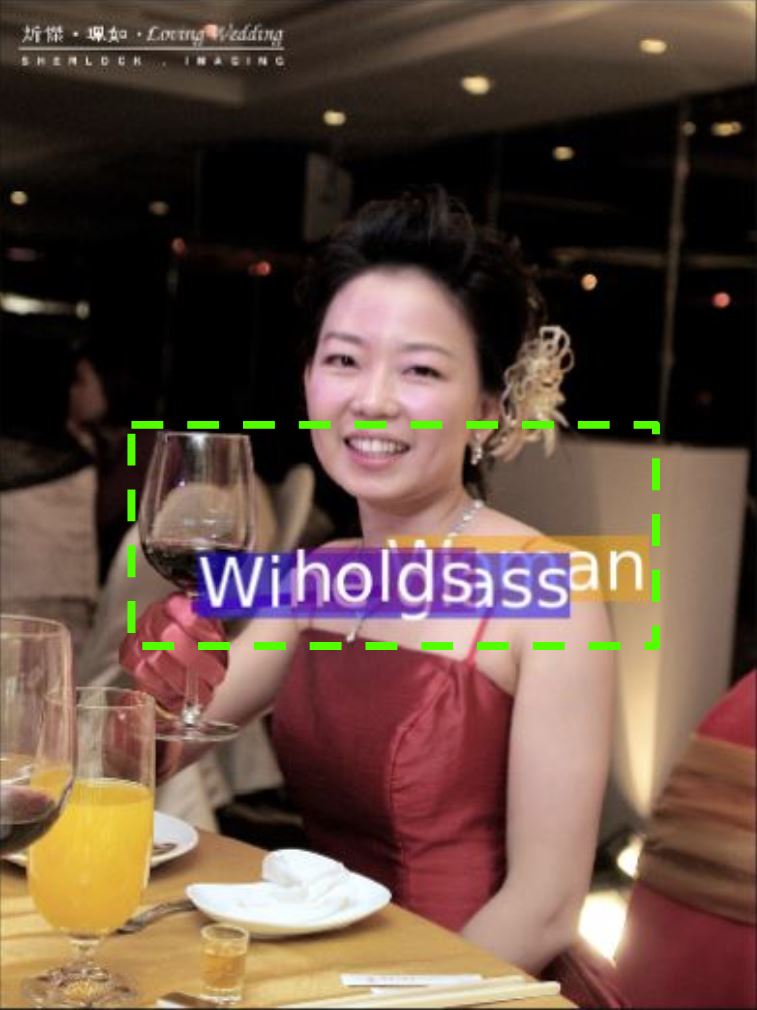}
        \end{subfigure}
    \end{subfigure}
    \centering
    \begin{subfigure}{0.63\columnwidth}
        \centering
        \begin{subfigure}{0.488\columnwidth}
          \adjincludegraphics[width=\textwidth,trim={0 0 0 {.2\height}},clip]{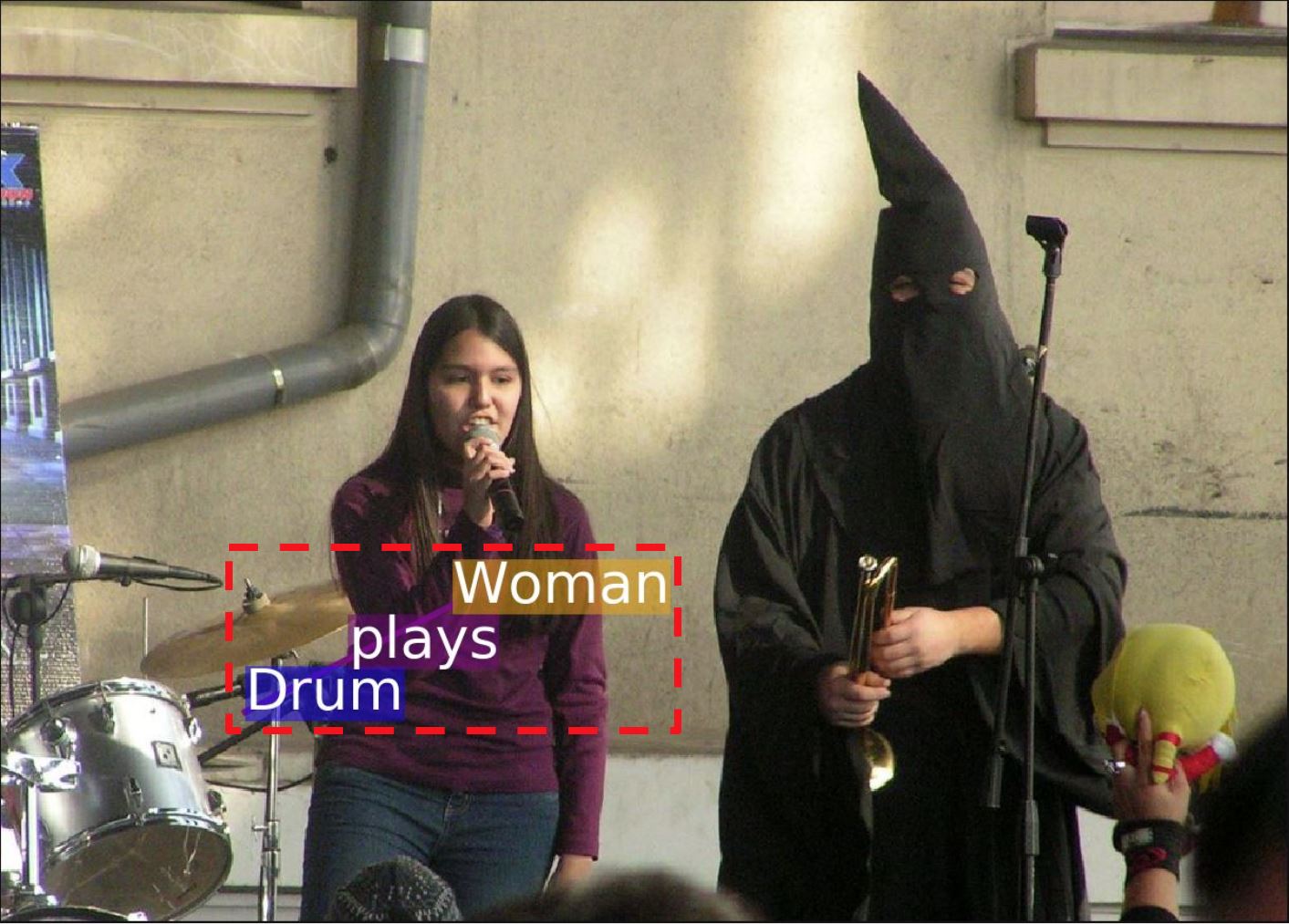}
        \end{subfigure}
        \centering
        \begin{subfigure}{0.49\columnwidth}
          \centering
          \adjincludegraphics[width=\textwidth,trim={0 0 0 {.2\height}},clip]{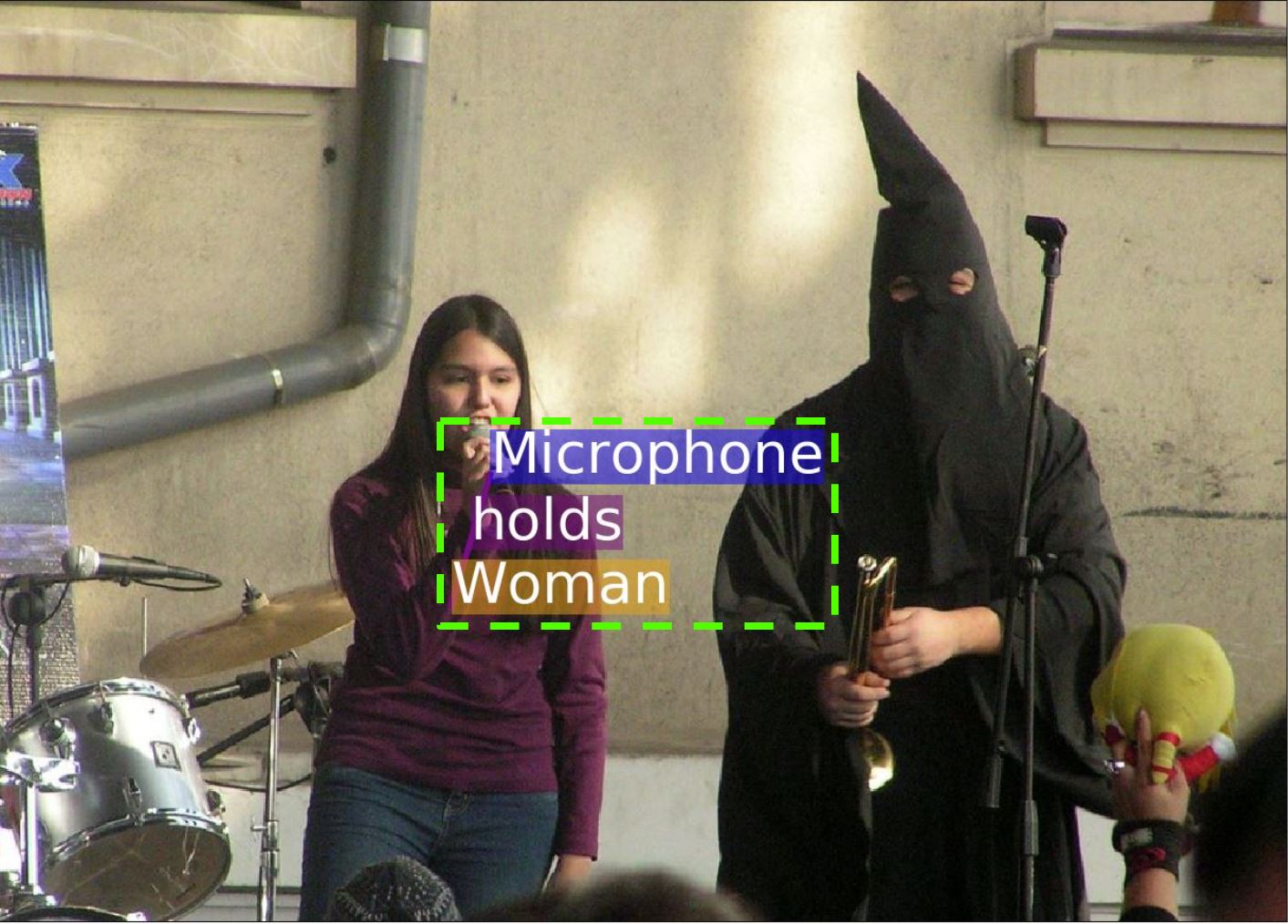}
        \end{subfigure}
    \end{subfigure}   
  \end{minipage}
  \centering
  \begin{minipage}{\linewidth}
    \centering
    \begin{subfigure}{0.33\columnwidth}
        \centering
        \begin{subfigure}{0.487\columnwidth}
          \centering
          \adjincludegraphics[width=\textwidth,trim={0 0 0 {.2\height}},clip]{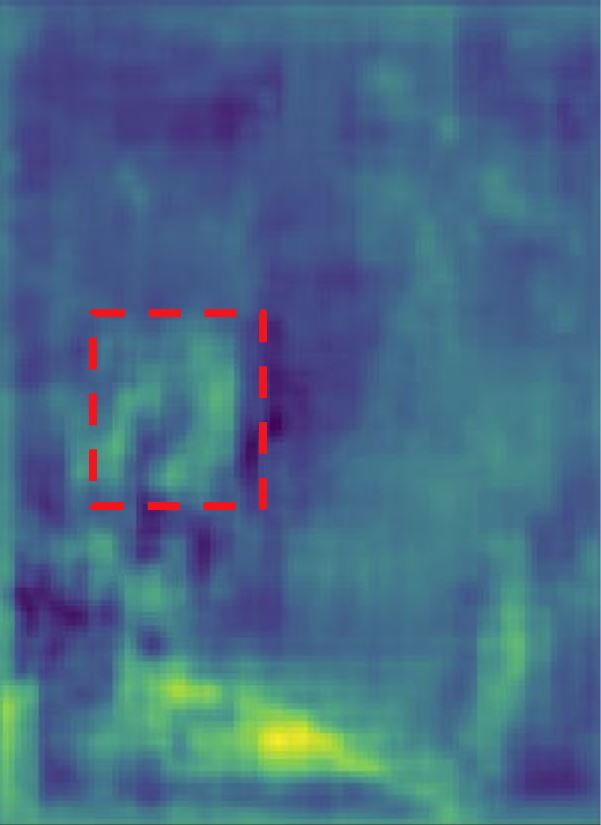}
        \captionsetup{font=small}
        \setlength\abovecaptionskip{-2ex}
        \caption*{$L_0$ only}
        \end{subfigure}
        \centering
        \begin{subfigure}{0.493\columnwidth}
          \centering
          \adjincludegraphics[width=\textwidth,trim={0 0 0 {.2\height}},clip]{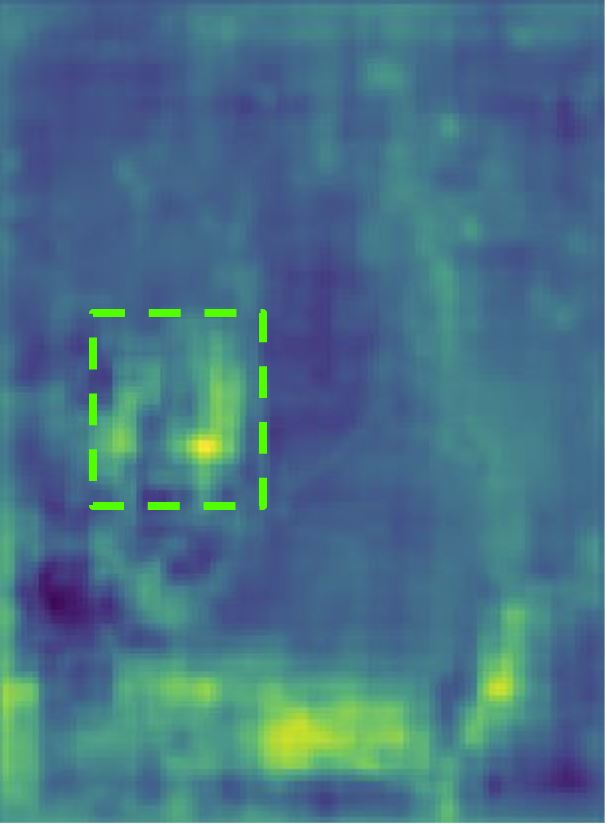}
        \captionsetup{font=small}
        \setlength\abovecaptionskip{-2ex}
        \caption*{with our losses}
        \end{subfigure}
    \captionsetup{font=small}
    \caption{Entity Instance Confusion}
    \label{fig:entity_confusion}
    \end{subfigure}
    \centering
    \begin{subfigure}{0.63\columnwidth}
        \centering
        \begin{subfigure}{0.49\columnwidth}
          \adjincludegraphics[width=\textwidth,trim={0 0 0 {.2\height}},clip]{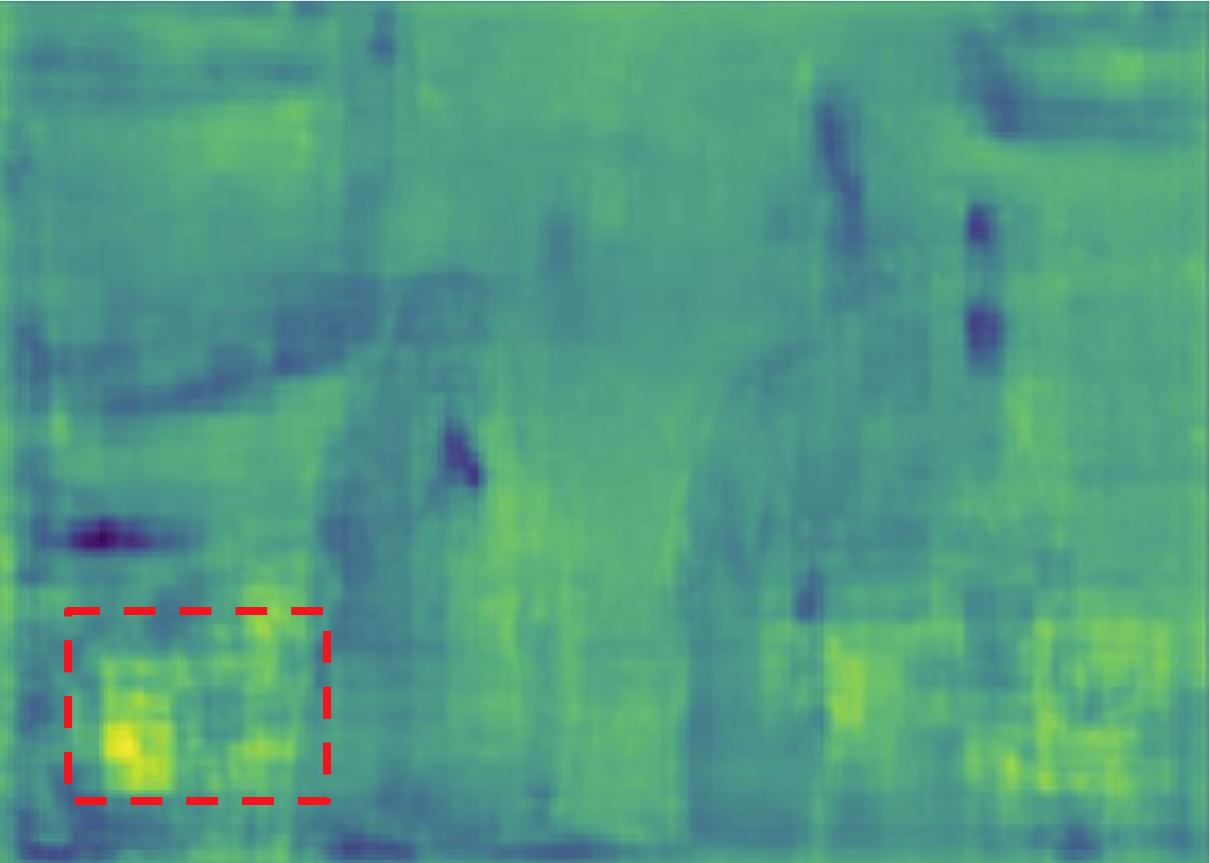}
        \captionsetup{font=small}
        \setlength\abovecaptionskip{-2ex}
        \caption*{$L_0$ only}
        \end{subfigure}
        \centering
        \begin{subfigure}{0.49\columnwidth}
          \centering
          \adjincludegraphics[width=\textwidth,trim={0 0 0 {.2\height}},clip]{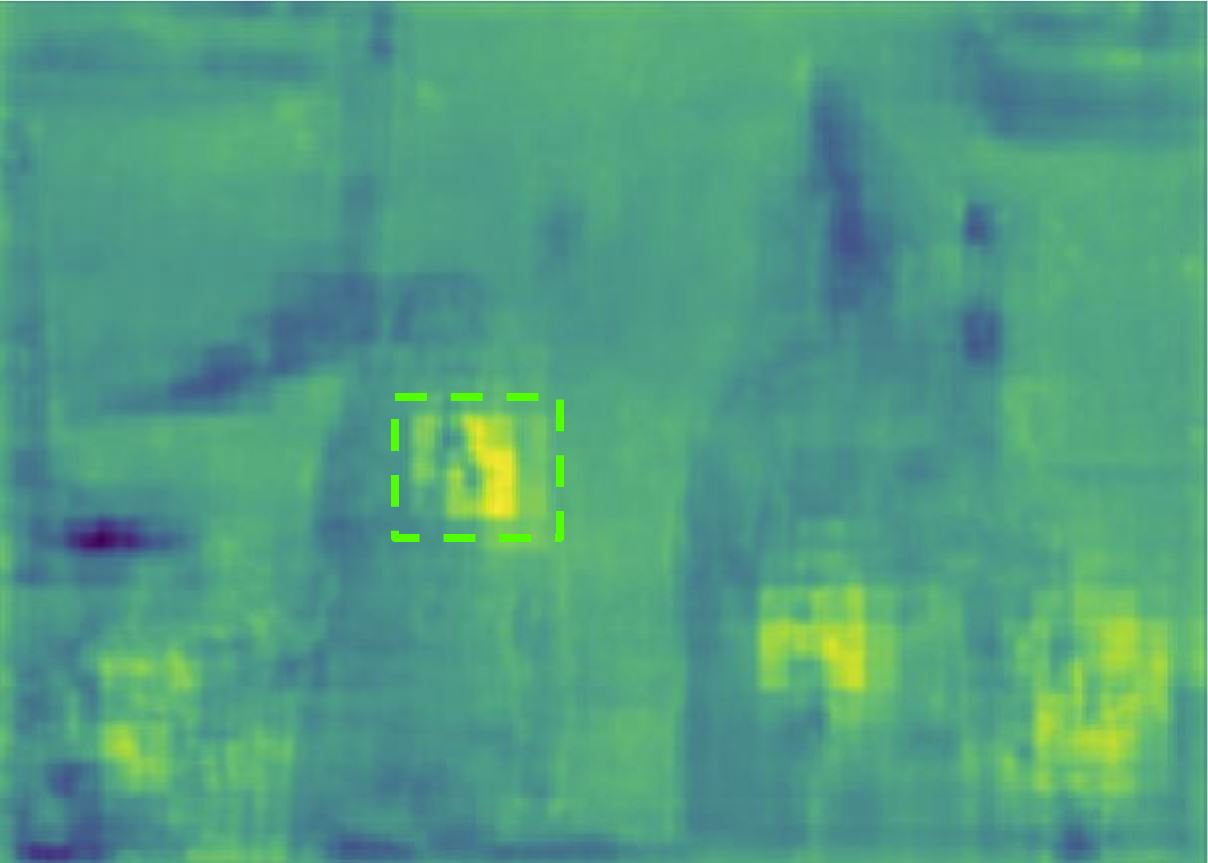}
        \captionsetup{font=small}
        \setlength\abovecaptionskip{-2ex}
        \caption*{with our losses}
        \end{subfigure}
    \captionsetup{font=small}
    \caption{Proximal Relationship Ambiguity}
    \label{fig:rel_ambiguity}
    \end{subfigure}   
  \end{minipage}
\captionsetup{font=small}
\caption{Example results of RelDN with $L_0$ only and with our losses. The top row shows RelDN outputs and the bottom row visualizes the learned predicate CNN features of the two models. Red and green boxes highlight the wrong and right outputs (the first row) or feature saliency (the second row). As it shows, our losses force the model to attend to the representative regions that discriminate the correct relationships against unrelated entity pairs, thus is able to disentangle entity instance confusion and proximal relationship ambiguity.}
\label{fig:vis}
\end{figure*}

We compare with other top models on the official evaluation server. The official test set is split into a Public and Private set with a 30\%/70\% split. The Public set is used as a dev set. We present individual results for both, as well as their weighted average under Overall in Table~\ref{tab:oi}.

\noindent\textbf{Visual Genome:} We follow the same train/val splits and evaluation metrics as \cite{zellers2018neural}. We train our entity detector initialized by COCO pre-trained weights. Following \cite{zellers2018neural}, we conduct three evaluations: scene graph detection(SGDET), scene graph classification (SGCLS), and predicate classification (PRDCLS). We report results for these tasks with and without the Graphical Contrastive Losses.


\noindent\textbf{VRD:} We evaluate our model with entity detectors initialized by ImageNet and COCO pre-trained weights. We use the same evaluation metrics as in \cite{yu17iccv}, which reports R@50 and R@100 for relationship predictions at 1, 10, and 70 predicates per entity pair.

\subsection{Loss Analysis}

\noindent\textbf{Loss Combinations:} We now look at whether our proposed losses reduce two aforementioned errors without affecting the overall performance, and whether all three losses are necessary. Results in Table \ref{tab:abl_L} show that combination of all the three losses with the N-way cross-entropy loss ($L_0+L_1+L_2+L_3$) has consistently superior performance over just $L_0$. Notably, AP$_{rel}$ on ``holds'' improves by from 41.84 to 43.09 (+1.3). It improves even more significantly from 36.04 to 41.04 (+5.0) on ``plays" and from 40.43 to 44.16 (+3.7) on ``interacts\_with" respectively. These three classes suffer the most from the two aforementioned problems. Our results also show that any subset of the losses is worse than the entire ensemble. We see that $L_0+L_1$, $L_0+L_2$ and $L_0+L_3$ are inferior to $L_0+L_1+L_2+L_3$, especially on ``holds'', ``plays'', and ``interacts\_with'', where the largest margin is 3.87 ($L_0+L_2$ \vs $L_0+L_1+L_2+L_3$ on ``play'').

\begin{table*}[t!]
\centering
\resizebox{1.9\columnwidth}{!}{
\begin{tabular}{l c c c | c c c | c c c | c c | c c | c c}
\hline
& \multicolumn{9}{c}{Graph Constraint} & \multicolumn{6}{c}{No Graph Constraint} \\
& \multicolumn{3}{c}{SGDET} & \multicolumn{3}{c}{SGCLS} & \multicolumn{3}{c}{PRDCLS} & \multicolumn{2}{c}{SGDET} & \multicolumn{2}{c}{SGCLS} & \multicolumn{2}{c}{PRDCLS} \\
Recall at & 20 & 50 & 100 & 20 & 50 & 100 & 20 & 50 & 100 & 50 & 100 & 50 & 100 & 50 & 100 \\
\hline
VRD\cite{lu2016visual} & - & 0.3 & 0.5 & - & 11.8 & 14.1 & - & 27.9 & 35.0 & - & - & - & - & - & - \\
Associative Embedding\cite{newell2017pixels} & 6.5 & 8.1 & 8.2 & 18.2 & 21.8 & 22.6 & 47.9 & 54.1 & 55.4 & 9.7 & 11.3 & 26.5 & 30.0 & 68.0 & 75.2 \\
Message Passing\cite{xu2017scenegraph} & - & 3.4 & 4.2 & - & 21.7 & 24.4 & - & 44.8 & 53.0 & - & - & - & - & - & - \\
Message Passing+\cite{zellers2018neural} & 14.6 & 20.7 & 24.5 & 31.7 & 34.6 & 35.4 & 52.7 & 59.3 & 61.3 & 22.0 & 27.4 & 43.4 & 47.2 & 75.2 & 83.6 \\
Frequency\cite{zellers2018neural} & 17.7 & 23.5 & 27.6 & 27.7 & 32.4 & 34.0 & 49.4 & 59.9 & 64.1 & 25.3 & 30.9 & 40.5 & 43.7 & 71.3 & 81.2 \\
Frequency+Overlap\cite{zellers2018neural} & 20.1 & 26.2 & 30.1 & 29.3 & 32.3 & 32.9 & 53.6 & 60.6 & 62.2 & 28.6 & 34.4 & 39.0 & 43.4 & 75.7 & 82.9 \\
MotifNet-NOCONTEXT\cite{zellers2018neural} & 21.0 & 26.2 & 29.0 & 31.9 & 34.8 & 35.5 & 57.0 & 63.7 & 65.6 & 29.8 & 34.7 & 43.4 & 46.6 & 78.8 & 85.9 \\
MotifNet-LeftRight\cite{zellers2018neural} & \textbf{21.4} & 27.2 & 30.3 & 32.9 & 35.8 & 36.5 & 58.5 & 65.2 & 67.1 & \textbf{30.5} & 35.8 & 44.5 & 47.7 & 81.1 & 88.3 \\
\hline
\textbf{RelDN, $L_0$ only} & 20.8 & 28.1 & 32.5 & \textbf{36.1} & 36.7 & 36.7 & 66.7 & 68.3 & 68.3 & 30.1 & 36.4 & \textbf{48.9} & \textbf{50.8} & 93.7 & 97.7 \\
\textbf{RelDN} & 21.1 & \textbf{28.3} & \textbf{32.7} & \textbf{36.1} & \textbf{36.8} & \textbf{36.8} & \textbf{66.9} & \textbf{68.4} & \textbf{68.4} & 30.4 & \textbf{36.7} & \textbf{48.9} & \textbf{50.8} & \textbf{93.8} & \textbf{97.8} \\
\hline
RelDN (X-101-FPN) & 22.5 & 31.0 & 36.7 & 38.2 & 38.9 & 38.9 & 67.2 & 68.7 & 68.8 & 32.6 & 40.0 & 51.7 & 53.6 & 94.0 & 97.8 \\
\hline
\end{tabular}
}
\setlength{\abovecaptionskip}{-5pt}
\captionsetup{font=small}
\caption{\rule{0pt}{15pt} Comparison with state-of-the-arts on VG. \textbf{$L_0$ only} is the RelDN without our losses. We also include results of our model with ResNeXt-101-FPN as the backbone for future work reference.}
\label{tab:vg}
\end{table*}


\begin{table*}[t!]
\resizebox{2.1\columnwidth}{!}{
\begin{tabular}{l c c c c | c c c c c c | c c c c c c}
\hline
& \multicolumn{2}{c}{Relationship} & \multicolumn{2}{c}{Phrase} & \multicolumn{6}{c}{Relationship Detection} & \multicolumn{6}{c}{Phrase Detection} \\
& \multicolumn{4}{c}{free k} & \multicolumn{2}{c}{k = 1} & \multicolumn{2}{c}{k = 10}  & \multicolumn{2}{c}{k = 70}  & \multicolumn{2}{c}{k = 1} & \multicolumn{2}{c}{k = 10}  & \multicolumn{2}{c}{k = 70} \\
Recall at & 50 & 100 & 50 & 100 & 50 & 100 & 50 & 100 & 50 & 100 & 50 & 100 & 50 & 100 & 50 & 100 \\
\hline
PPRFCN*\cite{zhang2017ppr} & 14.41 & 15.72 & 19.62 & 23.75  & - & - & - & - & - & - & - & - & - & - & - & - \\
VTransE* & 14.07 & 15.20 & 19.42 & 22.42  & - & - & - & - & - & - & - & - & - & - & - & - \\
SA-Full*\cite{Peyre17} & 15.80 & 17.10 & 17.90 & 19.50  & - & - & - & - & - & - & - & - & - & - & - & - \\
DR-Net*\cite{dai2017detecting} & 17.73 & 20.88 & 19.93 & 23.45 & - & - & - & - & - & - & - & - & - & - & - & - \\
ViP-CNN\cite{LiCVPR2017} & 17.32 & 20.01 &  22.78 & 27.91 & 17.32 & 20.01 & - & - & - & - & 22.78 & 27.91 & - & - & - & - \\
VRL\cite{liang2017deep} & 18.19 & 20.79 & 21.37 & 22.60  & 18.19 & 20.79 & - & - & - & - & 21.37 & 22.60 & - & - & - & - \\
CAI*\cite{Zhuang_2017_ICCV} & 20.14 & 23.39 & 23.88 & 25.26 & - & - & - & - & - & - & - & - & - & - & - & - \\
KL distilation\cite{yu17iccv} & 22.68 & 31.89 & 26.47 & 29.76 & 19.17 & 21.34 & 22.56 & 29.89 & 22.68 & 31.89 & 23.14 & 24.03 & 26.47 & 29.76 & 26.32 & 29.43 \\
Zoom-Net\cite{Yin_2018_ECCV} & 21.37 & 27.30 & 29.05 & 37.34 & 18.92 & 21.41 & - & - & 21.37 & 27.30 & 24.82 & 28.09 & - & - & 29.05 & 37.34 \\
CAI + SCA-M\cite{Yin_2018_ECCV} & 22.34 & 28.52 & 29.64 & 38.39 & 19.54 & 22.39 & - & - & 22.34 & 28.52 & 25.21 & 28.89 & - & - & 29.64 & 38.39 \\
\hline
\textbf{RelDN, $L_0$ only} (ImageNet) & 21.62 & 26.12 & 28.59 & 35.18 & 19.57 & 22.61 & 21.62 & 26.12 & 21.62 & 26.12 & 26.39 & 31.28 & 28.59 & 35.18 & 28.59 & 35.18 \\
\textbf{RelDN} (ImageNet) & 21.52 & 26.38 & 28.24 & 35.44 & 19.82 & 22.96 & 21.52 & 26.38 & 21.52 & 26.38 & 26.37 & 31.42 & 28.24 & 35.44 & 28.24 & 35.44 \\
\textbf{RelDN, $L_0$ only} (COCO) & 26.67 & 32.55 & 33.29 & 41.25 & 24.30 & 27.91 & 26.67 & 32.55 & 26.67 & 32.55 & 31.09 & \textbf{36.42} & 33.29 & 41.25 & 33.29 & 41.25 \\
\textbf{RelDN} (COCO) & \textbf{28.15} & \textbf{33.91} & \textbf{34.45} & \textbf{42.12} & \textbf{25.29} & \textbf{28.62} & \textbf{28.15} & \textbf{33.91} & \textbf{28.15} & \textbf{33.91} & \textbf{31.34} & \textbf{36.42} & \textbf{34.45} & \textbf{42.12} & \textbf{34.45} & \textbf{42.12} \\
\hline
\end{tabular}
}
\setlength{\abovecaptionskip}{-5pt}
\setlength\belowcaptionskip{-2ex}
\captionsetup{font=small}
\caption{\rule{0pt}{15pt} Comparison with state-of-the-art on VRD ($-$ means unavailable / unknown). Same with Table \ref{tab:vg}, \textbf{$L_0$ only} is the RelDN without our losses. ``Free k'' means considering $k$ as a hyper-parameter that can be cross-validated.}
\label{tab:vrd}
\end{table*}

\begin{table}[t!]
\centering
\resizebox{0.6\columnwidth}{!}{
\begin{tabular}{c | c c | c}
\hline
Team ID & Public & Private & Overall \\
\hline
radek & 0.289 & 0.201 & 0.227 \\
toshif & 0.256 & 0.228 & 0.237 \\
tito & 0.256 & 0.237 & 0.243 \\
Kyle & 0.280 & 0.235 & 0.249 \\
Seiji & 0.332 & 0.285 & 0.299 \\
\hline
\textbf{RelDN$^*$} & 0.327 & 0.299 & 0.308 \\
\textbf{RelDN} & 0.320 & 0.332 & \textbf{0.328} \\
\hline
\end{tabular}
}
\setlength\belowcaptionskip{-2ex}
\captionsetup{font=small}
\caption{Comparison with models from OpenImages Challenge.
RelDN$^*$ means using the same entity detector from \textit{Seiji}, the champion model. Overall is computed as 0.3*Public+0.7*Private. Note that this table uses the official mAP$_{rel}$ and mAP$_{phr}$ metrics.}
\label{tab:oi}
\end{table}

\begin{figure*}[t!]
  \centering
  \begin{subfigure}{2.1\columnwidth}
    \centering
    \begin{subfigure}{0.3\columnwidth}
      \centering
      \includegraphics[width=\textwidth]{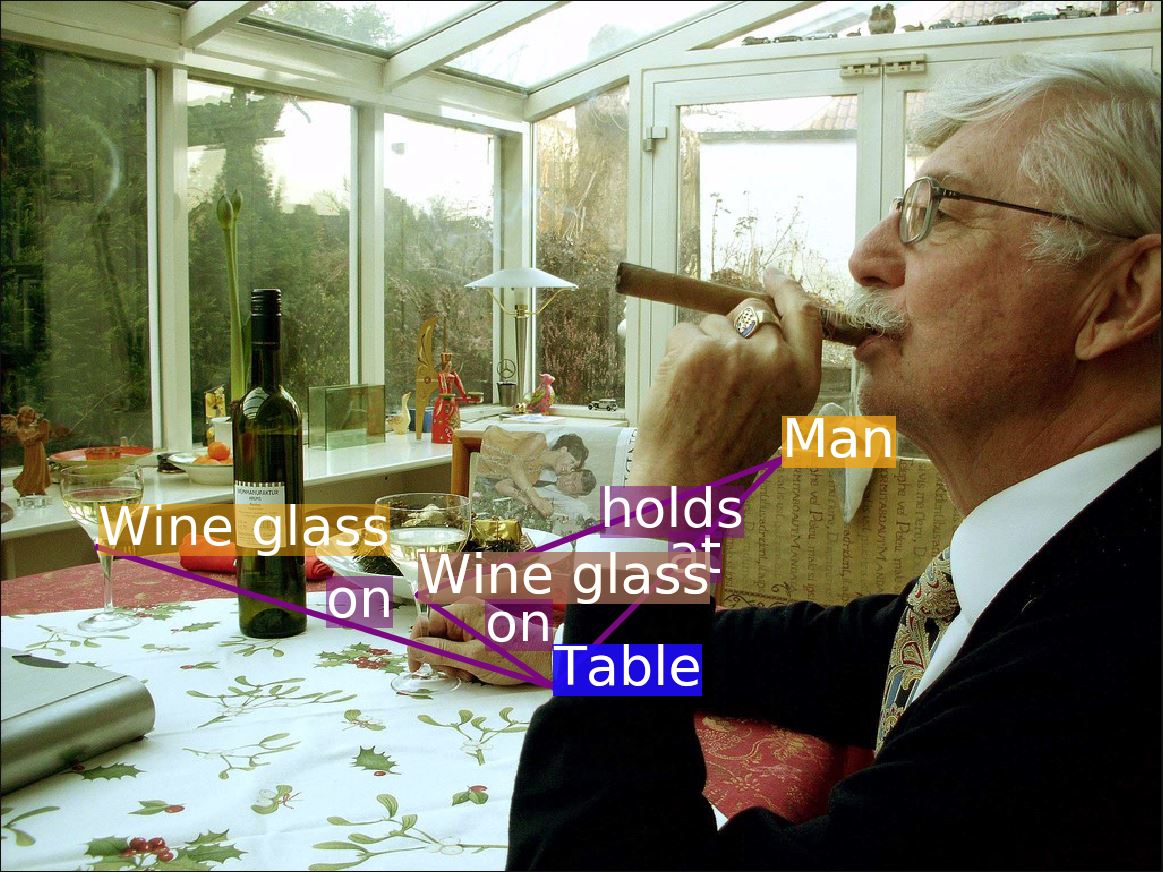}
    \end{subfigure}
    \centering
    \begin{subfigure}{0.335\columnwidth}
      \centering
      \includegraphics[width=\textwidth]{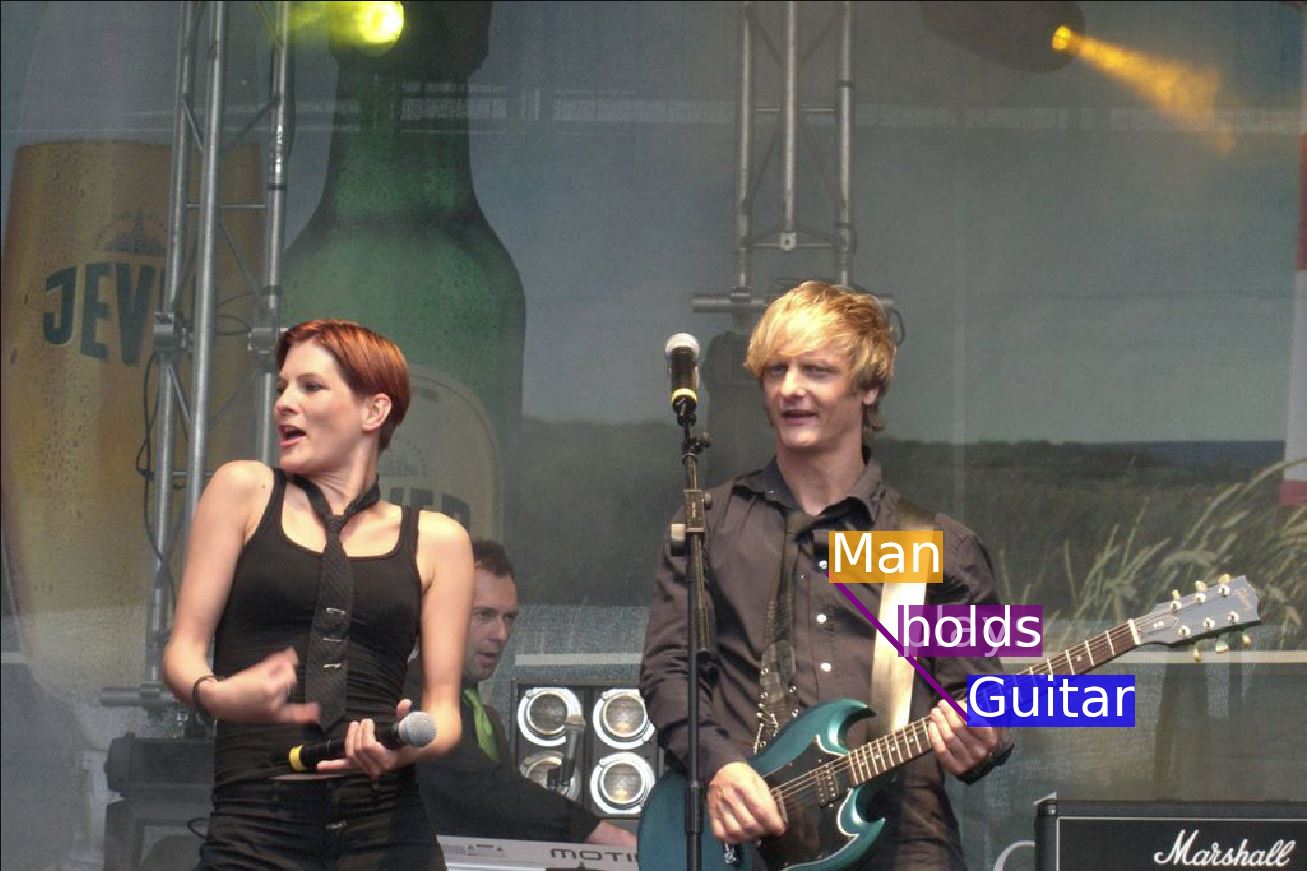}
    \end{subfigure}
    \centering
    \begin{subfigure}{0.355\columnwidth}
      \centering
      \includegraphics[width=\textwidth]{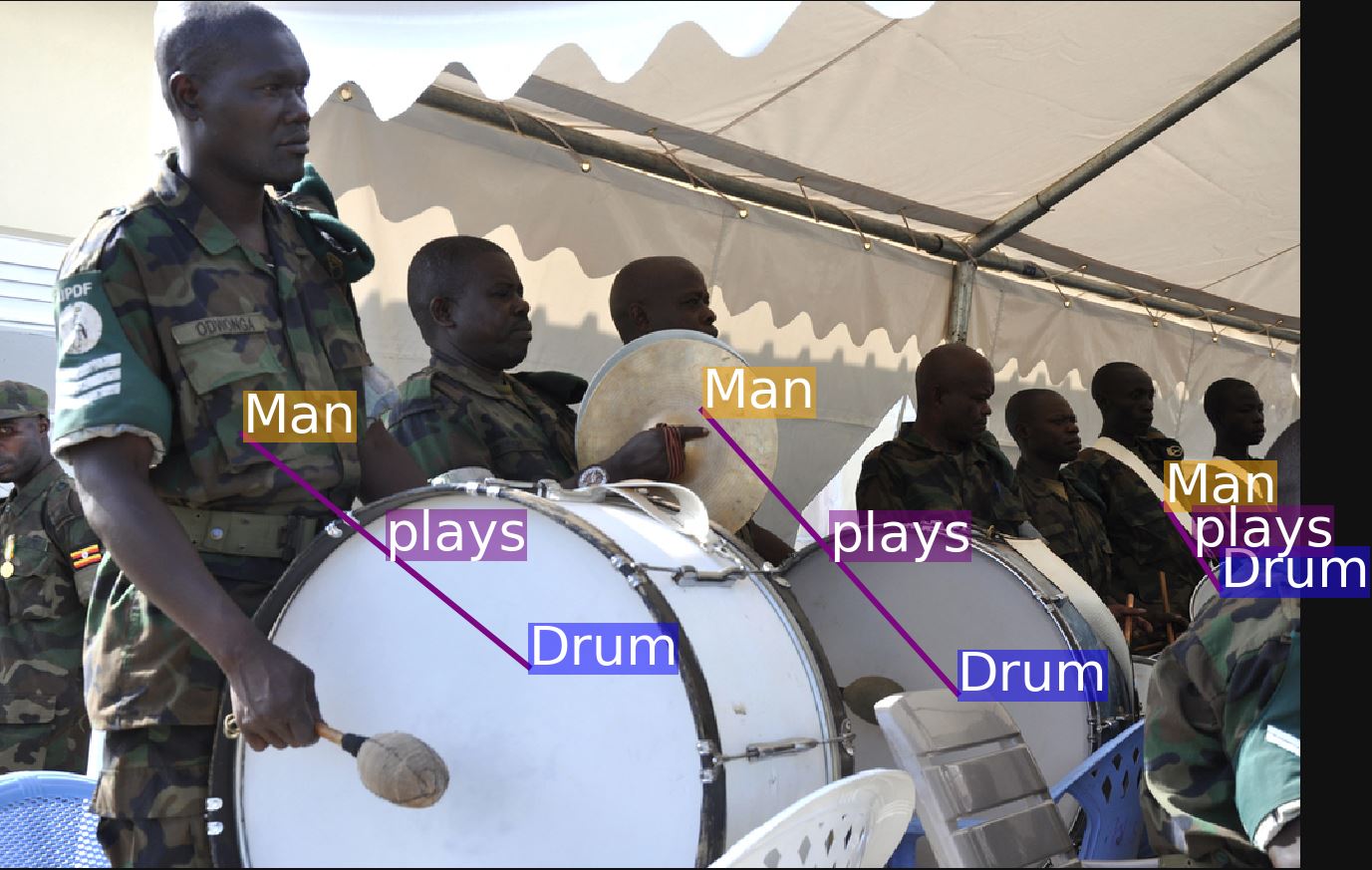}
    \end{subfigure}
  \end{subfigure}

  \centering
  \begin{subfigure}{2.1\columnwidth}
    \centering
    \begin{subfigure}{0.324\columnwidth}
      \centering
      \includegraphics[width=\textwidth]{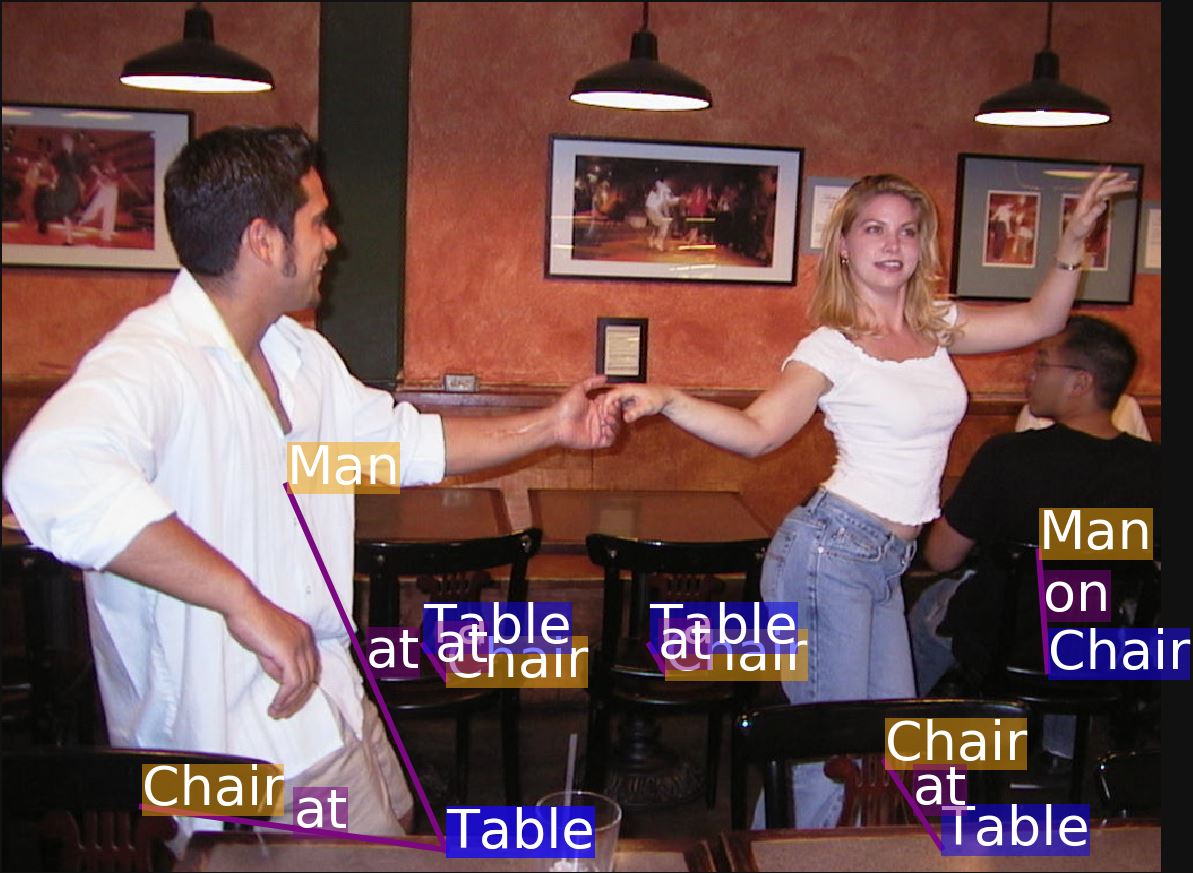}
    \end{subfigure}
    \centering
    \begin{subfigure}{0.325\columnwidth}
      \centering
      \includegraphics[width=\textwidth]{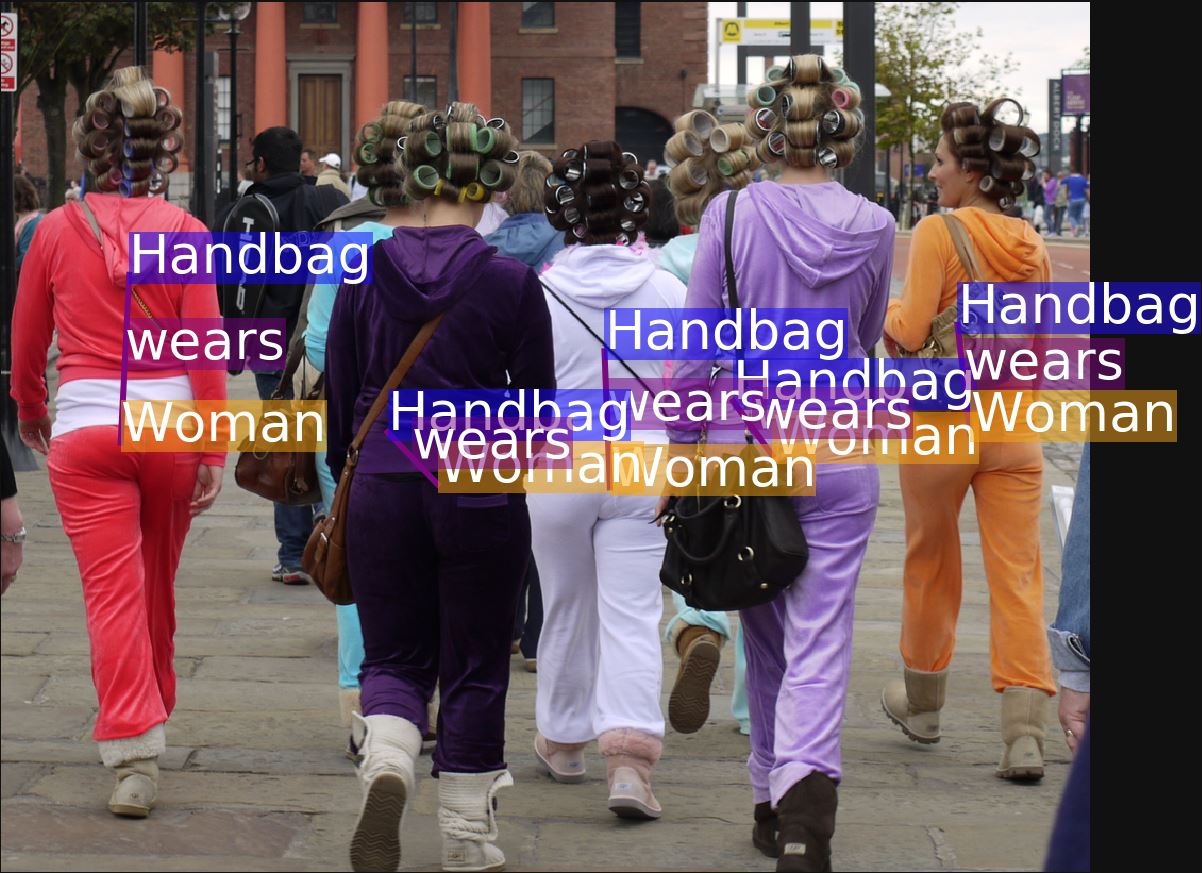}
    \end{subfigure}
    \centering
    \begin{subfigure}{0.341\columnwidth}
      \centering
      \includegraphics[width=\textwidth]{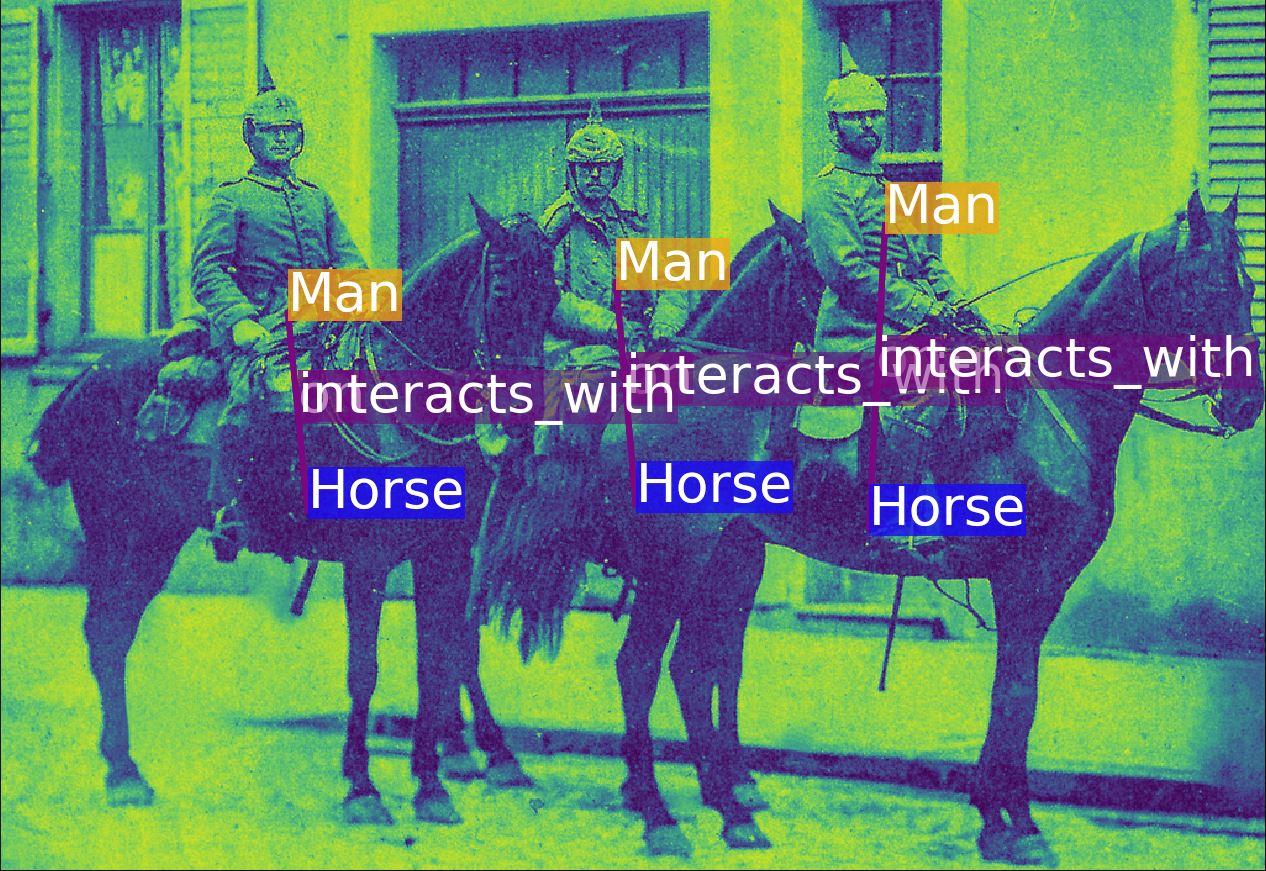}
    \end{subfigure}
  \end{subfigure}
  
  \centering
  \begin{subfigure}{2.1\columnwidth}
    \centering
    \begin{subfigure}{0.48025\columnwidth}
      \centering
      \adjincludegraphics[width=\textwidth,trim={0 {.088\height} 0 {.3\height}},clip]{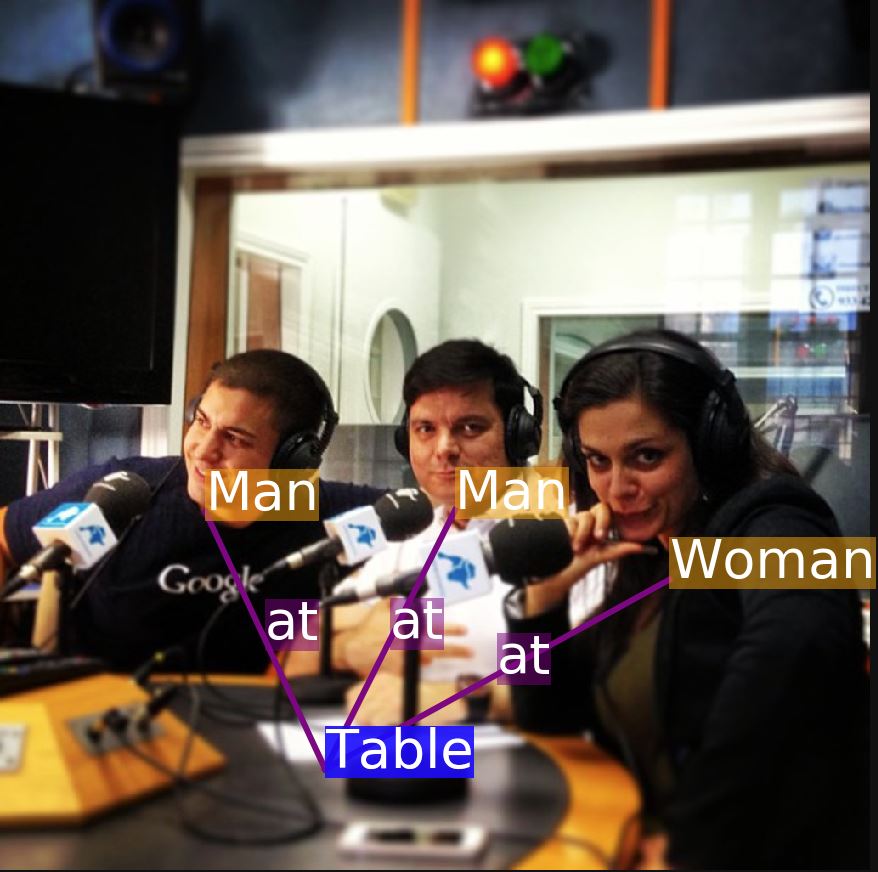} 
    \end{subfigure}
    \centering
    \begin{subfigure}{0.247\columnwidth}
      \centering
      \adjincludegraphics[width=\textwidth,trim={0 {.123\height} 0 0},clip]{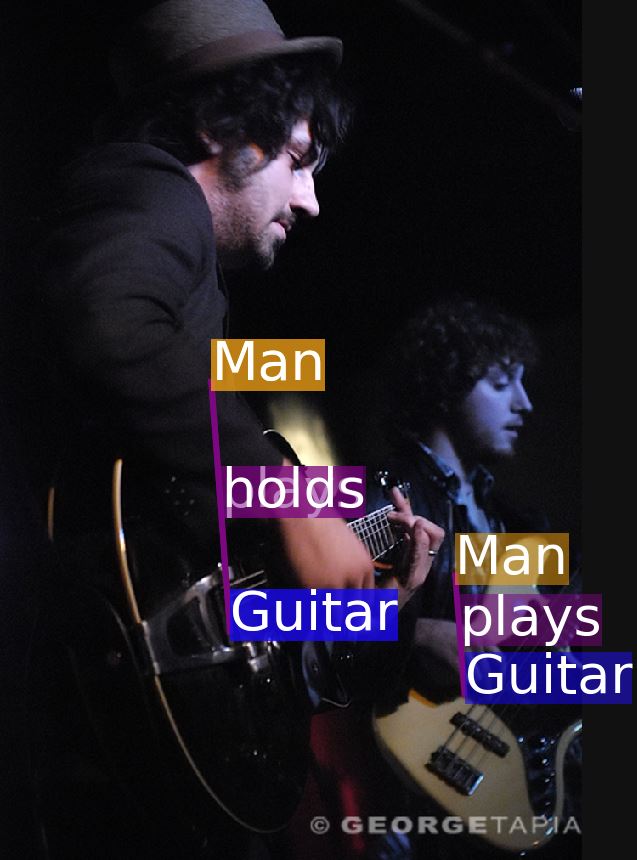}
    \end{subfigure}
    \centering
    \begin{subfigure}{0.26275\columnwidth}
      \centering
      \adjincludegraphics[width=\textwidth,trim={0 {.12\height} 0 0},clip]{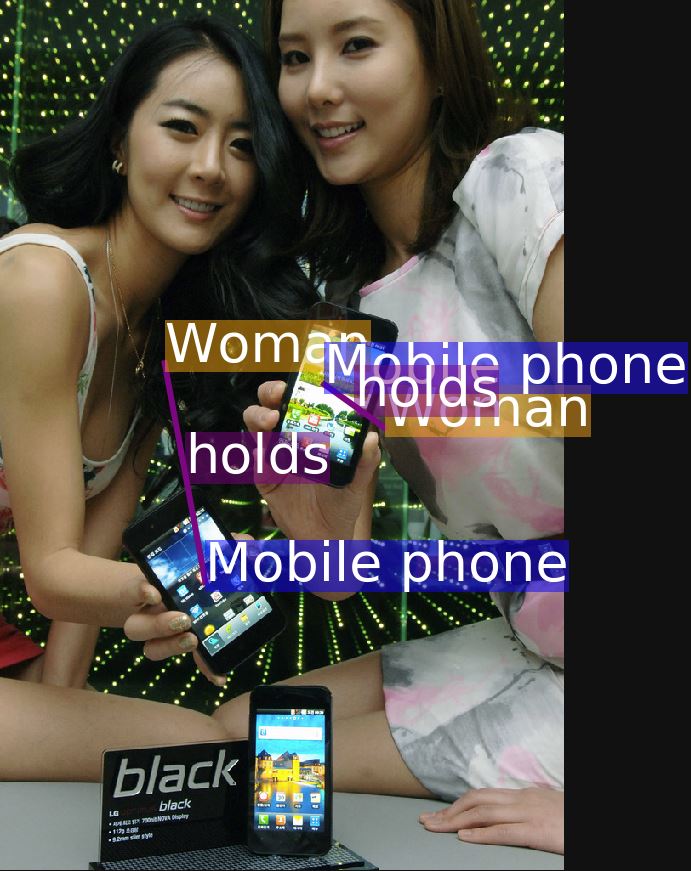} 
    \end{subfigure}
  \end{subfigure}
  
  \centering
  \begin{subfigure}{2.1\columnwidth}
    \centering
    \begin{subfigure}{0.3825\columnwidth}
      \centering
      \includegraphics[width=\textwidth]{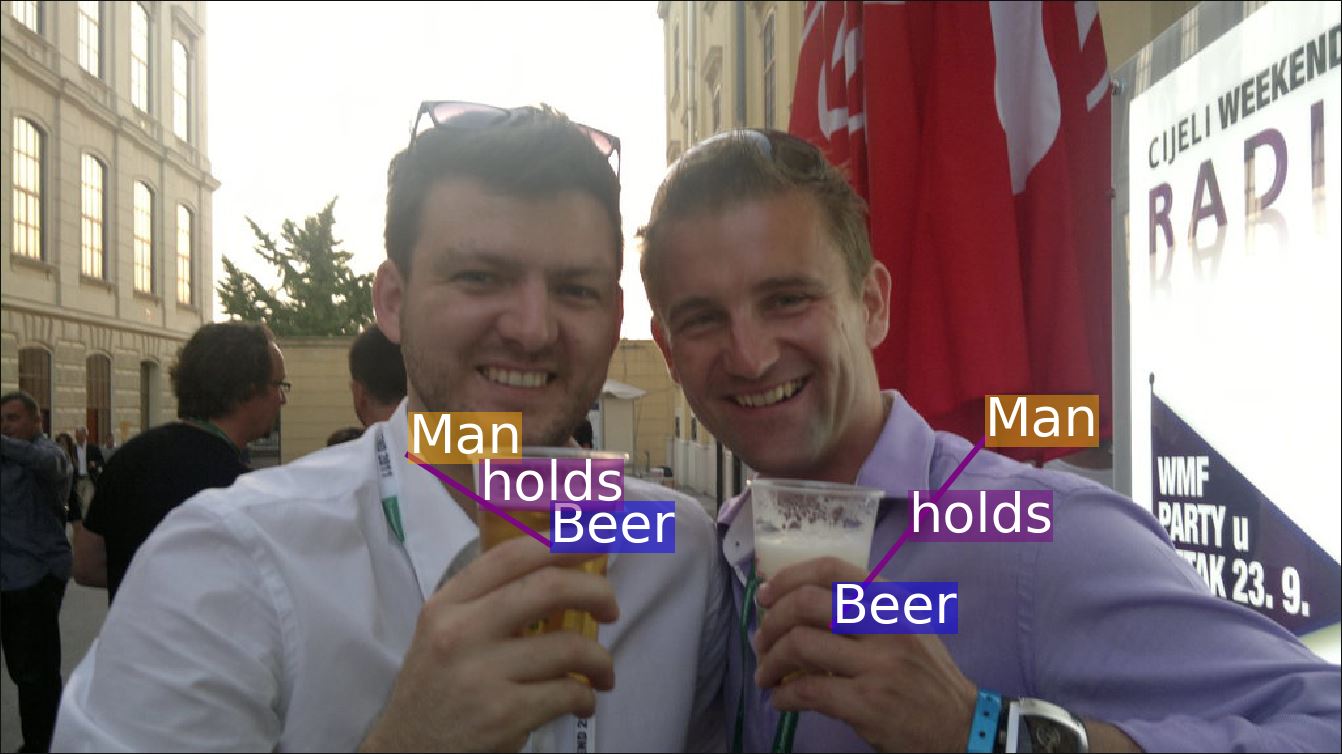} 
    \end{subfigure}
    \centering
    \begin{subfigure}{0.2865\columnwidth}
      \centering
      \includegraphics[width=\textwidth]{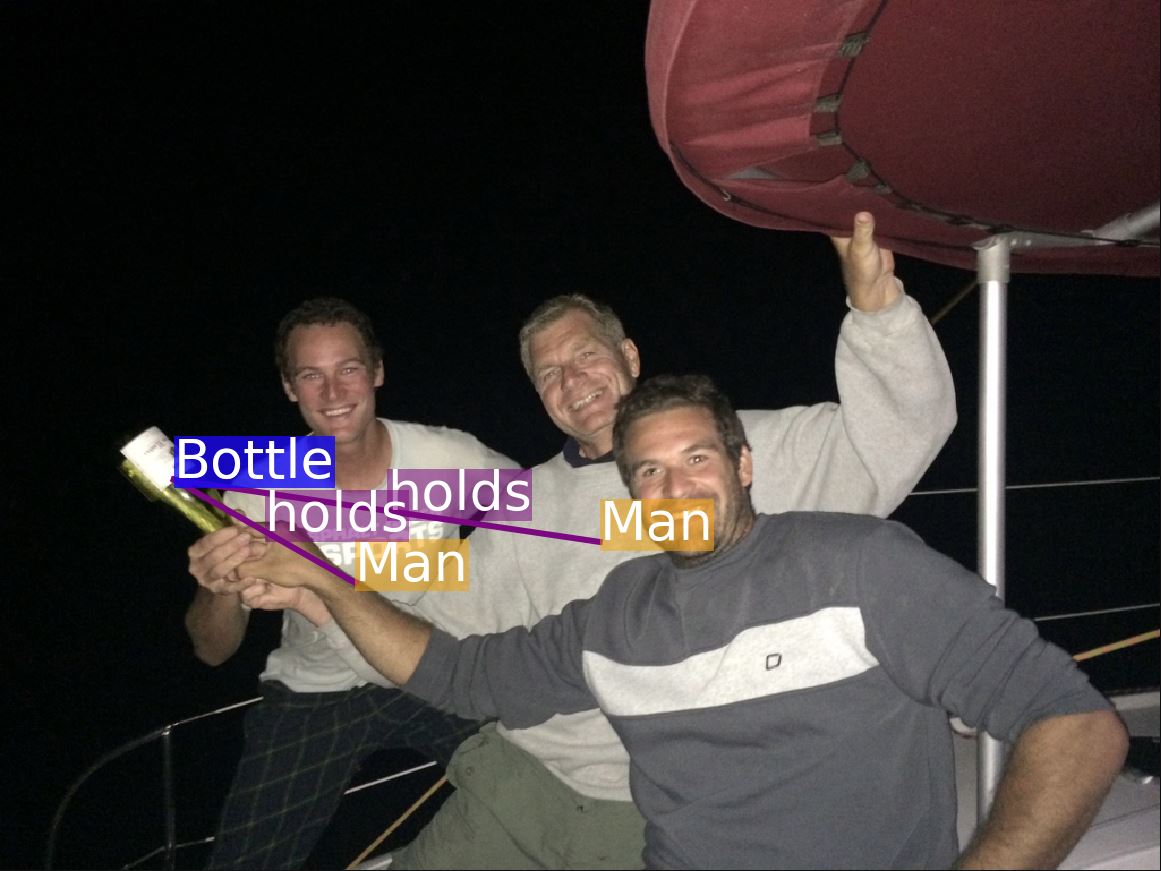}
    \end{subfigure}
    \centering
    \begin{subfigure}{0.3215\columnwidth}
      \centering
      \includegraphics[width=\textwidth]{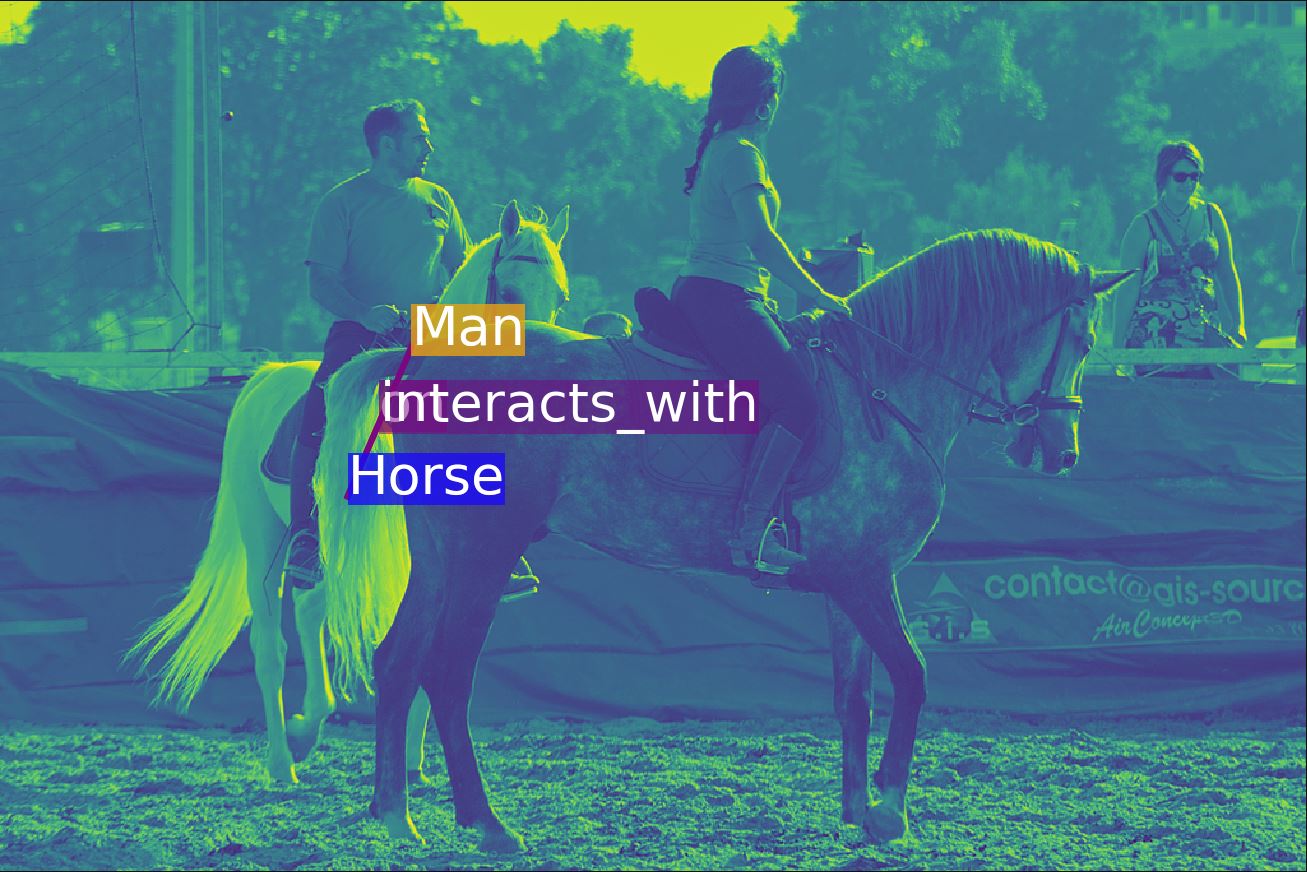} 
    \end{subfigure}
  \end{subfigure}
  
  \centering
  \begin{subfigure}{2.1\columnwidth}
    \centering
    \begin{subfigure}{0.358\columnwidth}
      \centering
      \includegraphics[width=\textwidth]{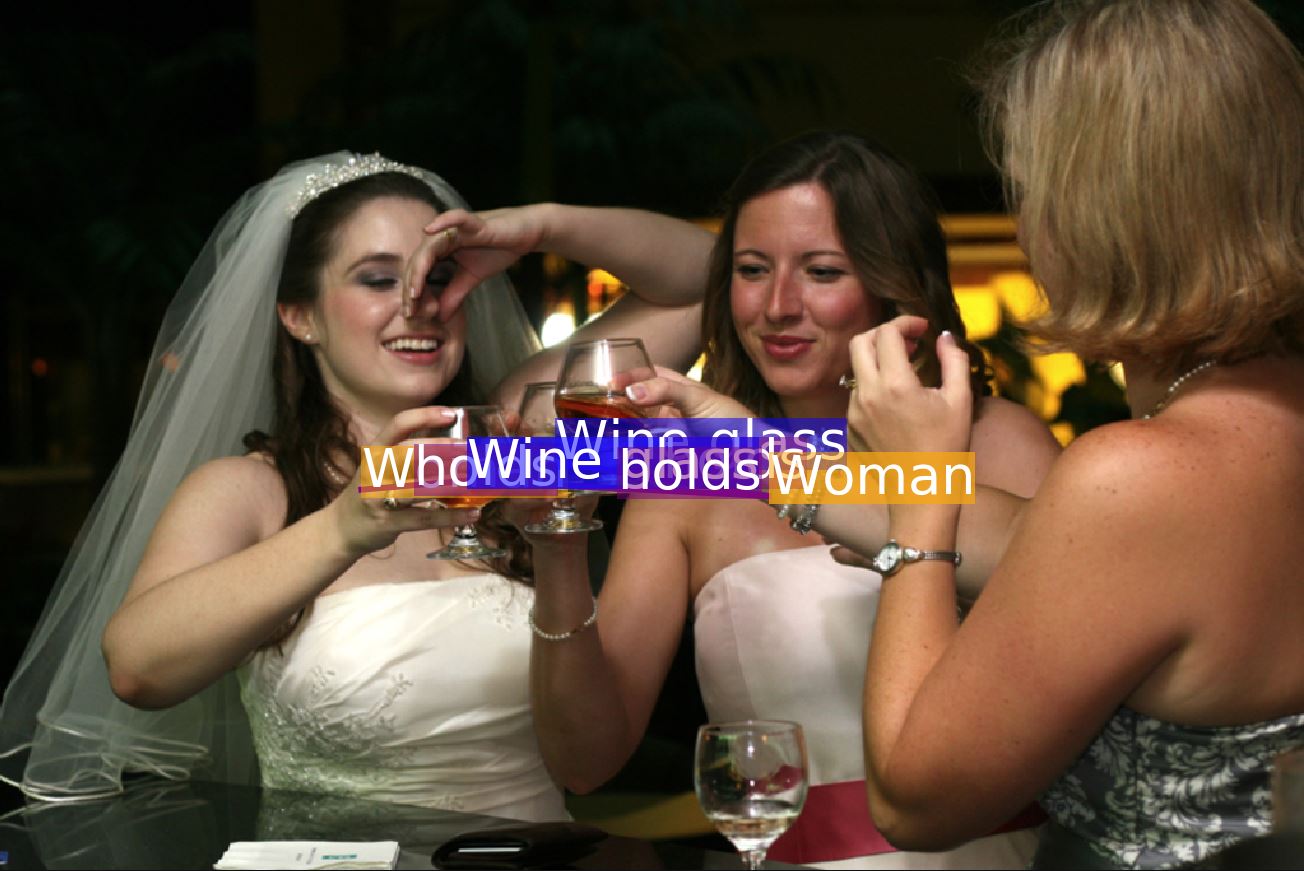} 
    \end{subfigure}
    \centering
    \begin{subfigure}{0.311\columnwidth}
      \centering
      \includegraphics[width=\textwidth]{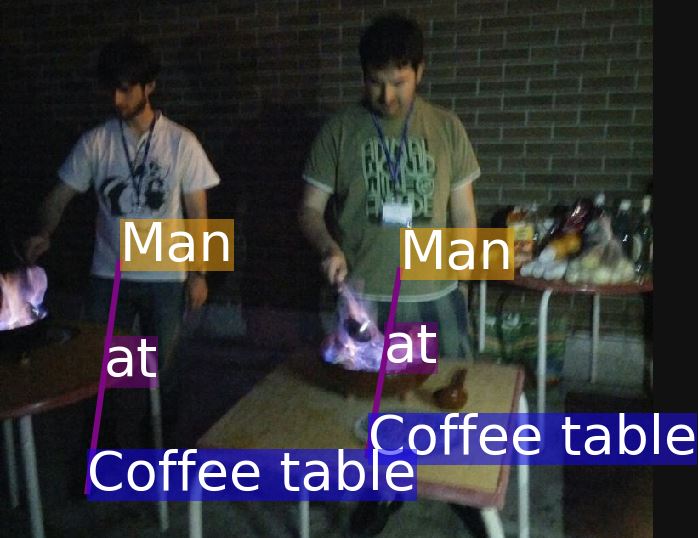}
    \end{subfigure}
    \centering
    \begin{subfigure}{0.321\columnwidth}
      \centering
      \includegraphics[width=\textwidth]{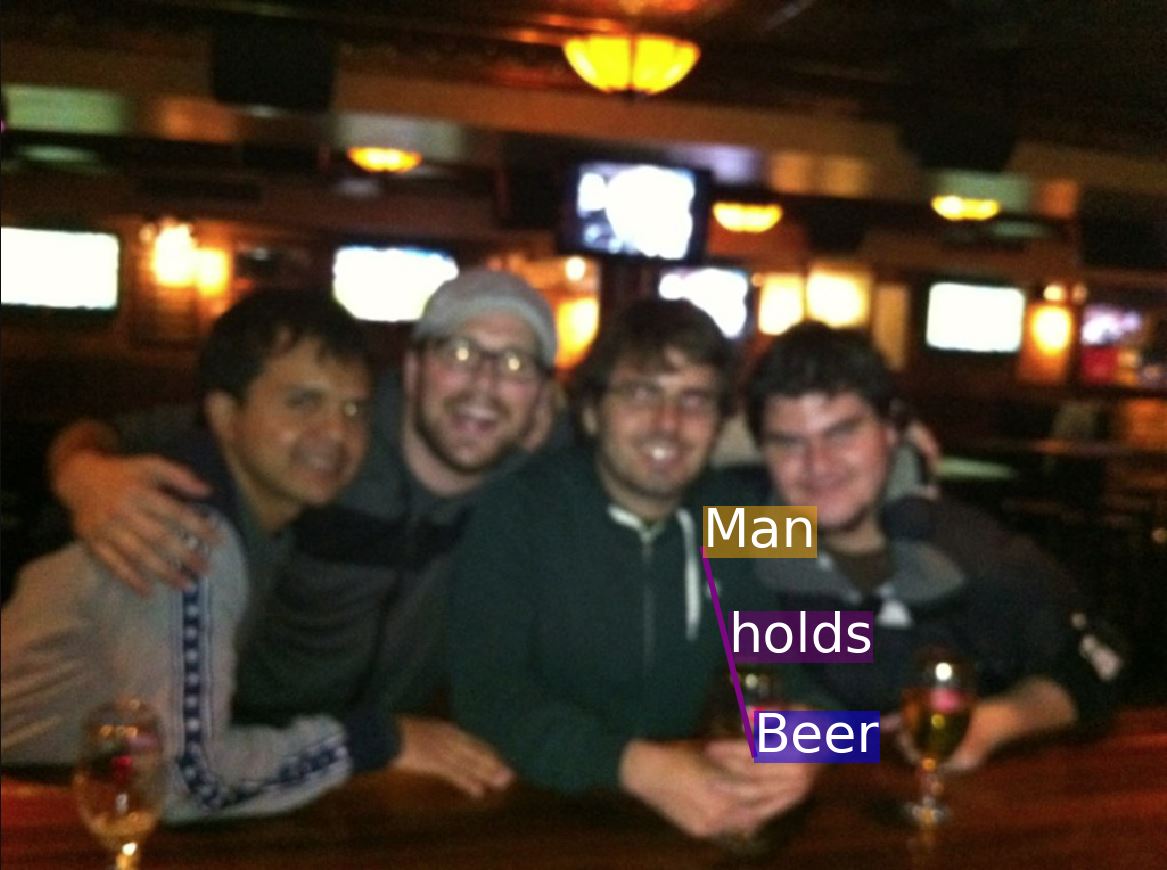} 
    \end{subfigure}
  \end{subfigure}
\setlength\belowcaptionskip{-2ex}
\captionsetup{font=small}
\caption{Example images of the 100 image subset with ground truth relationships. The subset contains five predicates where the Entity Instance Confusion and Proximal Relationship Ambiguity commonly occur.}
\label{fig:examples_special}
\end{figure*}

To better verify the isolated impact of our losses, we carefully sample a subset of 100 images containing five predicates that significantly suffer from the two aforementioned problems, selected via visual inspection on a random set of images. The five predicates are ``at'', ``holds'', ``plays'', ``interacts\_with'', and ``wears''. We sample them by looking at the raw images and select those with either entity instance confusion or proximal relationship ambiguity. Example images can be found in Figure \ref{fig:examples_special}. Table \ref{tab:abl_special_img} shows comparison of our losses with $L_0$ only on this subset. The overall gap is 1.4 and the largest gap is 4.1 at AP$_{rel}$ on ``holds''.

Figure \ref{fig:vis} shows two examples from this subset, one containing entity instance confusion and the other containing proximal relationship ambiguity. In Figure \ref{fig:entity_confusion} the model with only $L_0$ fails to identify the wine glass being held, while by adding our losses, the area surrounding the correct wine glass lights up. In Figure \ref{fig:rel_ambiguity} $\langle woman, plays, drum \rangle$ is incorrectly predicted since the $L_0$-only model mistakenly pairs the unplayed drum with the singer -- a reasonable error considering the amount of person-play-drum examples as well as the relative proximities between the singer and the drum. Our losses successfully suppress that region and attend to the correct microphone being held, demonstrating the effectiveness of our hard-negative sampling strategies.

\noindent\textbf{Margin Thresholds:} We study the effects of various values of the margin thresholds $\alpha_1,\alpha_2,\alpha_3$ used in Eq.\ref{eq:class_agnostic},\ref{eq:o_class_aware},\ref{eq:p_class_aware}. For each experiment, we set $\alpha_1=\alpha_2=\alpha_3=m$ while varying $m$. As shown in Table \ref{tab:abl_m}, we observe similar results with previous work \cite{kiros2014unifying,vendrov2015order} that $m=0.1$ or $m=0.2$ achieves the best performance. Note that $m=1.0$ is the largest possible margin, as our affinity scores range from 0 to 1. 


\subsection{Loss Analysis with the Official mAP metrics}

Here, we show our ablation studies using the official uniform-class-weighting evaluation metrics, \textit{mAP$_{rel}$}, \textit{mAP$_{phr}$} and \textit{score}. We also include \textit{mAP$_{rel}^\textbf{*}$}, \textit{mAP$_{phr}^\textbf{*}$} and \textit{score$^\textbf{*}$}, which is the standard mAP and score excluding ``under'' and ``hits'' in the evaluation.  Table \ref{tab:abl_L_no_weights} presents ablation study results on loss components. Table \ref{tab:abl_special_img_no_weights} shows comparison between the $L_0$-only model against the model with our losses on the 100 selected images. In Table \ref{tab:abl_L_no_weights} the variation of numbers using mAP and score demonstrates the necessity of de-emphasizing the extremely infrequent classes. Note that the mAP*-based columns show a similar trend to our wmAP-based results from the paper. In Table \ref{tab:abl_special_img_no_weights}, the model with our losses is still better than the $L_0$-only model by a non-trivial margin, mainly because the former outperform the latter on almost every per-class AP metric for those 5 selected classes. Note that since ``under'' and ``hits'' are not in the 100 image subset, there is no need to evaluate with \textit{mAP$_{rel}^\textbf{*}$}, \textit{mAP$_{phr}^\textbf{*}$} and \textit{score$^\textbf{*}$}.

\subsection{Model Analysis}
\label{subsec:model_analysis}

We conduct an effectiveness evaluation on the three modules of the RelDN. For the visual module, we also investigate the two skip-connections. As Table \ref{tab:abl_model} shows, the semantic module alone cannot solve relationship detection by using language bias only. By adding the basic visual feature, \ie, the $\langle$S,P,O$\rangle$ concatenation, we see a significant 4.7 gain, which is further improved by adding additional separate S,O skip-connections, especially at ``plays'' (+3.1), ``interacts\_with'' (+1.0), ``wears'' (+2.0) where subjects' or objects' appearance and poses are highly representative of the interactions. Finally, adding the spatial module gives the best results, and the most obvious gaps are at spatial relationships, \ie, ``at'' (+0.2), ``on'' (+0.2), ``inside\_of'' (+2.4).

\subsection{Comparison to State of the Art}

\noindent\textbf{OpenImages:} We present results compared with top 5 models from the Challenge in Table \ref{tab:oi}. We surpass the 1st place \textit{Seiji} by 4.7\% on Private set and 2.9\% on the full set, which is in fact a significant margin considering the low absolute scores and the large amount of test images (99,999 in total). Even using the same entity detector as \textit{Seiji}, we noticeable gaps (1.4\% and 0.8\%) on the two sets.


\noindent\textbf{Visual Genome:} Table \ref{tab:vg} shows that our model is better than state-of-the-arts on all metrics. It outperforms the previous best, MotifNet-LeftRight, by a 2.4\% gap on Scene Graph Detection (SGDET) with Recall@100 and by a 12.7\% gap on Predicate Classification (PRDCLS) with Recall@50. Note that although our entity detector is better than MotifNet-LeftRight on mAP at 50\% IoU (25.5 \vs 20.0), our implementation of Frequency+Overlap baseline (Recall@20: 16.2, Recall@50: 19.8, Recall@100: 21.5) is not better than their version (Recall@20: 21.0, Recall@50: 26.2, Recall@100: 30.1), indicating that our better relationship performance mostly comes from our model design.

We also observe that our losses achieve smaller gains over the standard cross-entropy loss setup than it does on OpenImages\_mini. The reasons are two-fold: 1) One of the few dominant relationship types in the Visual Genome dataset is possessive, \eg, ``ear of man'', which has much fewer entity confusion issues; 2) The $Recall@k$ metric is less strict than mAP. If there is an image with only one ground truth, then Recall@100 will always be 100\% as long as this ground truth target is within the top 100 model predictions, regardless of the ranking of the 100 outputs. As such, the small improvements in ranking the top 100 will not affect the score. Nevertheless, the improvements from our loss is still non-trivial and consistent on all metrics under different values of $k$.

In addition, we also show results using a better backbone, ResNeXt-101-FPN \cite{xie2017aggregated,lin2017feature}, for the entity detector in Table \ref{tab:vg}.

\noindent\textbf{VRD:} Table \ref{tab:vrd} presents results on VRD compared with state-of-the-art methods. Note that only \cite{Yin_2018_ECCV} specifically states that they use ImageNet pre-trained weights while others remain unknown. Therefore, we show results for pre-training on either ImageNet or COCO. Our model is competitive with those methods when pre-trained on ImageNet, but significantly outperforms when pre-trained on COCO. The gap between \textit{$L_0$ only} and the full model is smaller when pre-trained on ImageNet than on COCO. We believe the stronger localization features from pre-training on COCO is much easier for our model and losses to leverage.

\subsection{Qualitative Results}

\begin{figure*}[t!]
  \centering
  \begin{subfigure}{2.1\columnwidth}
    \centering
    \begin{subfigure}{0.32\columnwidth}
      \centering
      \includegraphics[width=\textwidth]{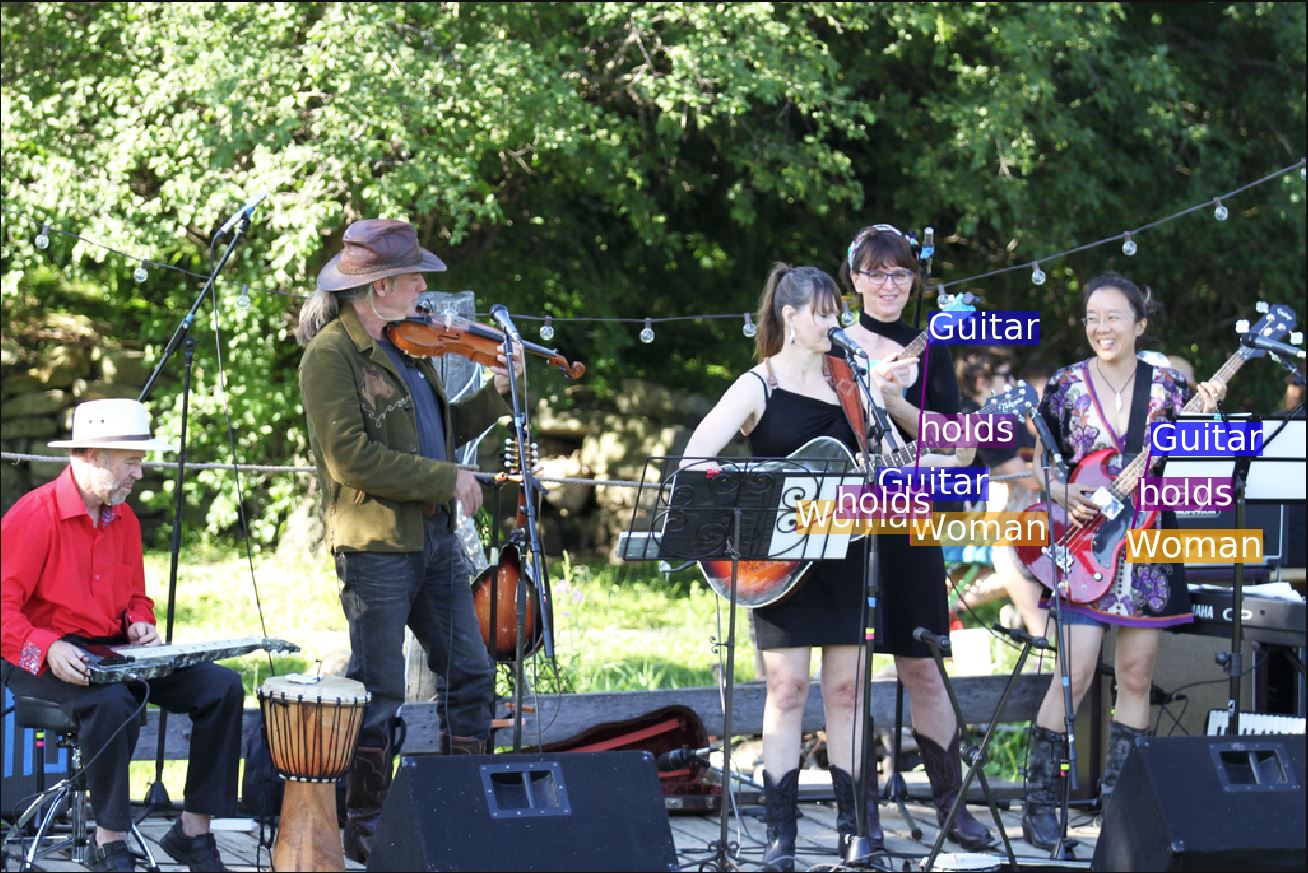}
    \end{subfigure}
    \centering
    \begin{subfigure}{0.32\columnwidth}
      \centering
      \includegraphics[width=\textwidth]{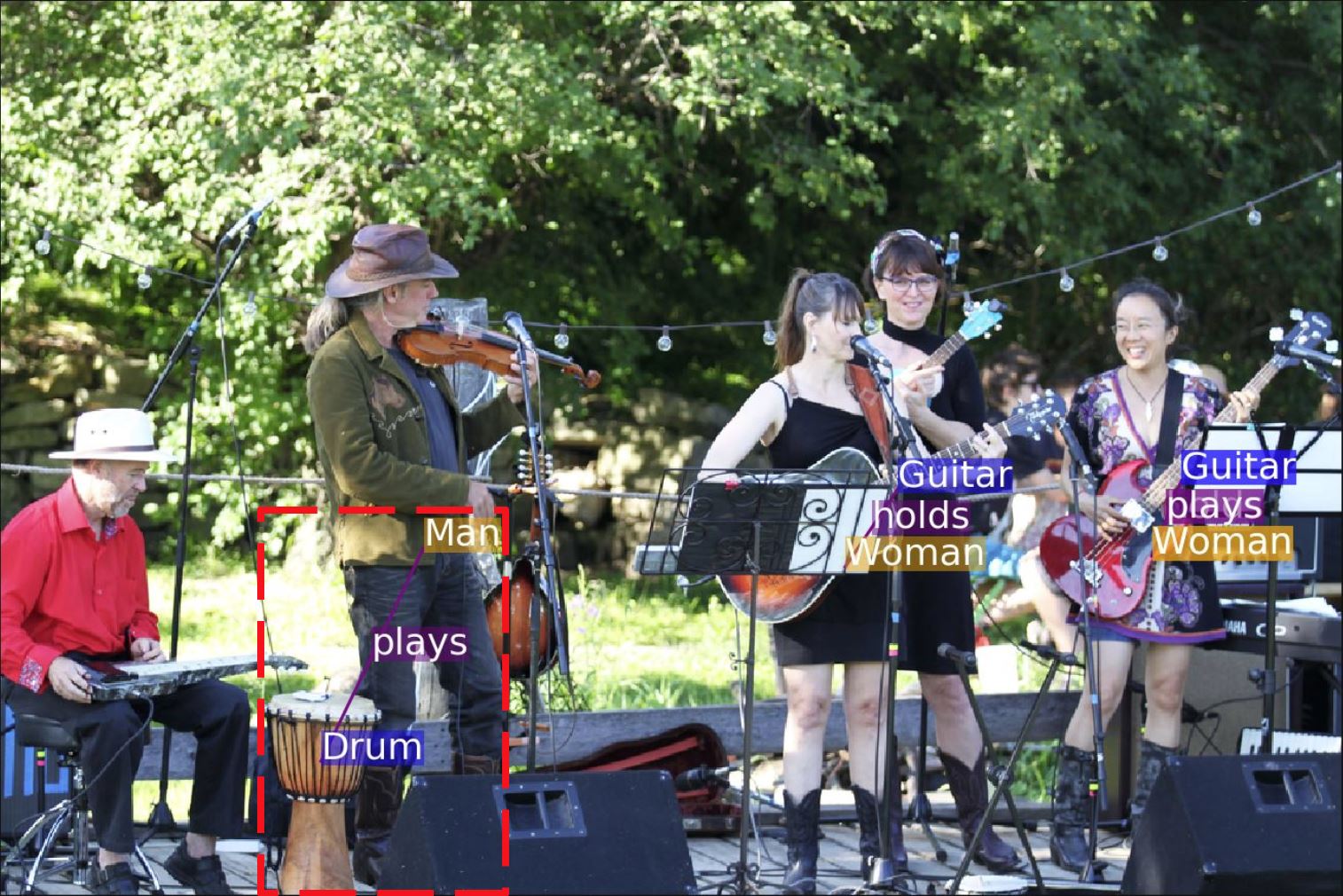}
    \end{subfigure}
    \centering
    \begin{subfigure}{0.32\columnwidth}
      \centering
      \includegraphics[width=\textwidth]{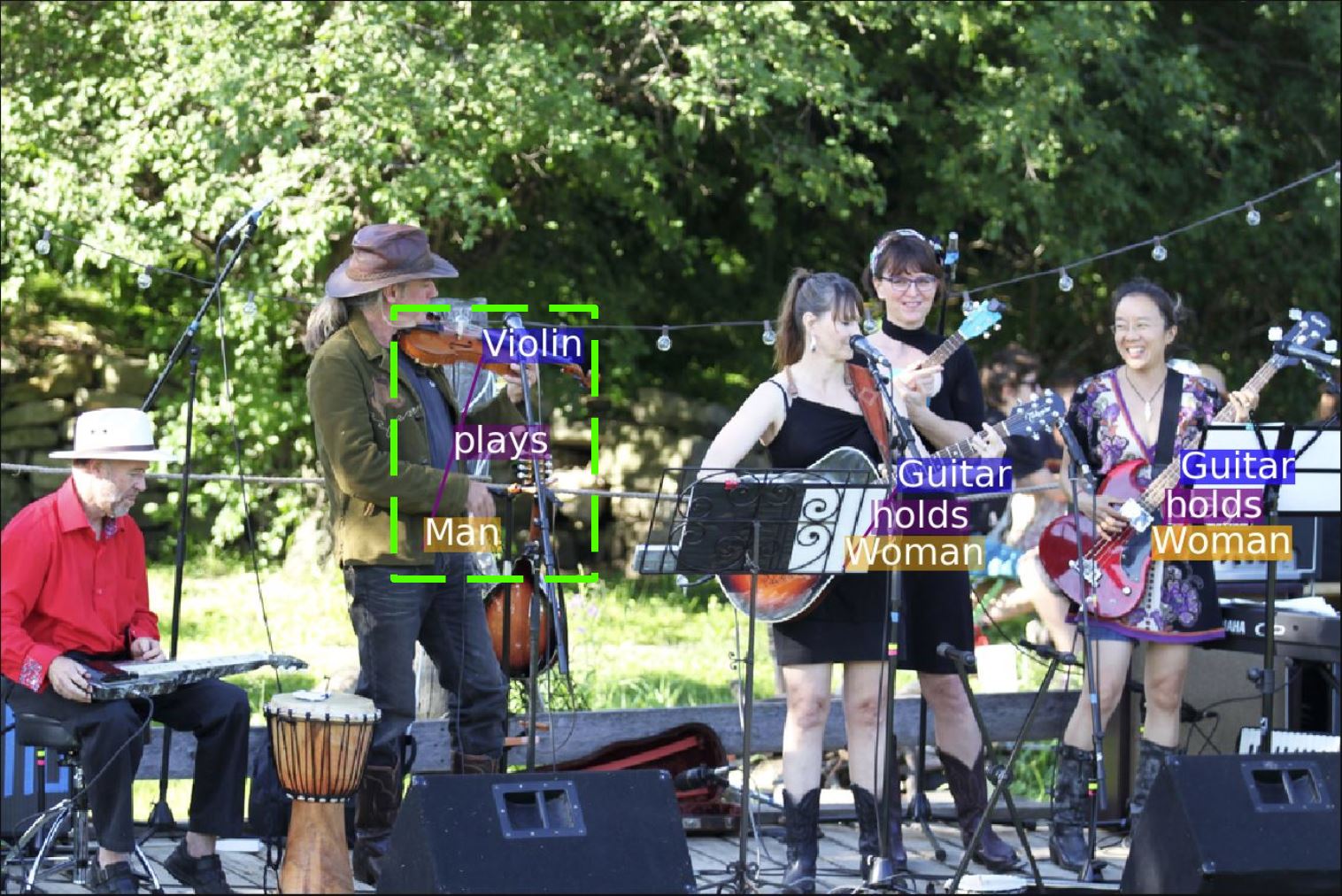}
    \end{subfigure}
  \end{subfigure}
  
  
  
  \centering
  \begin{subfigure}{2.1\columnwidth}
    \centering
    \begin{subfigure}{0.32\columnwidth}
      \centering
      \includegraphics[width=\textwidth]{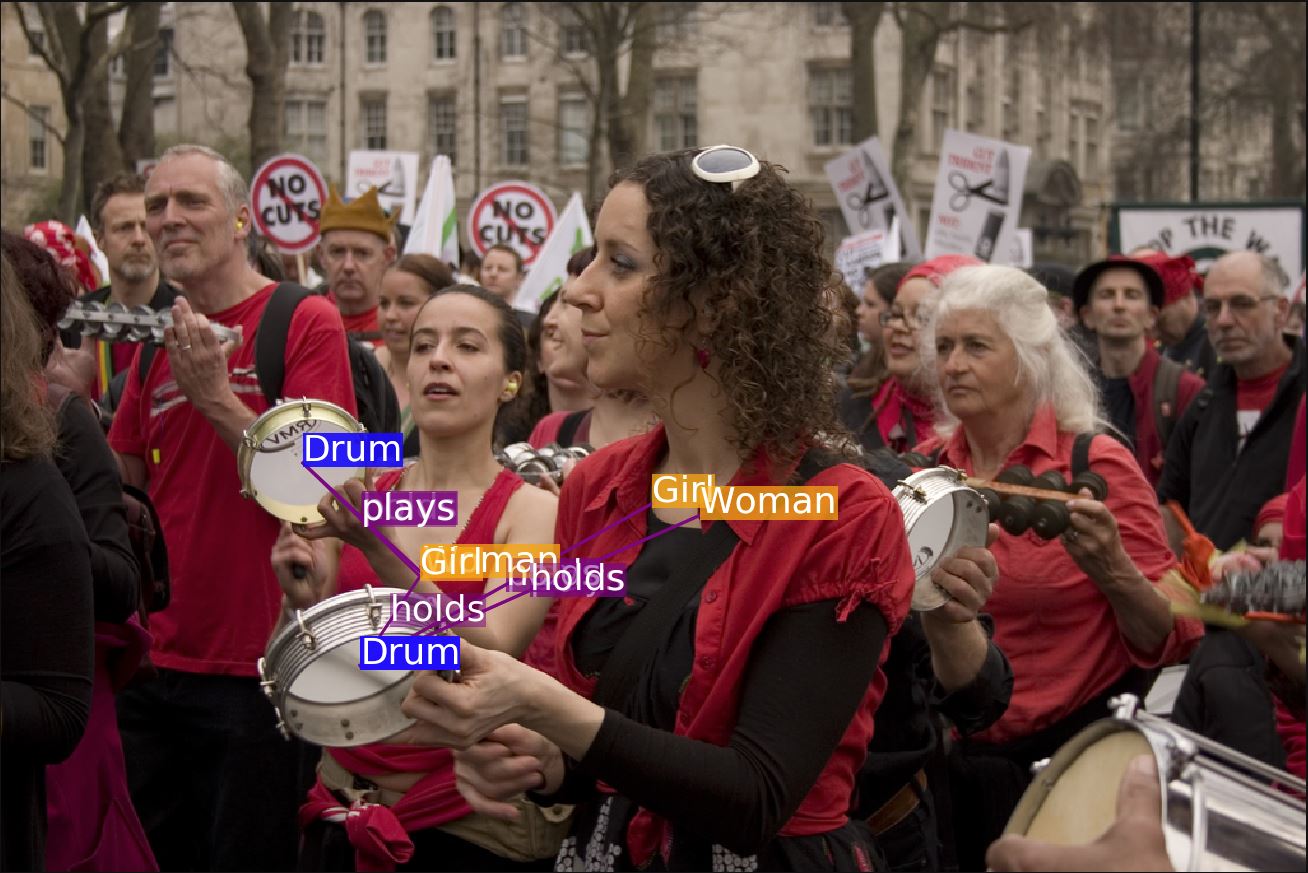}
    \end{subfigure}
    \centering
    \begin{subfigure}{0.32\columnwidth}
      \centering
      \includegraphics[width=\textwidth]{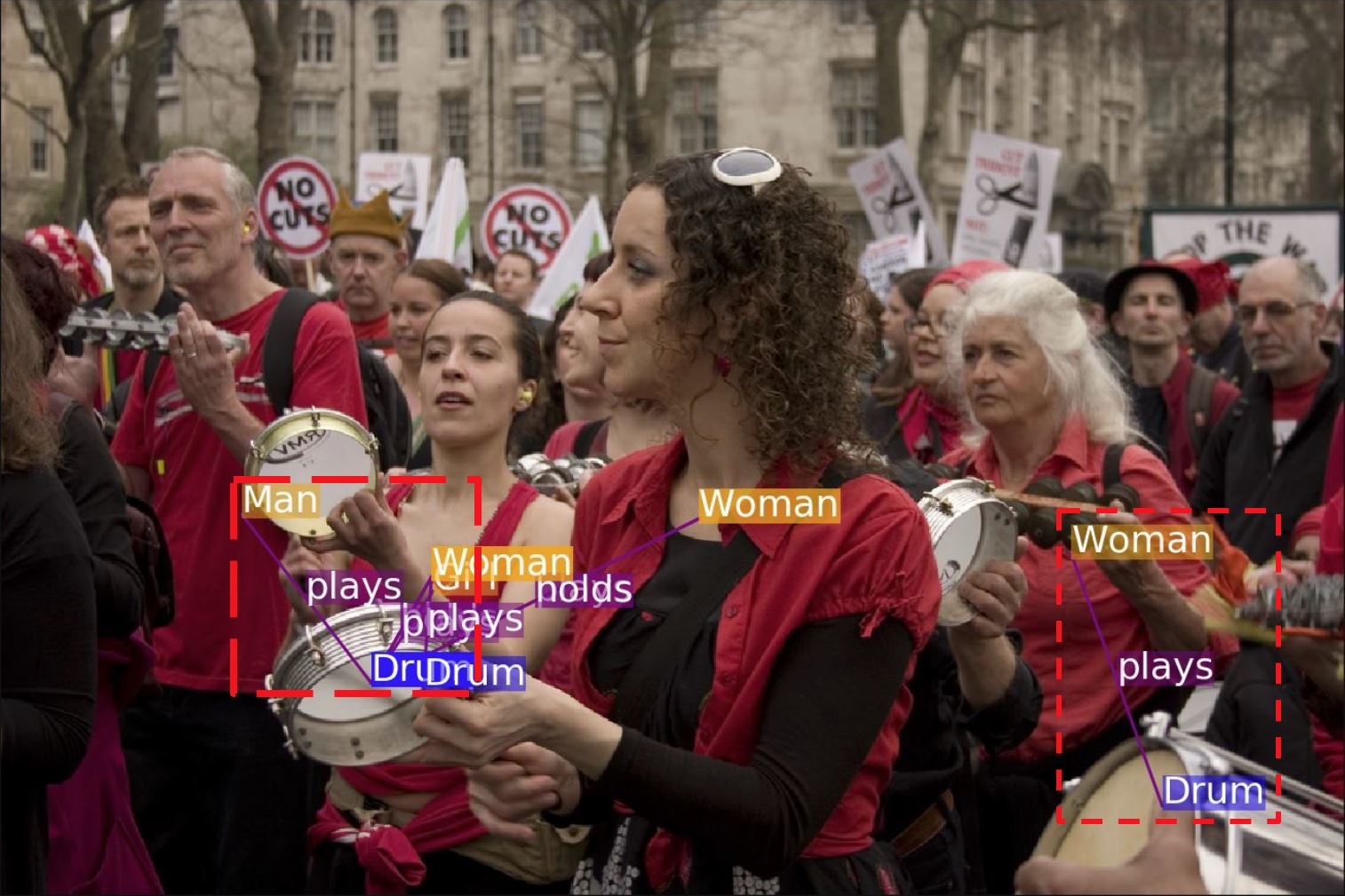}
    \end{subfigure}
    \centering
    \begin{subfigure}{0.32\columnwidth}
      \centering
      \includegraphics[width=\textwidth]{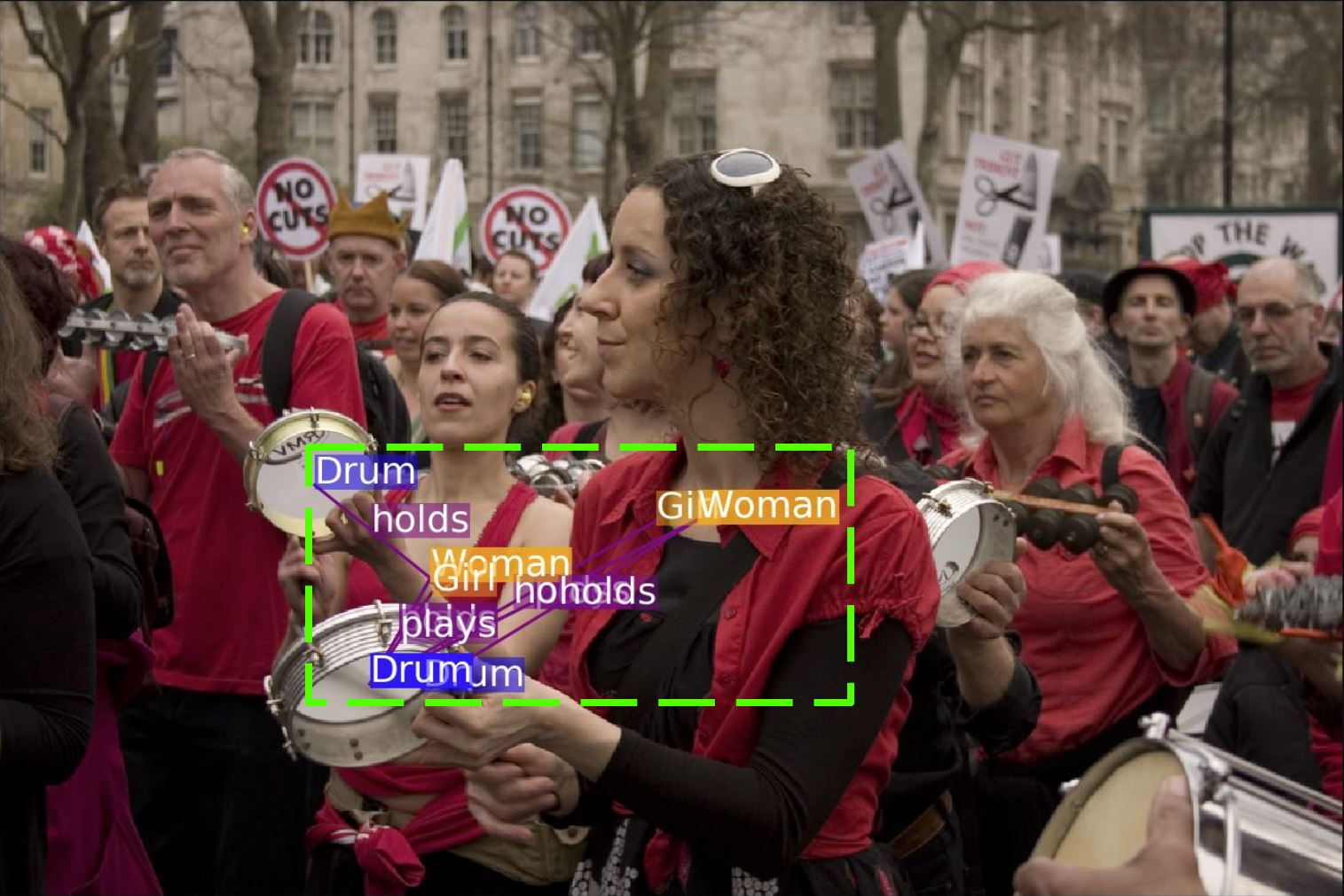}
    \end{subfigure}
  \end{subfigure}
  
  \centering
  \begin{subfigure}{2.1\columnwidth}
    \centering
    \begin{subfigure}{0.32\columnwidth}
      \centering
      \includegraphics[width=\textwidth]{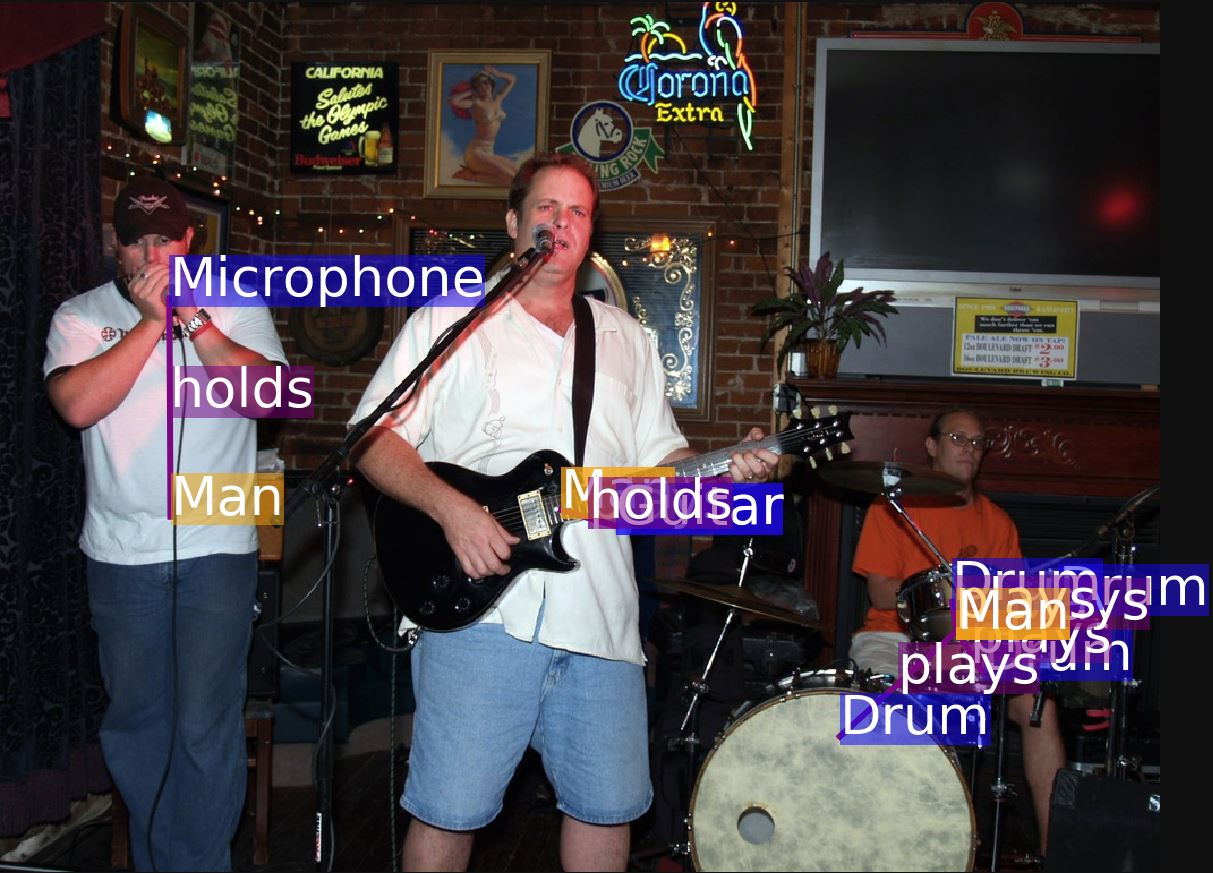}
    \end{subfigure}
    \centering
    \begin{subfigure}{0.32\columnwidth}
      \centering
      \includegraphics[width=\textwidth]{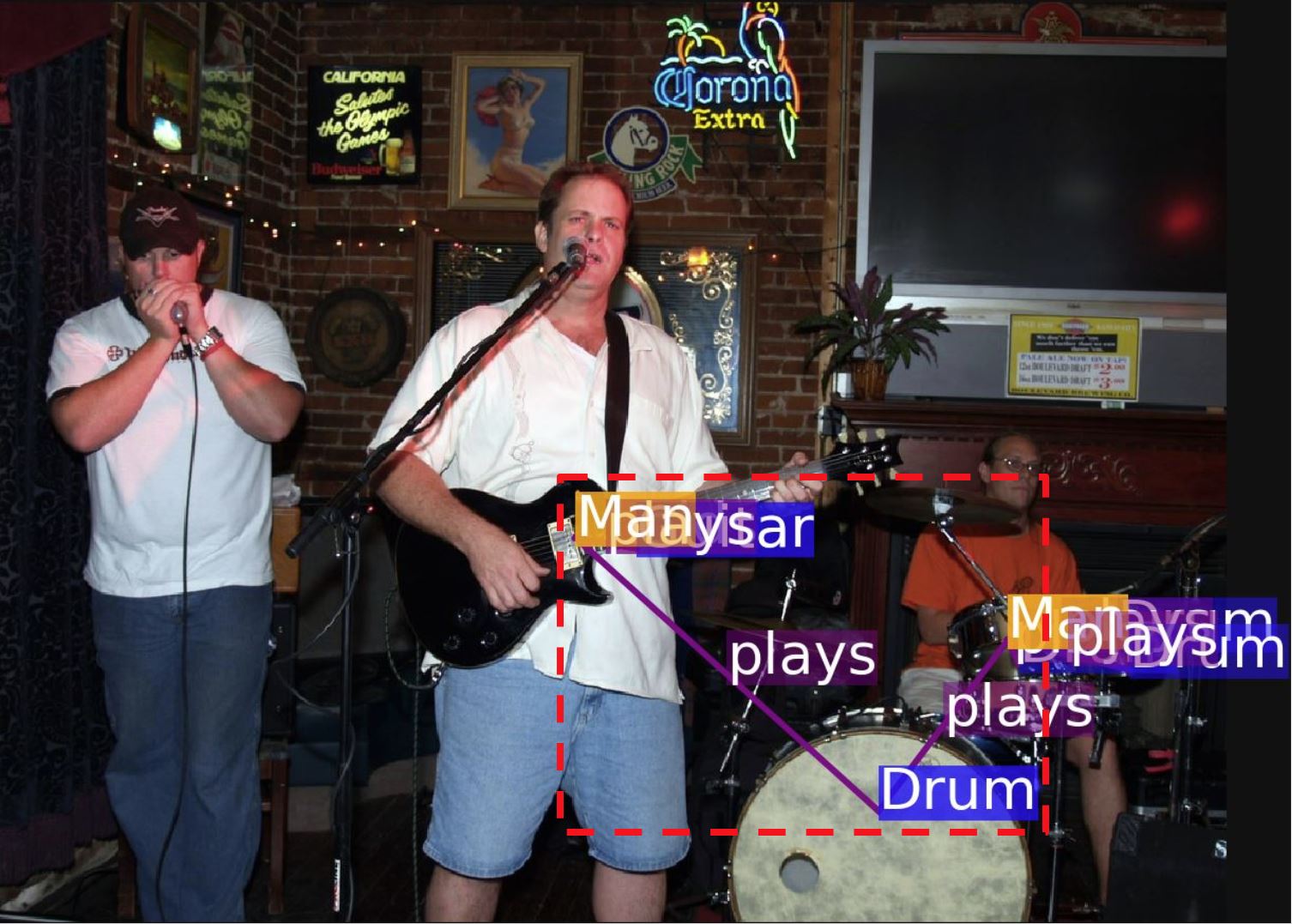}
    \end{subfigure}
    \centering
    \begin{subfigure}{0.32\columnwidth}
      \centering
      \includegraphics[width=\textwidth]{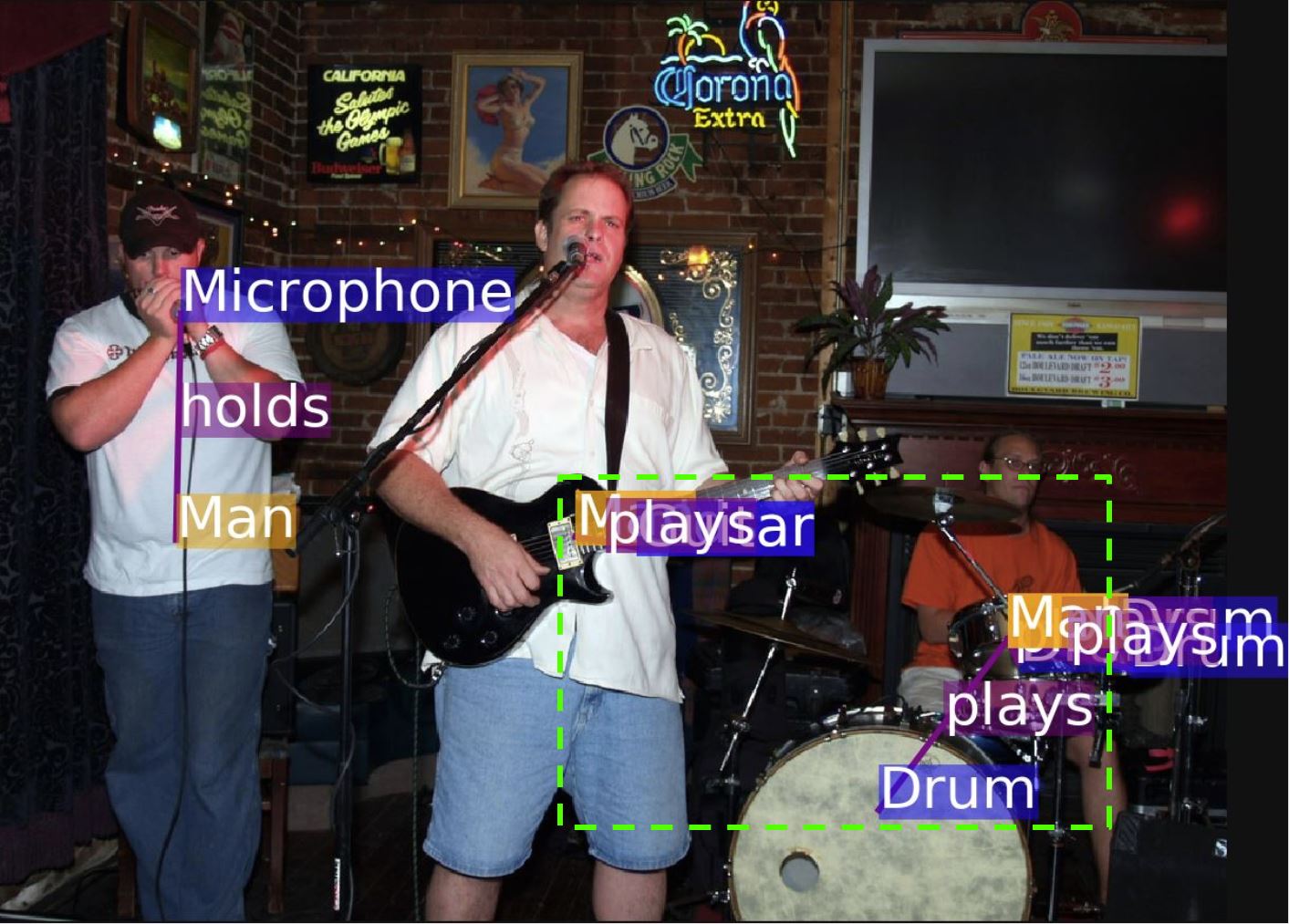}
    \end{subfigure}
  \end{subfigure}
  
  \centering
  \begin{subfigure}{2.1\columnwidth}
    \centering
    \begin{subfigure}{0.32\columnwidth}
      \centering
      \includegraphics[width=\textwidth]{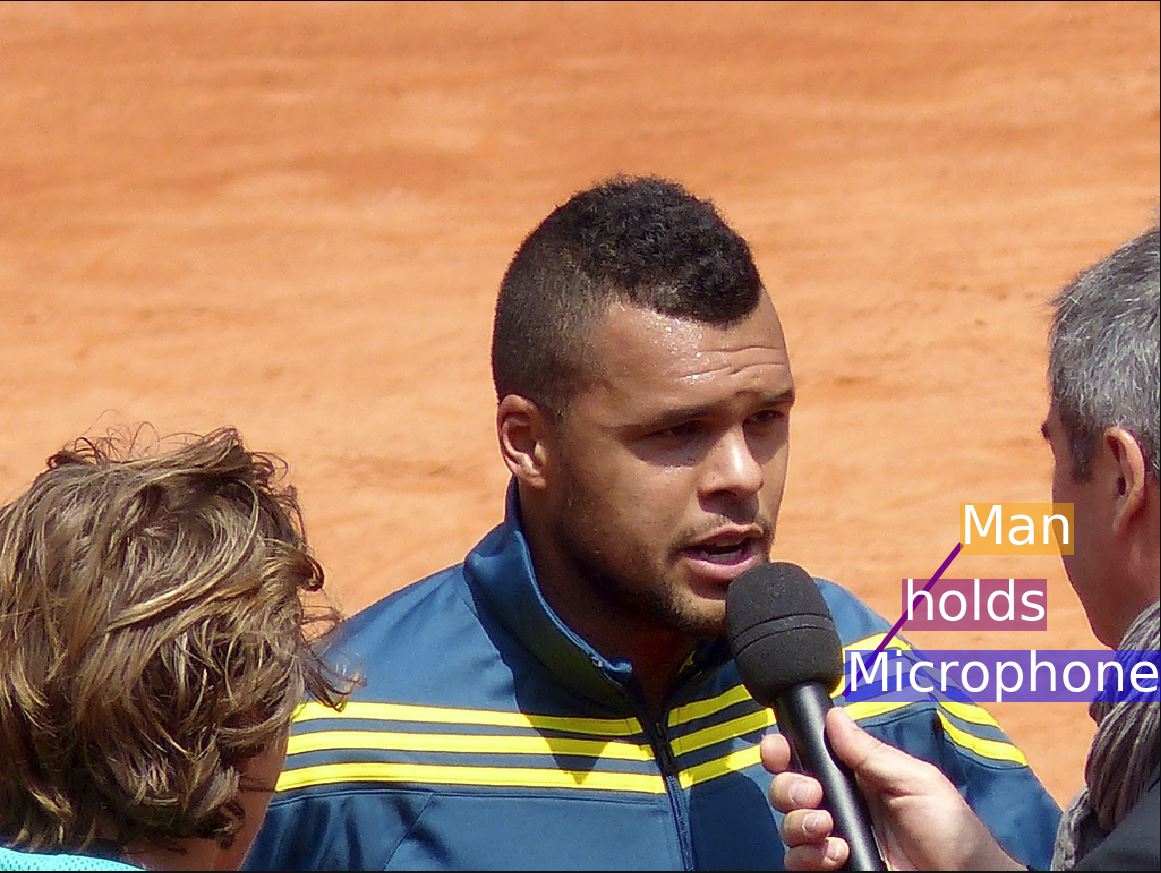}
    \end{subfigure}
    \centering
    \begin{subfigure}{0.32\columnwidth}
      \centering
      \includegraphics[width=\textwidth]{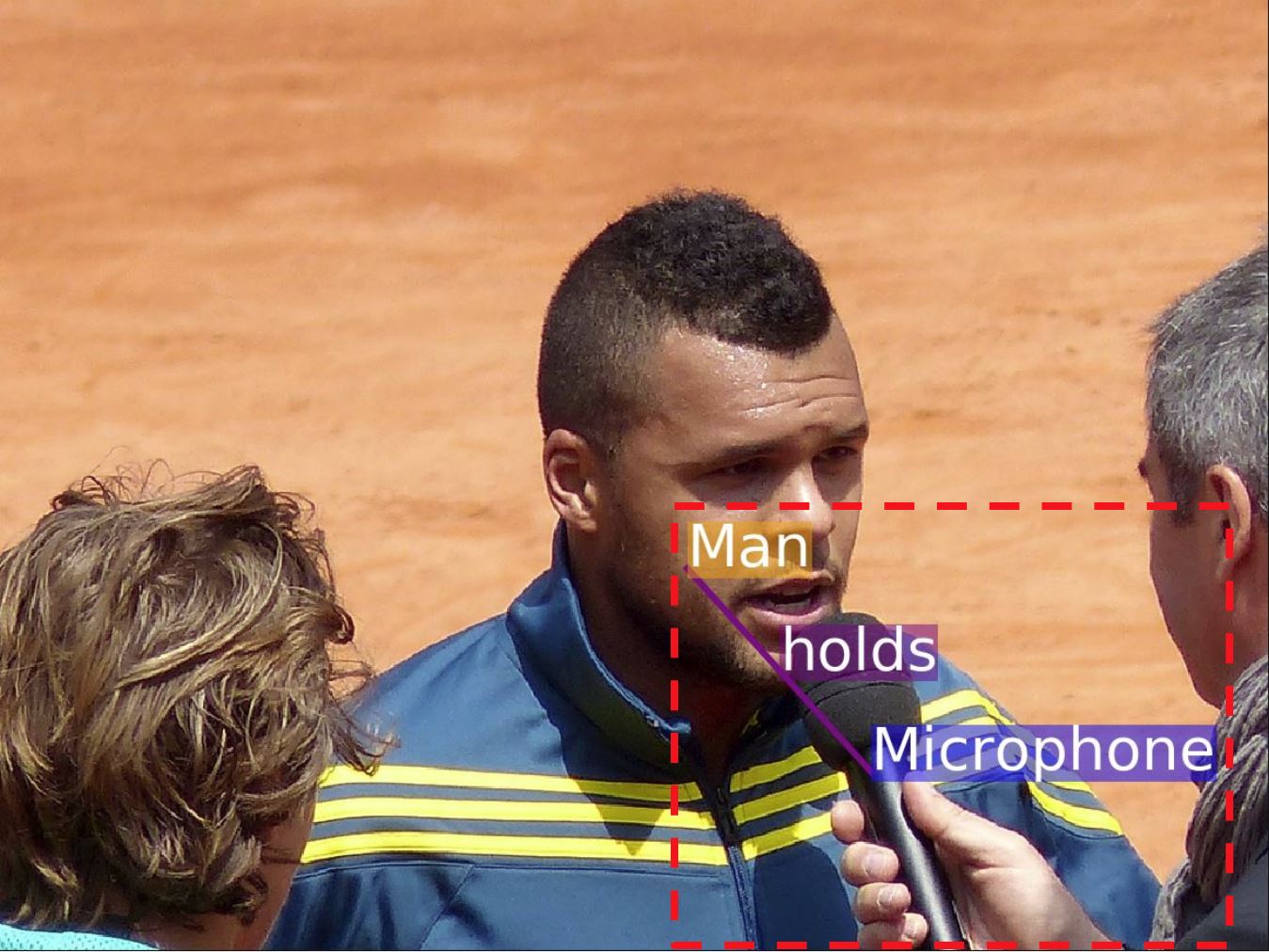}
    \end{subfigure}
    \centering
    \begin{subfigure}{0.32\columnwidth}
      \centering
      \includegraphics[width=\textwidth]{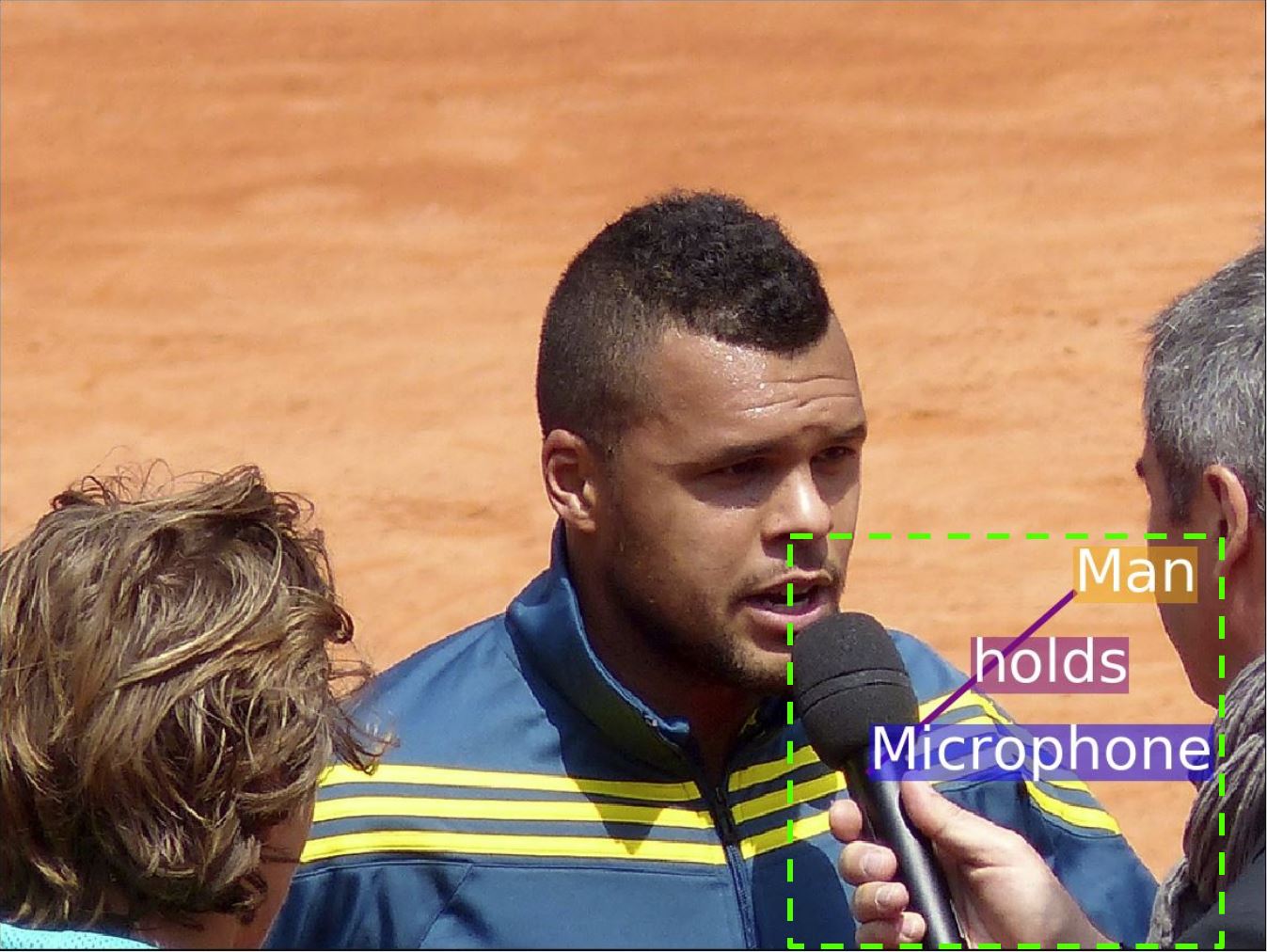}
    \end{subfigure}
  \end{subfigure}
  
  \centering
  \begin{subfigure}{2.1\columnwidth}
    \centering
    \begin{subfigure}{0.32\columnwidth}
      \centering
      \includegraphics[width=\textwidth]{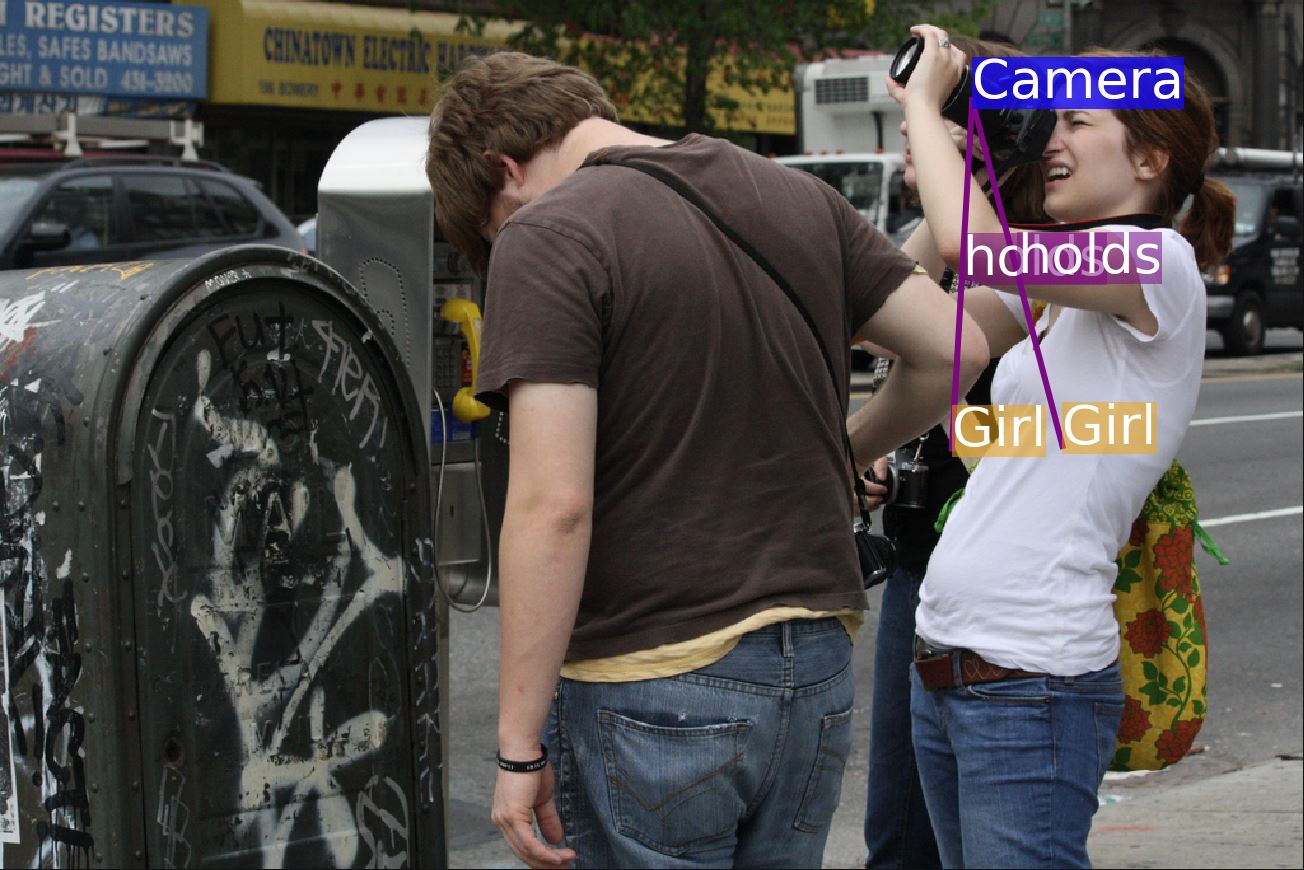}
    \caption{ground truth}
    \label{fig:gt}
    \end{subfigure}
    \centering
    \begin{subfigure}{0.32\columnwidth}
      \centering
      \includegraphics[width=\textwidth]{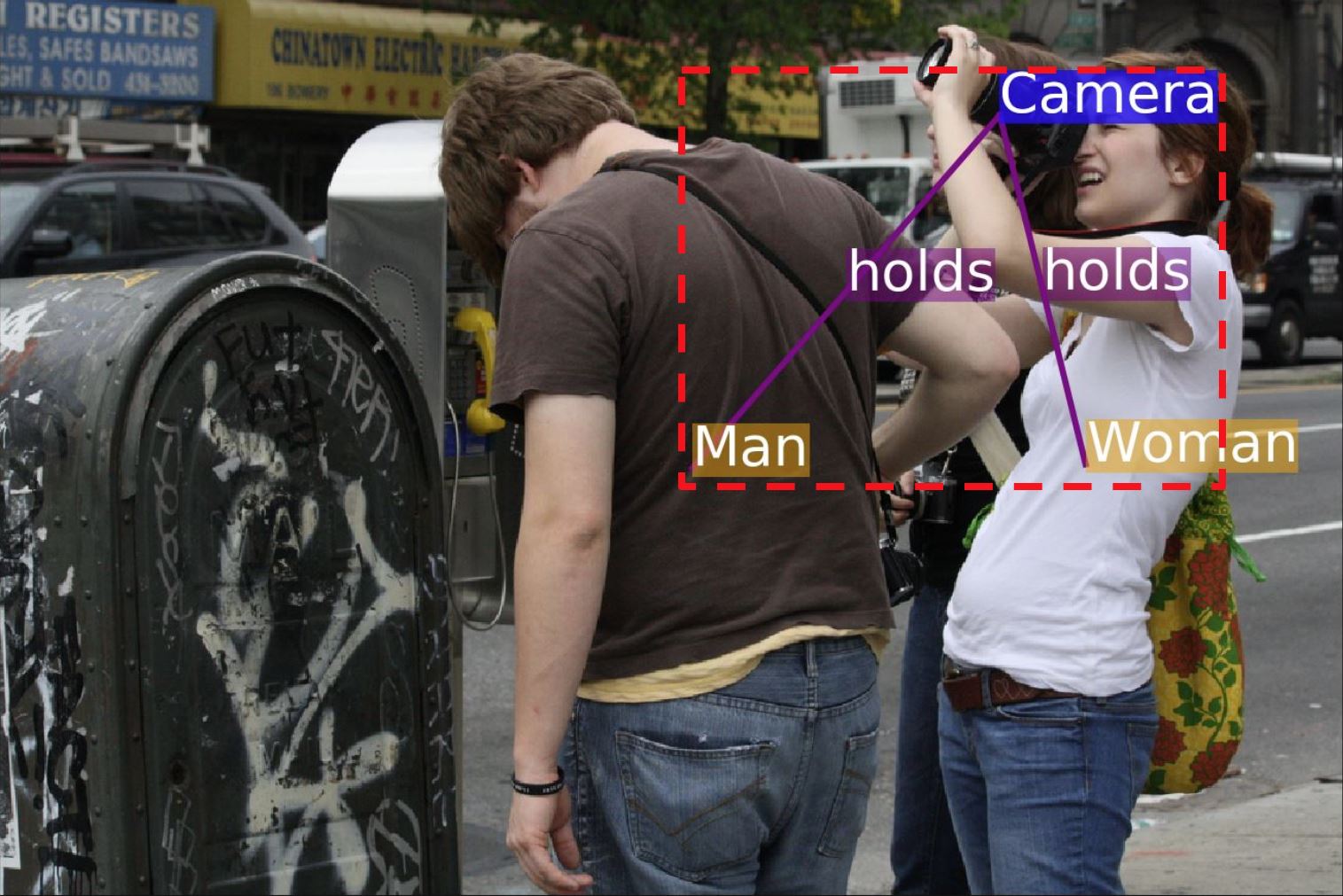}
    \caption{$L_0$ only}
    \label{fig:det_conv}
    \end{subfigure}
    \centering
    \begin{subfigure}{0.32\columnwidth}
      \centering
      \includegraphics[width=\textwidth]{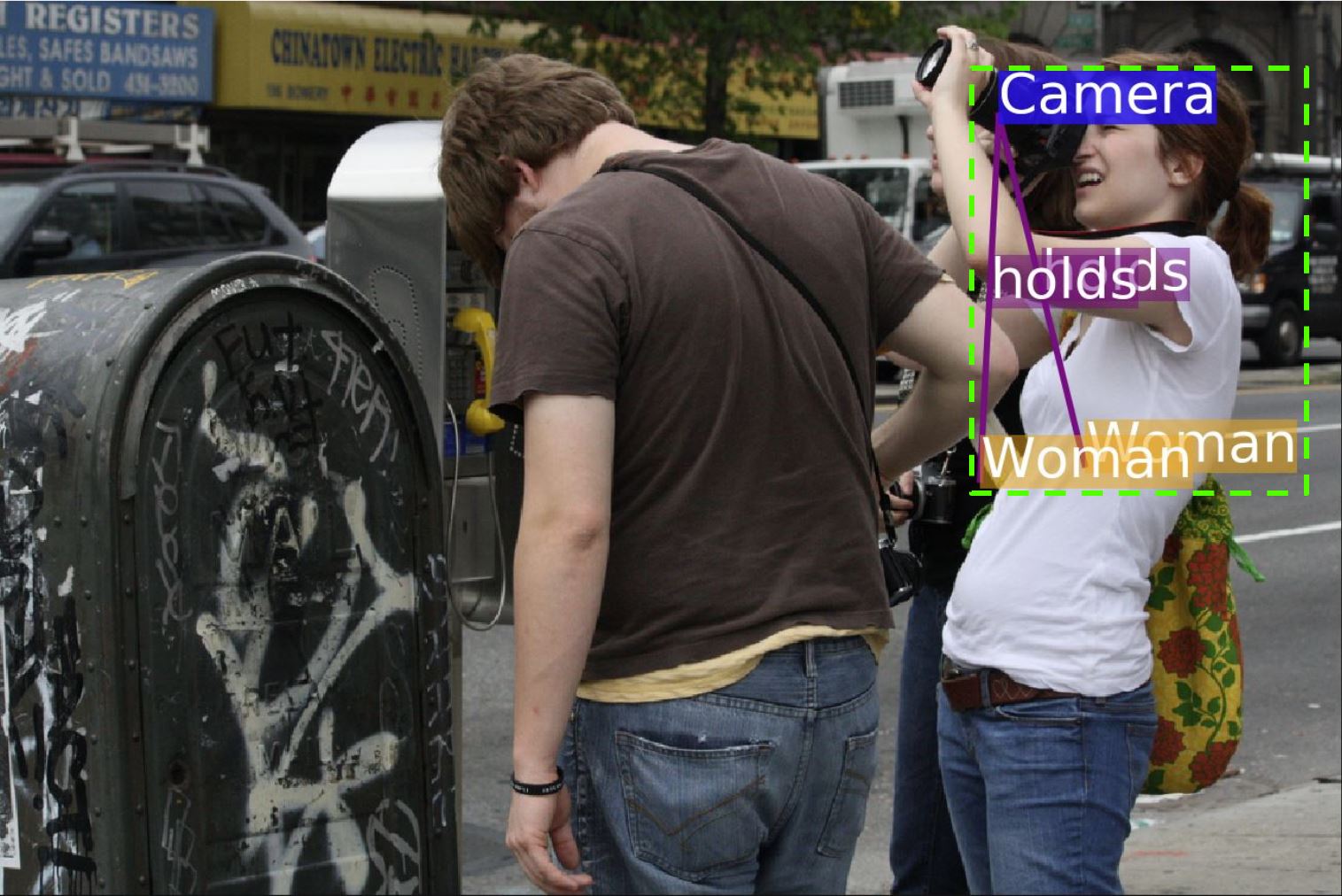}
    \caption{all losses}
    \label{fig:prd_conv}
    \end{subfigure}
  \end{subfigure}
\setlength\belowcaptionskip{-2ex}
\captionsetup{font=small}
\caption{Example images where RelDN with only $L_0$ predicts incorrectly while our loss succeeds. For each image we check the number of its ground truth relationships, then we output the same number of top predictions from a model to see its ranking accuracy. Red boxes in (b) highlight the false predictions from RelDN with $L_0$ only and green boxes in (c) highlight the correct ones from RelDN with all losses.}
\label{fig:qualitative}
\end{figure*}

In Figure \ref{fig:qualitative} we provide four example images where our losses correct the false predictions made by the $L_0$ only model. Both the Entity Instance Confusion and the Proximal Relationship Ambiguity issues are included here. In the fourth row, the $L_0$ only model is confused between two entity instances, \ie, which person is holding the microphone, while our losses manage to refer to the correct one. In the third row the relationship between the guitar player and the drum is ambiguous. Here, the $L_0$ only model fails by predicting a false-positive, but our model trained with all losses correctly detects no relationship there.

\section{Conclusion}
\label{sec:conclusion}

In this work we present methods to overcome two major issues in scene graph parsing: Entity Instance Confusion and Proximal Relationship Ambiguity. We show that traditional multi-class cross-entropy loss does not take advantage of intrinsic knowledge of structured scene graphs and is therefore insufficient to handle these two issues. To address that, we propose Graphical Contrastive Losses which effectively utilize semantic properties of scene graphs to contrast positive relationships against hard negatives. We carefully design three types of losses to solve the issues in three aspects. We demonstrate efficacy of our losses by adding it to a model built with the same pipeline, and we achieve state-of-the-art results on three datasets.

{\small
\bibliographystyle{ieee}
\bibliography{egbib}

\begin{thebibliography}{10}\itemsep=-1pt

\bibitem{openimages_challenge}
Openimages vrd challenge.
\newblock \url{https://storage.googleapis.com/openimages/web/challenge.html}.

\bibitem{chen2017query}
K.~Chen, R.~Kovvuri, and R.~Nevatia.
\newblock Query-guided regression network with context policy for phrase
  grounding.
\newblock In {\em ICCV}, 2017.

\bibitem{dai2017detecting}
B.~Dai, Y.~Zhang, and D.~Lin.
\newblock Detecting visual relationships with deep relational networks.
\newblock In {\em CVPR}, 2017.

\bibitem{gkioxari2017interactnet}
G.~Gkioxari, R.~Girshick, P.~Doll\'{a}r, and K.~He.
\newblock Detecting and recognizing human-object intaractions.
\newblock {\em CVPR}, 2018.

\bibitem{gupta2017aligned}
T.~Gupta, K.~J. Shih, S.~Singh, and D.~Hoiem.
\newblock Aligned image-word representations improve inductive transfer across
  vision-language tasks.
\newblock In {\em ICCV}, 2017.

\bibitem{hu2017modeling}
R.~Hu, M.~Rohrbach, J.~Andreas, T.~Darrell, and K.~Saenko.
\newblock Modeling relationships in referential expressions with compositional
  modular networks.
\newblock In {\em CVPR}, 2017.

\bibitem{hu2016natural}
R.~Hu, H.~Xu, M.~Rohrbach, J.~Feng, K.~Saenko, and T.~Darrell.
\newblock Natural language object retrieval.
\newblock In {\em CVPR}, 2016.

\bibitem{kiros2014unifying}
R.~Kiros, R.~Salakhutdinov, R.~S. Zemel, and et~al.
\newblock Unifying visual-semantic embeddings with multimodal neural language
  models.
\newblock {\em TACL}, 2015.

\bibitem{openimages}
I.~Krasin, T.~Duerig, N.~Alldrin, V.~Ferrari, S.~Abu-El-Haija, A.~Kuznetsova,
  H.~Rom, J.~Uijlings, S.~Popov, S.~Kamali, M.~Malloci, J.~Pont-Tuset, A.~Veit,
  S.~Belongie, V.~Gomes, A.~Gupta, C.~Sun, G.~Chechik, D.~Cai, Z.~Feng,
  D.~Narayanan, and K.~Murphy.
\newblock Openimages: A public dataset for large-scale multi-label and
  multi-class image classification.
\newblock {\em Dataset available from
  https://storage.googleapis.com/openimages/web/index.html}, 2017.

\bibitem{krishnavisualgenome}
R.~Krishna, Y.~Zhu, O.~Groth, J.~Johnson, K.~Hata, J.~Kravitz, S.~Chen,
  Y.~Kalantidis, L.-J. Li, D.~A. Shamma, et~al.
\newblock Visual genome: Connecting language and vision using crowdsourced
  dense image annotations.
\newblock {\em IJCV}, 2017.

\bibitem{LiCVPR2017}
Y.~Li, W.~Ouyang, and X.~Wang.
\newblock Vip-cnn: {A} visual phrase reasoning convolutional neural network for
  visual relationship detection.
\newblock In {\em CVPR}, 2017.

\bibitem{liang2017deep}
X.~Liang, L.~Lee, and E.~P. Xing.
\newblock Deep variation-structured reinforcement learning for visual
  relationship and attribute detection.
\newblock {\em arXiv preprint arXiv:1703.03054}, 2017.

\bibitem{lin2017feature}
T.-Y. Lin, P.~Doll{\'a}r, R.~B. Girshick, K.~He, B.~Hariharan, and S.~J.
  Belongie.
\newblock Feature pyramid networks for object detection.
\newblock In {\em CVPR}, 2017.

\bibitem{liu2017referring}
J.~Liu, L.~Wang, M.-H. Yang, et~al.
\newblock Referring expression generation and comprehension via attributes.
\newblock In {\em CVPR}, 2017.

\bibitem{lu2016visual}
C.~Lu, R.~Krishna, M.~Bernstein, and L.~Fei-Fei.
\newblock Visual relationship detection with language priors.
\newblock In {\em ECCV}, 2016.

\bibitem{luo2017comprehension}
R.~Luo and G.~Shakhnarovich.
\newblock Comprehension-guided referring expressions.
\newblock In {\em CVPR}, 2017.

\bibitem{mao2016generation}
J.~Mao, J.~Huang, A.~Toshev, O.~Camburu, A.~L. Yuille, and K.~Murphy.
\newblock Generation and comprehension of unambiguous object descriptions.
\newblock In {\em CVPR}, 2016.

\bibitem{mikolov2013distributed}
T.~Mikolov, I.~Sutskever, K.~Chen, G.~S. Corrado, and J.~Dean.
\newblock Distributed representations of words and phrases and their
  compositionality.
\newblock In {\em NIPS}, 2013.

\bibitem{mnih2013learning}
A.~Mnih and K.~Kavukcuoglu.
\newblock Learning word embeddings efficiently with noise-contrastive
  estimation.
\newblock In {\em NIPS}, 2013.

\bibitem{nagaraja2016modeling}
V.~K. Nagaraja, V.~I. Morariu, and L.~S. Davis.
\newblock Modeling context between objects for referring expression
  understanding.
\newblock In {\em ECCV}, 2016.

\bibitem{newell2017pixels}
A.~Newell and J.~Deng.
\newblock Pixels to graphs by associative embedding.
\newblock In {\em NIPS}, 2017.

\bibitem{Peyre17}
J.~Peyre, I.~Laptev, C.~Schmid, and J.~Sivic.
\newblock Weakly-supervised learning of visual relations.
\newblock In {\em ICCV}, 2017.

\bibitem{plummer2015flickr30k}
B.~Plummer, L.~Wang, C.~Cervantes, J.~Caicedo, J.~Hockenmaier, and S.~Lazebnik.
\newblock Flickr30k entities: Collecting region-to-phrase correspondences for
  richer image-to-sentence models.
\newblock In {\em ICCV}, 2015.

\bibitem{ren2015faster}
S.~Ren, K.~He, R.~Girshick, and J.~Sun.
\newblock Faster r-cnn: Towards real-time object detection with region proposal
  networks.
\newblock In {\em NIPS}, 2015.

\bibitem{rohrbach2016grounding}
A.~Rohrbach, M.~Rohrbach, R.~Hu, T.~Darrell, and B.~Schiele.
\newblock Grounding of textual phrases in images by reconstruction.
\newblock In {\em ECCV}, 2016.

\bibitem{vendrov2015order}
I.~Vendrov, R.~Kiros, S.~Fidler, and R.~Urtasun.
\newblock Order-embeddings of images and language.
\newblock In {\em ICLR}, 2016.

\bibitem{wang2016learning}
L.~Wang, Y.~Li, and S.~Lazebnik.
\newblock Learning deep structure-preserving image-text embeddings.
\newblock In {\em CVPR}, 2016.

\bibitem{xie2017aggregated}
S.~Xie, R.~Girshick, P.~Doll{\'a}r, Z.~Tu, and K.~He.
\newblock Aggregated residual transformations for deep neural networks.
\newblock In {\em CVPR}, 2017.

\bibitem{xu2017scenegraph}
D.~Xu, Y.~Zhu, C.~Choy, and L.~Fei-Fei.
\newblock Scene graph generation by iterative message passing.
\newblock In {\em CVPR}, 2017.

\bibitem{Yang_2018_ECCV}
J.~Yang, J.~Lu, S.~Lee, D.~Batra, and D.~Parikh.
\newblock Graph r-cnn for scene graph generation.
\newblock In {\em ECCV}, 2018.

\bibitem{Xu_2018_ECCV}
X.~Yang, H.~Zhang, and J.~Cai.
\newblock Shuffle-then-assemble: Learning object-agnostic visual relationship
  features.
\newblock In {\em ECCV}, 2018.

\bibitem{Yin_2018_ECCV}
G.~Yin, L.~Sheng, B.~Liu, N.~Yu, X.~Wang, J.~Shao, and C.~Change~Loy.
\newblock Zoom-net: Mining deep feature interactions for visual relationship
  recognition.
\newblock In {\em ECCV}, 2018.

\bibitem{yu2018mattnet}
L.~Yu, Z.~Lin, X.~Shen, J.~Yang, X.~Lu, M.~Bansal, and T.~L. Berg.
\newblock Mattnet: Modular attention network for referring expression
  comprehension.
\newblock In {\em CVPR}, 2018.

\bibitem{yu2016modeling}
L.~Yu, P.~Poirson, S.~Yang, A.~C. Berg, and T.~L. Berg.
\newblock Modeling context in referring expressions.
\newblock In {\em ECCV}, 2016.

\bibitem{yu17iccv}
R.~Yu, A.~Li, V.~I. Morariu, and L.~S. Davis.
\newblock Visual relationship detection with internal and external linguistic
  knowledge distillation.
\newblock In {\em ICCV}, 2017.

\bibitem{zagoruyko2016paying}
S.~Zagoruyko and N.~Komodakis.
\newblock Paying more attention to attention: Improving the performance of
  convolutional neural networks via attention transfer.
\newblock In {\em ICLR}, 2017.

\bibitem{zellers2018neural}
R.~Zellers, M.~Yatskar, S.~Thomson, and Y.~Choi.
\newblock Neural motifs: Scene graph parsing with global context.
\newblock In {\em CVPR}, 2018.

\bibitem{zhang2017visual}
H.~Zhang, Z.~Kyaw, S.-F. Chang, and T.-S. Chua.
\newblock Visual translation embedding network for visual relation detection.
\newblock In {\em CVPR}, 2017.

\bibitem{zhang2017ppr}
H.~Zhang, Z.~Kyaw, J.~Yu, and S.-F. Chang.
\newblock Ppr-fcn: Weakly supervised visual relation detection via parallel
  pairwise r-fcn.
\newblock In {\em Proceedings of the IEEE Conference on Computer Vision and
  Pattern Recognition}, pages 4233--4241, 2017.

\bibitem{zhang2017relationship}
J.~Zhang, M.~Elhoseiny, S.~Cohen, W.~Chang, and A.~Elgammal.
\newblock Relationship proposal networks.
\newblock In {\em CVPR}, 2017.

\bibitem{zhang2018large}
J.~Zhang, Y.~Kalantidis, M.~Rohrbach, M.~Paluri, A.~Elgammal, and M.~Elhoseiny.
\newblock Large-scale visual relationship understanding.
\newblock In {\em AAAI}, 2019.

\bibitem{zhang2018neurips}
J.~Zhang, K.~Shih, A.~Tao, B.~Catanzaro, and A.~Elgammal.
\newblock An interpretable model for scene graph generation.
\newblock {\em arXiv preprint arXiv:1811.09543}, 2018.

\bibitem{zhang2018introduction}
J.~Zhang, K.~Shih, A.~Tao, B.~Catanzaro, and A.~Elgammal.
\newblock Introduction to the 1st place winning model of openimages
  relationship detection challenge.
\newblock {\em arXiv preprint arXiv:1811.00662}, 2018.

\bibitem{Zhuang_2017_ICCV}
B.~Zhuang, L.~Liu, C.~Shen, and I.~Reid.
\newblock Towards context-aware interaction recognition for visual relationship
  detection.
\newblock In {\em ICCV}, 2017.

\end{thebibliography}
}

\end{document}